\documentclass{article} 
\usepackage{iclr2027_conference,times}

\usepackage{amsmath,amsfonts,bm}

\def\eqref#1{equation~\ref{#1}}

\def\1{\bm{1}}

\DeclareMathAlphabet{\mathsfit}{\encodingdefault}{\sfdefault}{m}{sl}
\SetMathAlphabet{\mathsfit}{bold}{\encodingdefault}{\sfdefault}{bx}{n}

\usepackage{hyperref}
\usepackage{url}
\usepackage{afterpage}
\usepackage{enumitem}

\usepackage{xspace}
\usepackage[font=small]{subcaption}
\usepackage{graphicx} 
\usepackage[dvipsnames]{xcolor}

\usepackage{lipsum}
\usepackage{multirow}
\usepackage{multicol}
\usepackage{soul}
\usepackage[cjk]{kotex}

\newcommand{\ie}{\textit{i.e.}\xspace}
\newcommand{\eg}{\textit{e.g.}\xspace}

\definecolor{starred}{RGB}{227, 72, 70} 
\definecolor{lightred}{RGB}{255, 210, 210} 
\newcommand{\ours}{\textsc{Omni-Catalyst}}

\usepackage{booktabs, multirow, graphicx, colortbl, xcolor}
\usepackage{makecell, tabularx}

\usepackage{tikz}
\usepackage{url} 
\definecolor{cornellred}{rgb}{0.7, 0.11, 0.11}
\definecolor{cadmiumgreen}{rgb}{0.0, 0.42, 0.24}
\definecolor{aliceblue}{rgb}{0.91, 0.94, 0.97}
\definecolor{darkblue}{rgb}{0.83, 0.89, 0.97}
\definecolor{Red7}{rgb}{0.941, 0.243, 0.243}
\definecolor{Green7}{RGB}{55, 178, 77}
\definecolor{Indigo}{RGB}{75, 0, 130}
\definecolor{Blue9}{rgb}{0.098,0.3,0.9}
\definecolor{think}{HTML}{B5739D}
\definecolor{omni}{HTML}{67AB9F}
\definecolor{Blue1}{HTML}{0000CC}
\definecolor{deepgreen}{HTML}{008000}
\definecolor{coral}{HTML}{EC7161}
\definecolor{myyellow}{HTML}{F7C752}
\definecolor{myorange}{HTML}{E07C3E}

\hypersetup{
  linkcolor = cornellred,
  citecolor  = cadmiumgreen,
  colorlinks = true,
  urlcolor = Blue9
}

\usepackage{wrapfig}
\newcommand{\up}[1]{\textsubscript{\textcolor{teal}{+#1}}}
\newcommand{\down}[1]{\textsubscript{\textcolor{red}{-#1}}}

\usepackage{MnSymbol}
\usepackage{bbold}

\title{Omni-Modal LLMs Know More Than They Say}

\author{
Na Min An\textsuperscript{1}
\quad Taekyung Kim\textsuperscript{2}
\quad Jaemyung Yu\textsuperscript{2}
\quad Hyunjung Shim\textsuperscript{1,$\dagger$}
\quad Dongyoon Han\textsuperscript{2,$\dagger$} \\
\textsuperscript{1}KAIST AI\quad
\textsuperscript{2}NAVER AI Lab \\
\texttt{\{naminan\}@kaist.ac.kr}
}

\iclrfinalcopy 
\begin{document}

\maketitle
\blfootnote{$\dagger$~Corresponding author}

\renewcommand{\thefootnote}{\arabic{footnote}}
\setcounter{footnote}{0}

\begin{abstract}
Omni-modal large language models (Omni-LLMs) extend LLMs to process heterogeneous modalities (\eg, vision, audio, and IMU signals) alongside language, enabling a broader range of multimodal tasks.
Yet we find that these models often fail to express information from non-textual modalities even when they perceive it correctly.
We term this disconnect \emph{knows-but-cannot-say}.
To investigate this phenomenon, we introduce Token-level Modality Alignment (TMA), which measures the alignment between individual modality representations and response representations.
Our analysis shows that non-textual representations are less aligned with response tokens and receive less attention than textual representations, suggesting a potential explanation for uneven modality preferences.
Motivated by these findings, we propose \ours{}, a simple \emph{perceive} $\rightarrow$ \emph{infer} strategy that complements the original modality inputs with self-generated text, making non-textual knowledge more accessible.
Across seven Omni-LLMs, \ours{} consistently improves performance (+7.29\%) and yields more balanced modality preferences, with further gains from diverse training strategies.
Further analysis demonstrates that the knowledge extraction phases increase TMA, suggesting that improved alignment makes non-textual information more accessible during response generation.
In summary, our findings uncover \textbf{how Omni-LLMs can better express modality knowledge through language}.
Our code is available at \url{https://github.com/naver-ai/OmniLLM}.
\end{abstract}

\begin{figure*}[h]
    \centering
    \vspace{-1.7em}
    \hspace{-0.5em}
    \begin{subfigure}[t]{0.75\linewidth}
    \vspace{0pt}
        \includegraphics[width=\linewidth]{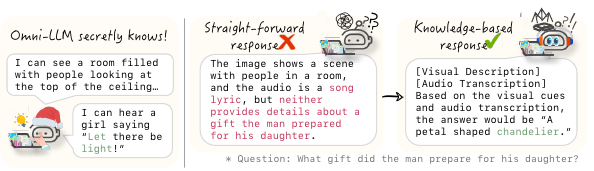}
        \vspace{-1.3em}
        \caption{Modality knowledge extraction}\label{fig:know_cannot_say}
    \end{subfigure}
    \hfill
    \hspace{-1.5em}
    \begin{subfigure}[t]{0.25\linewidth}
    \vspace{0pt}
        \includegraphics[width=\linewidth]{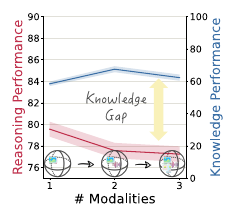}
        \vspace{-1.8em}
        \caption{Multimodality}\label{fig:modality_num_performance}
    \end{subfigure}
    \vspace{-0.5em}
    \caption{\textbf{The ``knows but cannot say'' problem in Omni-LLMs.}
    (a) Omni-LLM often embeds sufficient modality knowledge, yet it produces an incorrect answer when directly responding (left).
    By complementing the relatively weak modality, audio recovers the correct answer (right).
    (b) This phenomenon becomes more pronounced as the number of modalities increases.
    }
    \vspace{-1.2em}
    \label{fig:motivation}
\end{figure*}

\section{Introduction}
\label{sec1:intro}

Omni-modal large-language models (Omni-LLMs)~\citep{han2024onellm,cheng2024videollama,xie2024mini,hurst2024gpt,team2024chameleon,zhan2025anygptunifiedmultimodalllm,ye2025omnivinci,li2025baichuan,ai2025ming,xu2025omni,liu2025ola,chen2025blip3,xu2025qwen25omnitechnicalreport,qwenteam2026qwen35omnitechnicalreport,team2026longcat,gemmateam2026,team2026hyperclova,luo2026openomni,nguyen2026see,dai2026latentomni} integrate diverse sensory modalities within a single language-centric reasoning system\footnote{We use \emph{multimodal} and \emph{omni-modal}  to refer to models supporting two and $>2$ modalities, respectively.}. Their design broadly resembles the organization of human perception, in which specialized pathways process distinct sensory streams before integrating them for higher-level cognition~\citep{mcclure1916perception,stokes2021thinking,beck2025between}. Omni-LLMs encode inputs such as vision and audio through modality-specific encoders, then project these representations into a shared space for the underlying language model to generate a response. While this unified architecture enables richer interaction across sensory channels, Omni-LLMs often still underperform modality-specialized counterparts on the very tasks those models are designed for~\citep{zhu2025extending}. This raises a fundamental question: \emph{why do Omni-LLMs fail to fully benefit from the additional modalities available to them?} Limited training data and suboptimal training strategies may partly explain this gap. However, another possible explanation may be that the limitation lies not only in perception itself, but also in how effectively Omni-LLMs can access perceived information during the subsequent generation process.

In this work, we investigate the gap between what Omni-LLMs perceive and what they use to generate an answer. Figure~\ref{fig:motivation} presents a motivating illustration of this discrepancy and how it evolves as the number of input modalities increases. As shown in Figure~\ref{fig:know_cannot_say}, the Omni-LLM can correctly extract relevant modality information but fails to use it when directly answering the multimodal reasoning question. Remarkably, once the same modality-specific information is explicitly elicited in textual form, the model recovers the correct answer. This suggests a \emph{knows-but-cannot-say} problem~\citep{orgad2025llms,lu2026clip}: the model \emph{knows} the relevant information at the perception level, but that knowledge fails to reach the generation path unless it is re-expressed in text. Crucially, this problem intensifies as additional modalities are introduced (see Figure~\ref{fig:modality_num_performance}), precisely the regime Omni-LLMs are built for. Downstream accuracy collapses with each added modality while perception accuracy holds, isolating the bottleneck in read-out rather than encoding.

Our study investigates the hidden \emph{modality} knowledge problem through cross-modality representational alignment. Although Omni-LLMs are extensively trained to align heterogeneous modality representations~\citep{liang2021multibench,wang2025xmodbench,zhao2025unified,chowdhury2026magnet,li2026omnibench}, irreducible misalignment persists~\citep{liang2022mind,furst2022cloob,yaras2024explaining,fonseca2025text,li2025drcap,grassucci2026closing,schrodi2025two,eslami2025mitigate,yan2026beyond2,groger2026revisiting,chowers2026modality}. To effectively characterize this hidden gap, we introduce \textbf{Token-level Modality Alignment (TMA)}, which complements existing global and local alignment measures by quantifying how input modality information interacts with language generation at a token-level.
In \S\ref{sec3:analyses}, we find that TMA effectively explains why Omni-LLMs cannot fully access non-text modality representations. Notably, modality tokens with higher TMA show greater sensitivity in downstream performance, but not for knowledge extraction, and this effect is exacerbated with more modalities.

Such representational mismatch induces prevailing modality imbalance issue, in which Omni-LLMs rely on modalities that are easier to access rather than those most relevant to the query~\citep{chen2025omnixr,wang2026omni,yan2026beyond,zheng2026overcoming,yang2025audio,huang2022modality,tjandrasuwita2025understanding,fang2026align,reilly2026mixture}. To address this issue, we explore in \S\ref{sec4:method} a simple strategy that renders \textbf{modality-specific information into texts} before answer generation. We propose \ours{} for both training-free and trainable settings and show that complementary, self-generated modality descriptions substantially improve omni-modal reasoning performance (from \S\ref{sec5:1} to \S\ref{sec6:2}). Importantly, while \ours{} is particularly beneficial for complementing weaker modalities in multimodal inputs (\eg, audio given vision-audio), it also generalizes to single-modality settings, consistently improving performance across diverse modalities such as fMRI, IMU, and point clouds. The self-generated descriptions preserve modality-specific semantics while residing in the same representational space as the subsequent answering tokens, thereby bridging heterogeneous modality inputs and language generation. Overall, our results suggest that how much Omni-LLMs can make encoded information usable during generation matters. Our contributions can be summarized as follows: 
\begin{itemize}[leftmargin=*, itemsep=1pt] 
    \item \textbf{Diagnosing omni-modal representations:}
    We trace modality imbalance in Omni-LLMs through representational misalignment between modality-specific inputs and generation representations, using our proposed Token-level Modality Alignment (TMA) metric.
    \item \textbf{Extracting modality knowledge into the generation space}:
    We propose \ours{}, which self-generates descriptions for weaker modalities, steering Omni-LLMs toward the correct answer with their representations acting as an effective semantic and modality intermediate.
    \item \textbf{Attaining modality-balanced gains}: Across diverse Omni-LLMs and benchmarks, \ours{} improves omni-modal reasoning by 7.29\% and increases TMA and attention in the knowledge-extraction phase with enhanced modality balance.
\end{itemize}

\section{Background}
\label{sec2:rel_work}

\noindent\textbf{Omni-LLMs.}
Omni-LLMs integrate heterogeneous modalities within a unified architecture to support multimodal understanding and reasoning~\citep{zhu2025extending,koutoupis2026more,cicchetti2026triangle}.
Early efforts such as VAST~\citep{chen2023vast} established large-scale vision-audio-subtitle-text pretraining, while
One-LLM~\citep{han2024onellm} progressively incorporated additional modalities through mixture-of-experts projection.
Subsequent studies show that modality-specific or joint omni-modal fine-tuning alone does not guarantee genuine omni-modal capability~\citep{zhu2025extending}, and that
bimodal supervision provides limited transfer to omni-modal reasoning~\citep{liu2026boosting}.
NExT-OMNI~\citep{luo2026next} extends the setting to any-to-any generation using discrete flow matching,
More recently, inference-time methods have sought to improve modality use without additional training:
OmniXR~\citep{chen2025omnixr} explicitly extracts modality information before reasoning and replaces the modalities,
ThinkOmni~\citep{guanthinkomni2026} delegates reasoning to external text-only models~\citep{yang2025qwen3technicalreport,guo2025deepseek}, and
CoM~\citep{luo2026chain} decomposes inference into planner, reasoner, and decider roles.
While these approaches improve omni-modal inference through architectural, training, or prompting strategies, our work instead investigates \emph{why} non-textual information can remain underutilized despite being encoded by the model.

\noindent\textbf{Knowing without saying.}
A growing body of work reveals a gap between information encoded internally and information expressed in model outputs.
In language models, intermediate activations can decode answer-related information even when greedy decoding produces an incorrect answer~\citep{orgad2025llms}.
Similar discrepancies arise in vision-language models:
models may attend to correct visual evidence despite answering incorrectly~\citep{liu2026seeing},
encode spatial information that is linearly decodable but not reflected in predictions~\citep{liao2026decodable},
or internally capture negation and spatial/multilingual information that does not reliably reach the output~\citep{lu2026clip,an2026blindpositionbiasedlanguage}.
\citet{zhang2026visual} further shows that useful latent information can be silenced without influencing predictions.
Complementarily, \citet{gou2026reasoning} decouple perception from reasoning using an external captioner, demonstrating that richer visual descriptions can improve reasoning.
These findings motivate our central question: whether Omni-LLMs similarly \emph{know} information in non-textual representations but fail to effectively \emph{use} it during generation.

\noindent\textbf{Modality imbalance.}
Omni-LLMs are designed to integrate heterogeneous modalities, yet their representations and inference behavior need not be balanced.
Representationally, modality-specific embeddings can remain misaligned despite being mapped into a shared latent space, a phenomenon known as the modality gap.
Prior work addresses this gap through alignment and uniformity objectives (U+A)~\citep{wang2020understanding}, post-hoc embedding shifts (MG)~\citep{liang2022mind}, and representation standardization (I0T)~\citep{an-etal-2025-i0t}.
However, reducing representational gaps alone does not necessarily improve downstream performance.
Behaviorally, \citet{yan2026beyond} show that Omni-LLMs may strongly favor visual information even when audio, image, and text are equally available,
while HumanOmni-Speaker~\citep{bai2026humanomni} finds reliance on static visual cues over audio-visual alignment in multi-speaker settings.
Together, these results highlight that representational alignment and modality utilization are distinct; our work connects the two by measuring token-level modality alignment and examining how it relates to downstream use.
\section{Token-Level Modality Alignment and Generation Accessibility}
\label{sec3:analyses}

In this section, we investigate how Omni-LLMs access encoded modality information during response generation.
Two considerations guide our analysis.
First, useful modality information must be accessible during \emph{generation}, so alignment among input representations alone is insufficient.
Second, answering a query requires selectively using relevant \emph{tokens}; high average alignment only across samples, at a dataset-level, may obscure whether those tokens are accessible.
Existing encoder-level multimodal model measures primarily characterize alignment between modalities in a shared latent space~\citep{liang2022mind, huh2024position}. However, such encoder-level alignment may not directly apply to generative Omni-LLMs, whose ultimate responses come from the decoder's internal representations and support more than two modalities. We therefore introduce a token-level diagnostic metric that measures modality alignment between arbitrary input modality representations and the representations underlying response generation, and examine what this alignment implies about how modality knowledge is accessed.
\subsection{Modality Representational Misalignment}
\begin{wrapfigure}{r}{0.5\linewidth}
  \centering
  \vspace{-1.2em}
  \hspace{-1.2em}
    \includegraphics[width=1.05\linewidth]{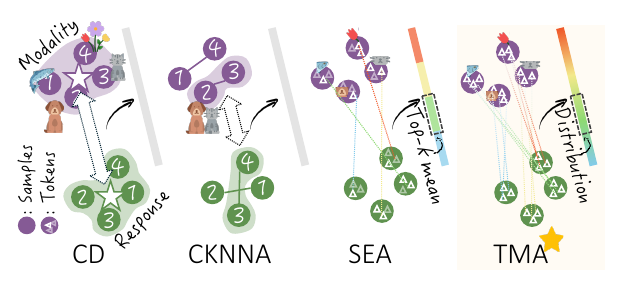}
  \vspace{-2.2em}
  \caption{\textbf{Comparison of modality alignment metrics.}
  Unlike prior metrics, our TMA captures the token-level alignment distribution per sample.
  }
  \vspace{-1.5em}
  \label{fig:alignment_metrics}
\end{wrapfigure}

\noindent\textbf{Evaluation of Omni-LLM modality representations.}  
We first highlight the limitations of existing metrics.
Representative modality alignment (or gap) metrics, such as centroid distance (CD)~\citep{liang2022mind} and CKNNA~\citep{huh2024position} (Figure~\ref{fig:alignment_metrics}), are developed for dual encoders, where 1) each modality sample yields a full embedding; 2) paired embeddings (\eg, an image and its caption) are trained to be semantically aligned. 
That means, each modality spans numerous tokens (often thousands for video); hence, a per-sample embedding may blur which tokens are task-relevant.
Second, input modalities are complementary rather than semantically similar: an audio track is not a paraphrase of the video, nor is the text query.
While \citet{qin2025chain} emphasize alignment to the decoder, they aggregate the final alignment score across tokens. 
Thus, for generative Omni-LLMs, the key question is not how closely the input modalities align dataset-wise, but how strongly each modality token aligns with the representations underlying response generation.  
Yet, token-level granularity alone does not guarantee an effective measurement of modality alignment. For instance, \citet{yin2025sea,xiang2025understanding} examine token-level alignment only across input modalities and aggregate it as a top-$k$ mean (SEA in Figure~\ref{fig:alignment_metrics}). This motivates a new, fine-grained alignment metric that scores each modality token by its maximum similarity to response tokens and tracks its \emph{full distribution} across modality tokens (TMA in Figure~\ref{fig:alignment_metrics}).

\noindent\textbf{Token-level modality alignment (TMA) distributions.}
For a modality $m$, sample $i$, and modality token $j$,
let $H_i^{(m)}={\mathbf{h}_{i,j}^{(m)}}{j=1}, ..., {N_m},$ and $H_i^{(r)}={\mathbf{h}_{i,k}^{(r)}}, {k=1}, ..., {N_r}$
denote the modality and response hidden states, respectively.
We extract the Omni-LLM’s last-decoder-layer hidden state at modality-specific placeholder positions
(\eg, \texttt{Qwen2.5-Omni}’s \texttt{<|video-pad|>}) and the generated response-token positions.
For each modality token $\mathbf{h}_{i,j}^{(m)}$, we define its TMA score as
\begin{equation}
    s_{i,j}^{m\rightarrow r}
    = \cos\!\left(\mathbf{h}_{i,j}^{(m)}, \mathbf{h}_{i,k^*_{i,j}(m\rightarrow r)}^{(r)}\right), 
    \qquad
    \text{s.t.}  \quad
    k^*_{i,j}(m\rightarrow r)
    = \arg\max_{k}
      \cos\!\left(\mathbf{h}_{i,j}^{(m)}, \mathbf{h}_{i,k}^{(r)}\right).
\label{eq:tma}
\end{equation}
Higher TMA indicates stronger alignment between a modality token and the representations underlying the model's generation.
We can examine the TMA distribution at multiple granularities beyond the token level.
at the \emph{sample level}, $\mathcal{S}_i^m=\{s_{i,j}^{m\rightarrow r}\}_{j=1}^{N_m}$ characterizes how alignment is distributed within an input; and at the \emph{dataset level}, $\{\mathcal{S}_i^m\}_{i=1}^{N}$ captures variation across examples and modality conditions. 
Unlike aggregate alignment measures, TMA therefore preserves both within-sample and across-sample variation, providing a finer-grained view of how modality representations connect to generation.
The TMA distribution also offers a lens into whether the model retains relevant knowledge and expresses it effectively in its output.

\begin{figure*}[t!]
    \centering
    \begin{subfigure}{0.72\textwidth}
        \includegraphics[width=\textwidth]{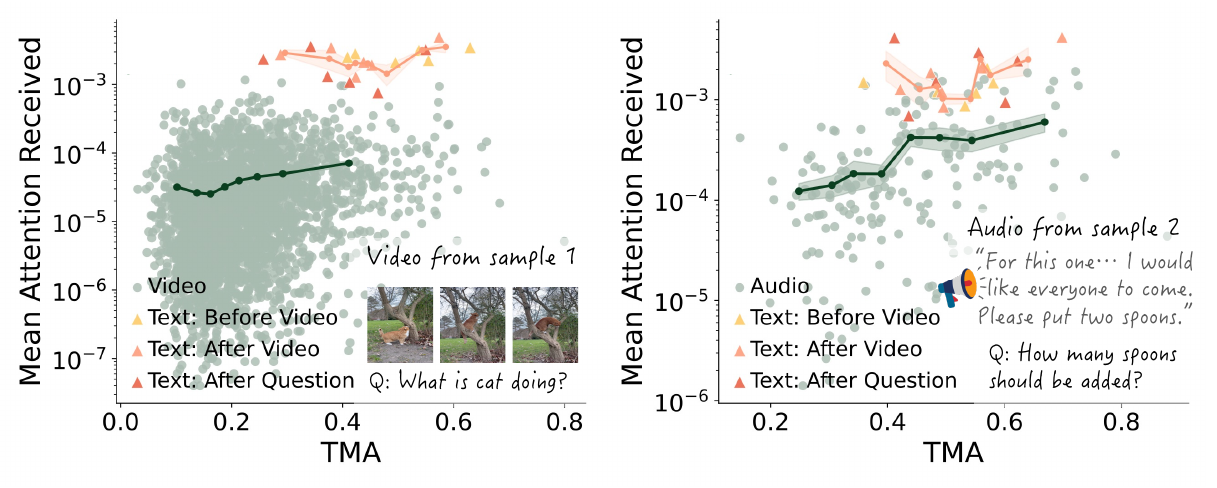}
        \vspace{-1.7em}
        \caption{Relation between TMA and attention received at the sample-level}\label{fig:tma_vs_attn}
    \end{subfigure}
    \hfill
    \begin{subfigure}{0.27\textwidth}
        \includegraphics[width=\textwidth]{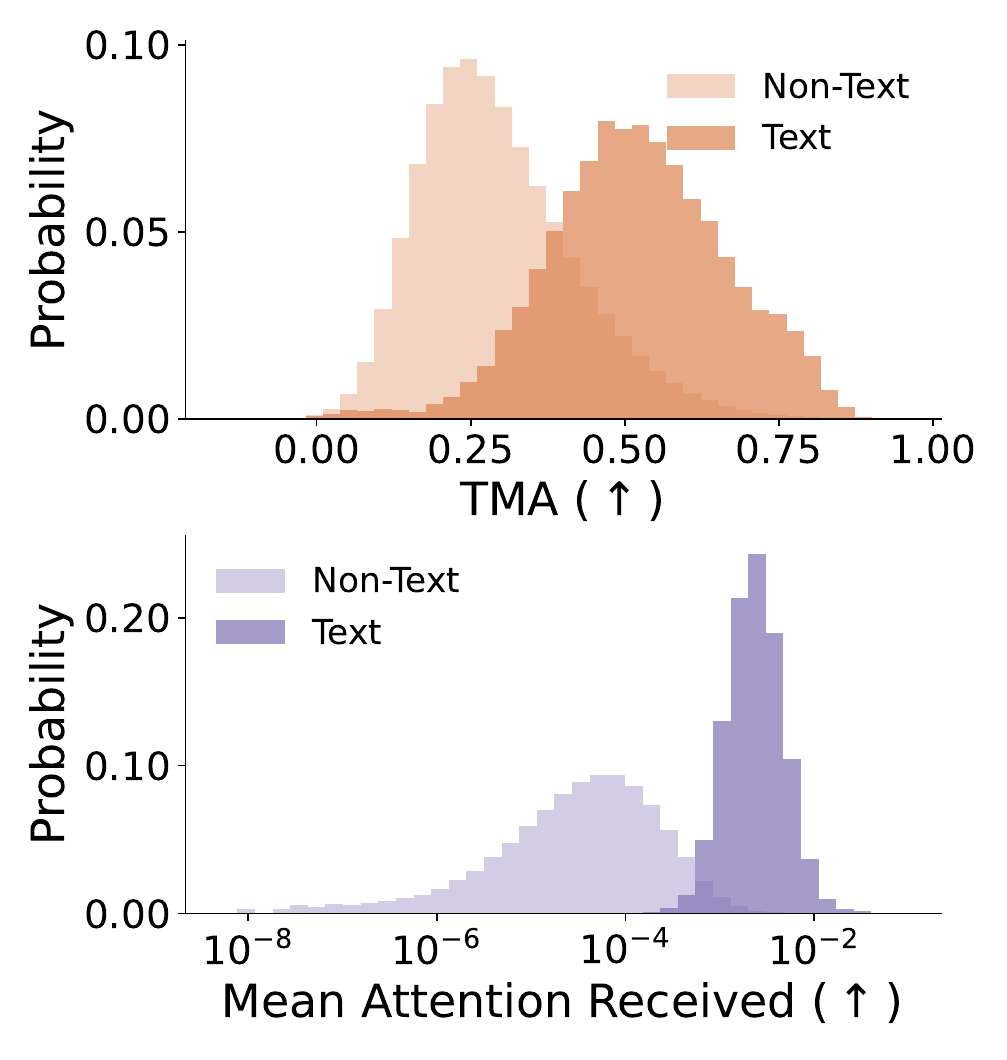}
        \vspace{-1.7em}
        \caption{Overall distribution}\label{fig:tma_and_attn}
    \end{subfigure}
    \vspace{-1.7em}
    \caption{\textbf{TMA distributions and accessibility to the generation pathway.}
    (a) Within samples, input modality (\eg, video and audio) tokens with higher TMA tend to receive higher attention weights from generated-token queries relatively, but exhibit low absolute TMA values.
    In contrast, text tokens yield higher attention weights and TMA values than non-text tokens, regardless of insertion position.
    (b) Across samples, text representations systematically exhibit higher TMA and attention than non-textual representations, revealing their greater accessibility to language-centric generation. 
    Note that TMA measures not the amount of information, but its alignment with generation.}
    \vspace{-2em}
\end{figure*}

\subsection{TMA Reflecting Limited Accessibility to Response Generation}
\noindent\textbf{TMA well captures how modality representations reach generation.} 
We use \emph{generation accessibility} to describe how an input representation participates in subsequent response generation, rather than whether it contains relevant information.
Since TMA measures the compatibility of an input representation with the response representation space, we examine its relationship with the generation-time attention weights from the same last decoder layer and greedy generation pass used for TMA.
For sample $i$, let $\mathcal{T}_i$ denote the set of decoding steps and $M_i^m$ denote the input-token positions of modality $m$.
The mean attention received by position $j\in M_i^m$ is
$
\bar a_{i,j}^{m}
=
\frac{1}{|\mathcal{T}_i|\,H}
\sum_{t\in\mathcal{T}_i}\sum_{h=1}^{H}
\operatorname{softmax}_{j}\!\left(
\frac{(\mathbf{q}_{i,t}^{h})^\top\mathbf{k}_{i,j}^{h}}{\sqrt{d_k}}
\right),
$
Across samples, higher TMA modality tokens tend to receive higher attention for both video and audio; however, the association is modest, and their absolute attention remains low (Figure~\ref{fig:tma_vs_attn}).
This distinction is crucial, as TMA and attention capture related but different aspects of generation accessibility.
TMA measures how closely an input representation aligns with representations underlying the generated response, whereas attention measures how much the decoder attends to that representation during generation.
Importantly, the two measures have different distributions: attention strongly separates text (inserted at different positions) from non-text tokens, while TMA distributions overlap (Figure~\ref{fig:tma_and_attn}).
Thus, non-textual representations can exhibit response-level alignment comparable to text representations even when they receive substantially less generation-time attention, a distinction that attention alone would obscure.
Conversely, text tokens, even when semantically unrelated to the queries and inserted at different positions (\ie, before modality 1, between modalities, or after modality 2), consistently receive substantially higher attention weights than non-textual tokens and also exhibit higher TMA overall (Figures~\ref{fig:tma_vs_attn} and~\ref{fig:tma_and_attn}).
Together, these patterns indicate that Omni-LLMs prefer textual representations during language-centric generation with higher generation accessibility. 

\begin{wrapfigure}{r}{0.6\linewidth}
    \centering
    \vspace{-1.5em}
    \begin{subfigure}{0.49\linewidth}
        \includegraphics[width=\textwidth]{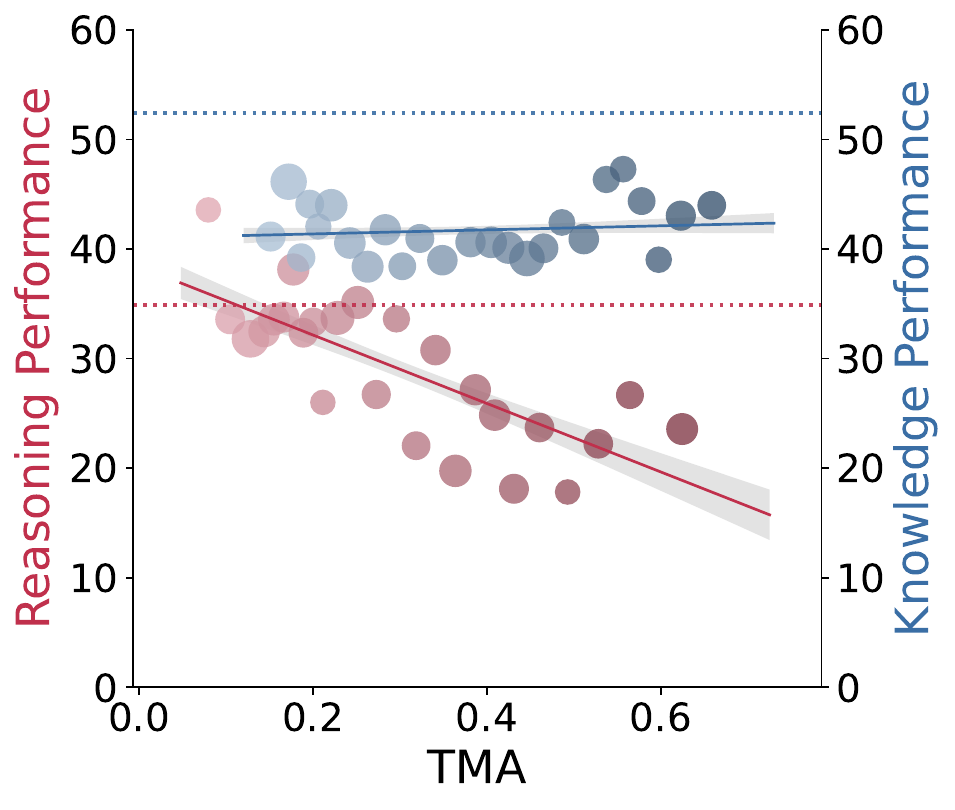}
        \vspace{-1.5em}
        \caption{TMA explanation}\label{fig:reasoning_vs_knowledge}
    \end{subfigure}
    \hfill
    \begin{subfigure}{0.49\linewidth}
        \includegraphics[width=\textwidth]{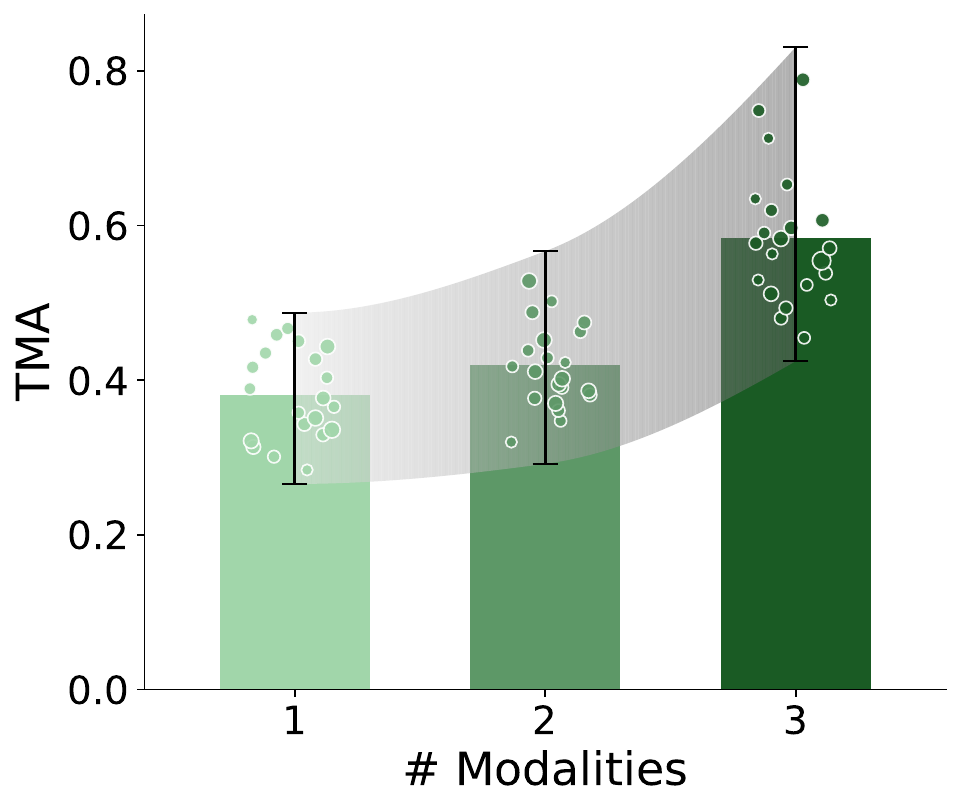}
        \vspace{-1.5em}
        \caption{Modality \# effects}\label{fig:modality_num_TMA}
    \end{subfigure}  
    \vspace{-0.5em}
    \caption{\textbf{TMA reveals a gap between what Omni-LLMs know and what they can express.}
    (a) Knowledge performance remains high regardless of the TMA of masked tokens, whereas answering performance is substantially lower and further degrades when higher-TMA tokens are masked.
    (b) Increasing the number of input modalities increases both the mean and variance of TMA, indicating more heterogeneous accessibility to response generation.
    }
    \vspace{-1em}
\end{wrapfigure}

\noindent\textbf{TMA distinguishes what the model \emph{knows} from what it \emph{says}.}
We then investigate whether low generation accessibility of non-textual modality tokens implies that the model lacks the underlying modality information.
To differentiate retained knowledge from its expression in downstream responses~\citep{li2026omnibench},
we evaluate both knowledge extraction and direct answering by masking tokens per sample. 
Masking tokens across TMA levels has little effect on knowledge performance, whereas overall downstream performance degrades substantially (Figure~\ref{fig:reasoning_vs_knowledge}), resulting in a noticeable gap between the two.
This contrast suggests that modality information can remain recoverable even when its corresponding tokens are weakly aligned with the representations underlying generation.
The TMA distribution therefore characterizes not whether the model holds information, but how strongly that information is aligned with its generation, suggesting a \emph{know-but-cannot-say} gap.

\noindent\textbf{When does the accessibility problem become worse?}
Finally, we examine how TMA changes as the number of input modalities increases.
Both its mean and variance rise with additional modalities (Figure~\ref{fig:modality_num_TMA}).
The higher mean indicates stronger average alignment with response representations, while the higher variance reveals greater differences among modality tokens.
Consequently, although some representations become more accessible as modalities accumulate, others remain relatively \emph{less accessible} to generation.
Thus, introducing new modalities does not uniformly translate additional modality information into generation;
instead, it can widen the variation in how readily that information is expressed,
contributing to a gap between what it knows and what it can express.

\section{Method: \ours{}}
\label{sec4:method}

\noindent\textbf{From TMA diagnosis to methodology design.}
Our analysis in \S\ref{sec3:analyses} showed that Omni-LLMs struggle to use retained modality information during generation, especially as the number of modalities increases.
TMA characterizes this accessibility at the token level, highlighting the importance of selectively accessing modality information relevant to the response.
Since text tokens align more closely with response representations, \ours{} complements sensory inputs with self-generated textual descriptions.
These descriptions express modality knowledge in the model's native generation space while preserving the original sensory representations, providing an additional pathway for relevant information to inform the response.

\noindent\textbf{Preliminary.} Conventional generative Omni-LLMs~\citep{qwen2025qwen25technicalreport,gemmateam2026} learn a function $f(\mathbf{X},q)\rightarrow \mathbf{y}$, where $\mathbf{X}=\{X_m\} _{m\in\mathcal{M}}$ denotes modality inputs, $q$ the query, and $\mathbf{y}=(y_1,\dots,y_T)$ the response.
Each modality-specific encoder maps its input to a latent representation, which is then processed by a shared language model:
$
    h_m = \mathcal{E}^m(X_m), \quad 
    \mathbf{y}=f_{\mathrm{LLM}}(\{h_m\}_{m\in\mathcal{M}},q),
$
where $\mathcal{E}^m$ and $h_m$ denote the encoder and latent representation for modality $m$.

\subsection{Self-Generating Modality Descriptions}
\label{method:training-free}
We introduce a simple training-free formulation of \ours{} that complements each non-textual representation with a self-generated modality description:
$
    d_m=f_{\mathrm{LLM}}(X_m;p_{\mathrm{desc}}), \quad
    \mathbf{y}=f_{\mathrm{LLM}}(\{h_m,d_m\}_{m\in\mathcal{M}},q;p_{\mathrm{ans}}),
$
where $p_{\mathrm{desc}}$ and $p_{\mathrm{ans}}$ denote the prompts for description and answer generation, respectively.
Both stages use the same underlying LLM.
Crucially, the generated description does not replace $h_m$; rather, it provides an additional textual pathway derived from the same modality information.
Thus, the description need not exhaustively reconstruct the original input, but should preserve task-relevant information in a form that is more directly compatible with the LLM's language-based processing.
From the perspective of TMA, this intervention exposes modality information through a representation that is more compatible with subsequent language generation, while preserving the original sensory representation.

Generating descriptions in a separate pass incurs additional inference cost.
We therefore also consider a single-pass procedure in which the modality descriptions and the final response are generated jointly:
$
(\mathbf{d},\mathbf{y})=f(\mathbf{X},q; p_{\mathrm{desc}} \circ p_{\mathrm{ans}}),
$
where $\mathbf{d}=\{d_m\}_{m\in\mathcal{M}}$, $\circ$ denotes prompt concatenation, and both $\mathbf{d}$ and $\mathbf{y}$ are generated conditioned on the same modality inputs and query.
Thus, the two variants of \ours{} only differ in whether modality descriptions are generated separately or jointly with the final response.
Most existing Omni-modal models employ LLMs as their foundation models and therefore naturally support text generation, which our method relies on. Models based on alternative foundation architectures~\citep{xie2025show,zhou2025transfusion,luo2026next} could likewise adapt our method by generating modality ``descriptions'' in a form compatible with their respective generation processes.

\subsection{Refining Modality Descriptions}
\label{method:training}

Self-generated modality descriptions provide additional textual pathways for modality information, but their usefulness for downstream prediction is not guaranteed.
We therefore develop a general framework to improve both their generation and use.
Let $\mathbf{X}=\{X_m\}_{m\in\mathcal{M}}$ denote the non-textual modality inputs, $q$ a multiple-choice question with candidate options, and $y$ the correct option.
Hereafter, we denote self-generated and ground-truth descriptions by $\hat{\mathbf{d}}=\{\hat d_m\}_{m\in\mathcal{M}}$ and $\mathbf{d}=\{d_m\}_{m\in\mathcal{M}}$.

Our framework considers two complementary objectives:
\emph{using} modality descriptions and \emph{generating} them.
The former trains the model to use self-generated descriptions for downstream prediction:
$
    \mathcal{L}_{\mathrm{U}}
    =-\log p_\theta\!\left(y \mid \mathbf{X},\hat{\mathbf{d}},q\right).
$
Here, $\hat{\mathbf{d}}$ is generated from the same input before the prediction step, following the two-pass training-free formulation of
\ours{} (see the Appendix for the single-pass variant).
The latter trains the model to generate modality descriptions from the raw inputs:
$
    \mathcal{L}_{\mathrm{G}}(\mathbf{z})
    =-\sum_{t=1}^{T_\mathbf{z}}\log p_\theta\!\left(z_t \mid z_{<t},\mathbf{X},q_{\mathrm{d}}\right),
$
where $\mathbf{z}=(z_1,\dots,z_{T_\mathbf{z}})$ denotes a target modality description and $q_{\mathrm{d}}$ specifies the description-generation instruction.

When ground-truth descriptions are available, we can directly supervise their generation:
$
\mathcal{L}_{\mathrm{P}}=\mathcal{L}_{\mathrm{G}}(\mathbf{d}).
$
When they are unavailable, we instead use self-generated descriptions as targets, $\mathbf{z}=\hat{\mathbf{d}}$ (concatenated over modalities), and apply a self-training or refinement strategy to select or improve these targets.
These objectives can be combined to first improve description generation and then refine their downstream use.
\begin{table*}[t]
\centering
\begin{minipage}[t]{0.58\linewidth}
\vspace{0pt}
\centering
\small
\setlength{\tabcolsep}{3pt}
\resizebox{\linewidth}{!}{%
\begin{tabular}{l|cccc|c}
\toprule
\multirow{2}{*}{\textbf{Method}} & \multicolumn{2}{c}{\textbf{AVHBench}} & \multirow{2}{*}{\textbf{OmniBench}} & \multirow{2}{*}{\textbf{WorldSense}} & \multirow{2}{*}{\textbf{\emph{Avg.}}} \\
 & \textbf{A-Hal} & \textbf{V-Hal} & & & \\
\midrule

\multicolumn{5}{l}{\textbf{\textit{Specialist Models}}} \\
EchoInk$^*$      & 72.4 & 68.3 & 46.8 & 39.3 & 56.7 \\
Omni-R1$^*$      & 75.8 & 65.8 & 47.6 & 39.7 & 57.2\\
HumanOmniV2$^*$  & 81.7 & 73.6 & 43.0 & 40.5 & 59.7 \\

\midrule
\multicolumn{5}{l}{\textbf{\textit{Training-free Models}}} \\
\texttt{Qwen2.5-Omni-3B}   & 77.8 & 62.2 & 43.8 & 39.1 & 55.7 \\
\ \ CoM~\citep{luo2026chain}       & 73.5\down{4.3} & \textbf{80.5}\up{18.3} & 43.8\down{0.0} & 40.7\up{1.6} & 59.6\up{3.9} \\
\rowcolor{gray!15} \ \ \ours{}  & \textbf{77.9}\up{0.1} & \textbf{80.5}\up{18.3} & \textbf{49.6}\up{5.8} & \textbf{44.6}\up{5.5} & \textbf{63.2}\up{7.5} \\

\texttt{Qwen2.5-Omni-7B}   & 73.2 & 66.8 & 46.5 & 39.2 & 56.4 \\
\ \ ThinkOmni~\citep{guanthinkomni2026} & 34.1\down{-39.1} & 46.2\down{-20.6} & 43.6\down{-2.9} & 33.4\down{-5.8} & 39.3\down{17.1} \\
\ \ CoM~\citep{luo2026chain} & \textbf{78.2}\up{5.0} & 79.7\up{12.9} & 49.4\up{2.9} & 41.9\up{2.7} & 62.3\up{5.9} \\
\rowcolor{gray!15} \ \ \ours{}  & 76.4\up{3.2} & \textbf{82.0}\up{15.2} & \textbf{51.7}\up{5.2} & \textbf{45.7}\up{6.5} & \textbf{64.0}\up{7.6} \\

\bottomrule
\end{tabular}%
}
\vspace{-0.9em}
\caption{\textbf{Comparison across different training-free methods.} \ours{} consistently outperforms corresponding baselines without an external reasoning model (ThinkOmni) and without an adaptive agentic framework (CoM).}
\vspace{-1.5em}
\label{tab:main_results}
\end{minipage}%
\hspace{0.02\linewidth}%
\begin{minipage}[t]{0.40\linewidth}
\vspace{0pt}
\centering
\small
\setlength{\tabcolsep}{3pt}
\resizebox{\linewidth}{!}{%
\begin{tabular}{l|ccc|c}
\toprule
\textbf{Method} & \textbf{Music} & \multicolumn{1}{c}{\begin{tabular}[c]{@{}c@{}}\textbf{Sound}\\ \textbf{Event}\end{tabular}} & \textbf{Speech} & \textbf{\emph{Avg.}} \\
 
\midrule
\texttt{gemma-4-e2b-it} & 25.5 & 16.2 & 22.4 & 21.4 \\
Self-CoT & 29.3\up{3.8} & 18.1\up{1.9} & 26.7\up{4.3}  & 24.7\up{3.3} \\
\rowcolor{gray!15}  \ours{}  & \textbf{43.4}\up{17.9} & \textbf{20.4}\up{4.2} & \textbf{34.0}\up{11.6} & \textbf{32.6}\up{11.2} \\

\midrule

\texttt{gemma-4-e4b-it} & 28.3 & 23.4 & 29.7 & 27.1 \\
Self-CoT & 34.0\up{5.7} & 25.7\up{2.3} & 34.5\up{4.8}  & 31.4\up{4.3} \\
\rowcolor{gray!15}  \ours{}  & \textbf{48.1}\up{19.8} & \textbf{31.3}\up{7.9} & \textbf{42.2}\up{12.5} & \textbf{40.5}\up{13.4}\\

\midrule
\texttt{Qwen2.5-Omni-3B} & 48.1 & 44.9 & 38.4 & 43.8 \\
Self-CoT & \textbf{51.9}\up{3.8} & 41.9\down{3.0} & 38.7\up{0.3} & 44.2\up{0.4} \\
\rowcolor{gray!15}  \ours{}  & 49.6\up{1.5} & \textbf{53.8}\up{8.9} & \textbf{48.3}\up{9.9} & \textbf{49.6}\up{5.8} \\

\midrule

\texttt{Qwen2.5-Omni-7B} & 49.1 & 51.0 & 39.6 & 46.6 \\
Self-CoT & \textbf{55.7}\up{6.6} & 48.7\down{2.3} & 40.0\up{0.4} & 48.1\up{1.5} \\
\rowcolor{gray!15}  \ours{}  & 50.9\up{1.8} & \textbf{51.7}\up{0.7} & \textbf{52.5}\up{12.9} & \textbf{51.7}\up{5.1} \\

\bottomrule
\end{tabular}%
}
\vspace{-0.8em}
\caption{\textbf{Comparison against Self-CoT.} \ours{} generally improves performance over the baselines across different audio types.}
\label{tab:ablation_audiotype_results}
\end{minipage}
\vspace{-1.5em}
\end{table*}
\section{Experimental Results}
\label{sec5}

This section examines whether the benefit of explicitly rendering modality-specific information into text extends across model families, tasks, and modality types.

\noindent\textbf{Setup.} We primarily apply \ours{} on \texttt{gemma-4-E2B/E4B-IT}~\citep{gemmateam2026}, \texttt{Qwen2.5-Omni-3B/7B}~\citep{xu2025qwen25omnitechnicalreport}, and \texttt{OneLLM}~\citep{han2024onellm}, across multimodal reasoning (OmniBench~\citep{li2026omnibench}; WorldSense~\citep{hong2025worldsense}), modality-conditioned reasoning (AVHBench~\citep{sung2025avhbench}), modality preference (OmniPreference~\citep{yan2026beyond}), and single-modality-to-text tasks (Appendix).
For training-based variants, we use 91k training samples from VALID~\citep{Huu2024VALID}.
We instantiate three existing self-training strategies--STaR~\citep{zelikman2022star}, R3V~\citep{cheng2025vision}, and SD-Zero~\citep{he2026self}--using modality descriptions (rather than reasoning traces) for self-improvement.
More details are provided in the Appendix.

\subsection{Generalization across Tasks}
\label{sec5:1}

\begin{wrapfigure}{r}{0.35\linewidth}
\centering
\vspace{-2.5em}
\includegraphics[width=\linewidth]{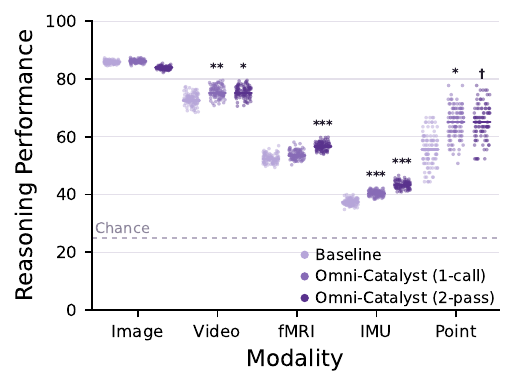}
\vspace{-2em}
\caption{\textbf{Modality extension.}
\ours{} is applicable to various modalities.
}
\label{fig:fmri_imu_point}
\vspace{-3.0em}
\end{wrapfigure}
Compared to prior competitive 
specialists~\citep{xing2025echoink,yang2025humanomniv2,zhong2026omni} 
and training-free methods~\citep{guanthinkomni2026,luo2026chain}, \ours{} (\S~\ref{method:training-free}) consistently attains superior performance across benchmarks (Table~\ref{tab:main_results}).
On OmniBench and WorldSense, where successful inference requires jointly using visual and audio information, CoM yields only 0--3\% gains, whereas \ours{} produces $+$5--7 improvements ($p<0.001$).
This advantage also holds across audio categories (Table~\ref{tab:ablation_audiotype_results}), especially consistently improving \texttt{gemma} performance by 8--18\%, approximately 2--14\% better than Self-CoT.

We further evaluate \ours{}'s generalizability across diverse modalities, including fMRI, IMU, and point clouds, in Figure~\ref{fig:fmri_imu_point}.
\ours{} significantly enhances performance ($\dagger$: $p<0.1$; *: $p<0.05$; **: $p<0.01$; ***: $p<0.001$), especially for the \emph{weaker} modalities showing relatively lower baseline performance, such as fMRI and IMU.
Representing modality-specific information in text provides a general mechanism for making heterogeneous modality information more accessible to language generation, without requiring external reasoning models~\citep{guanthinkomni2026} or complex inference configurations~\citep{luo2026chain}.

\subsection{Applicability to Different Training Methods}
\label{sec5:train}

\begin{wrapfigure}{r}{0.35\linewidth}
\centering
\vspace{-3em}
\includegraphics[width=\linewidth]{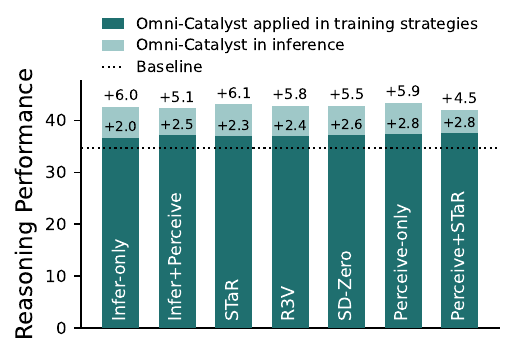}
\vspace{-2.5em}
\caption{\textbf{\ours{} adaptability.}
Applying \ours{} during both training and inference further improves Omni-LLM performance.
}
\label{fig:omnicatalyst_training}
\vspace{-1.5em}
\end{wrapfigure}

As shown in Figure~\ref{fig:omnicatalyst_training}, \ours{} can be internalized in the model weights across training strategies (\S~\ref{method:training}), improving over the untrained baseline by 2--3\%.
Notably, training the model to explicitly \emph{perceive} modality information ($\mathcal{L}_{\mathrm{P}}$) yields a larger gain (+2.8\%) than directly optimizing inference ($\mathcal{L}_{\mathrm{U}}$), even though the latter closely matches the test-time format.
Perception-only training also outperforms self-training methods (STaR, R3V, and SD-Zero).
Applying training-free \ours{} (\S~\ref{method:training-free}) on top of each approach yields an additional 5.1--6.1\% gain on average, bringing all variants to a comparable performance range of 42.4--43.6\%.
Finally, applying the overall best self-training method (STaR; +6.1\%) to the perceive-only-trained model does not further improve performance, although it produces higher-quality out-of-distribution audio captions~\citep{drossos2020clotho} (Appendix).
Overall, these results suggest that \ours{} complements rather than depends on any particular training strategy, while highlighting the value of improving modality perception before optimizing downstream inference.
\vspace{-1em}
\begin{wrapfigure}{r}{1\linewidth}
  \centering
  \vspace{-1.2em}
  \begin{subfigure}[b]{\linewidth}
  \vspace{0pt}
    \includegraphics[width=\linewidth]{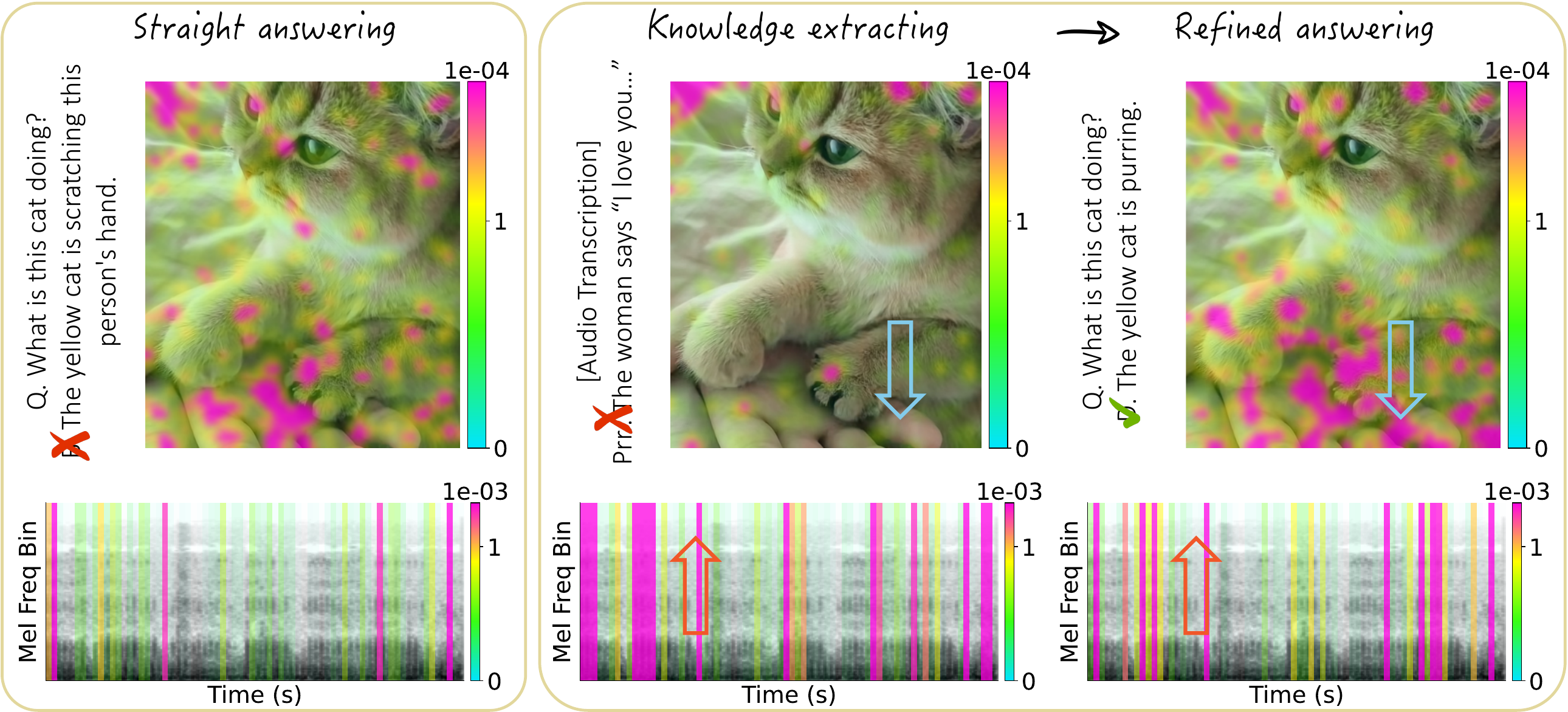}
    \vspace{-1.4em}
    \caption{Raw attention weights}
  \end{subfigure}
  \begin{subfigure}[b]{\linewidth}
  \vspace{0pt}
    \includegraphics[width=\linewidth]{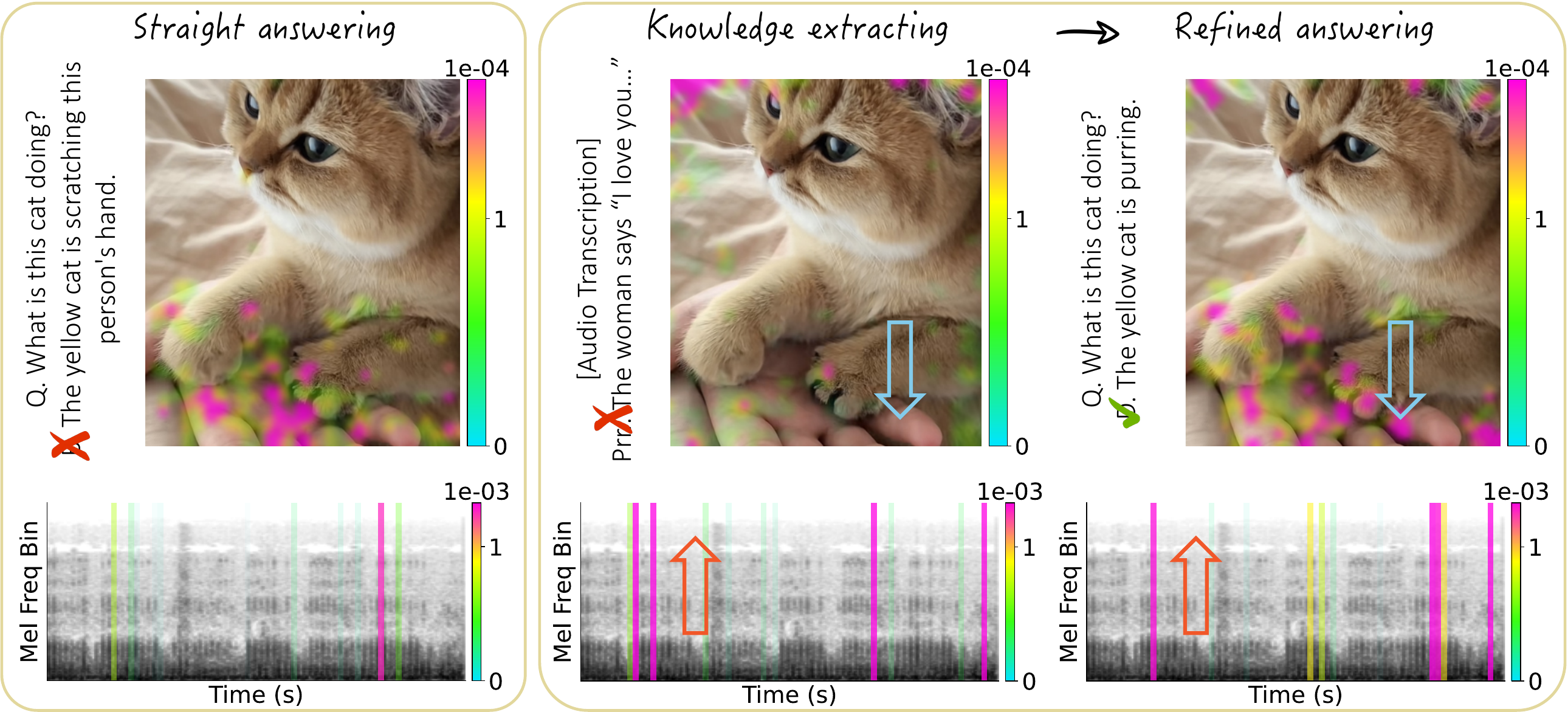}
    \vspace{-1.4em}
    \caption{Attention weights + TMA filter}
  \end{subfigure}
  \vspace{-1.8em}
  \caption{\textbf{Probe of how Omni-LLM attention changes.}
  The knowledge-extraction step in \ours{} enables the model to ``hear'' better, which in turn affects the refined answering stage, showing more balanced ``seeing'' and ``hearing.''
  Our TMA filter, applied on top of the raw attention weights on the original modalities, shows clearer visualizations of where the generated response aligns (\emph{compare} (a) and (b)).
  }
  \label{fig:attn_tma_qualitative}
  \vspace{-1em}
\end{wrapfigure}

\section{Empirical Analyses}
\label{sec6}
\subsection{How does \ours{} bridge the diagnosed modality–generation gap?}

\begin{figure}[t]
  \centering
  \vspace{-0.5em}
  \begin{minipage}[t]{0.65\linewidth}
    \vspace{0pt}
    \centering
    \begin{subfigure}[b]{0.49\linewidth}
      \includegraphics[width=\linewidth]{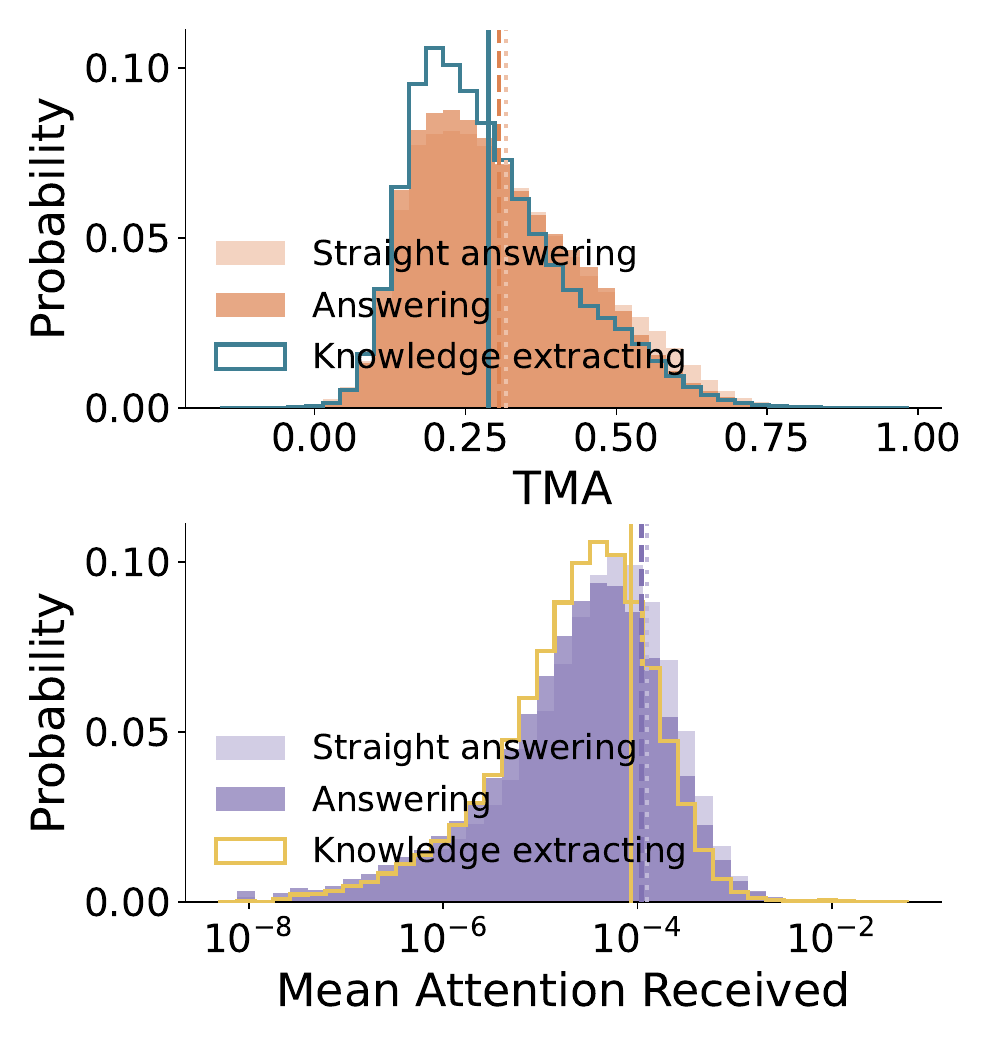}
      \vspace{-1.8em}
      \caption{Video distribution}
      \label{fig:tma_attn_video}
    \end{subfigure}\hfill
    \begin{subfigure}[b]{0.49\linewidth}
      \includegraphics[width=\linewidth]{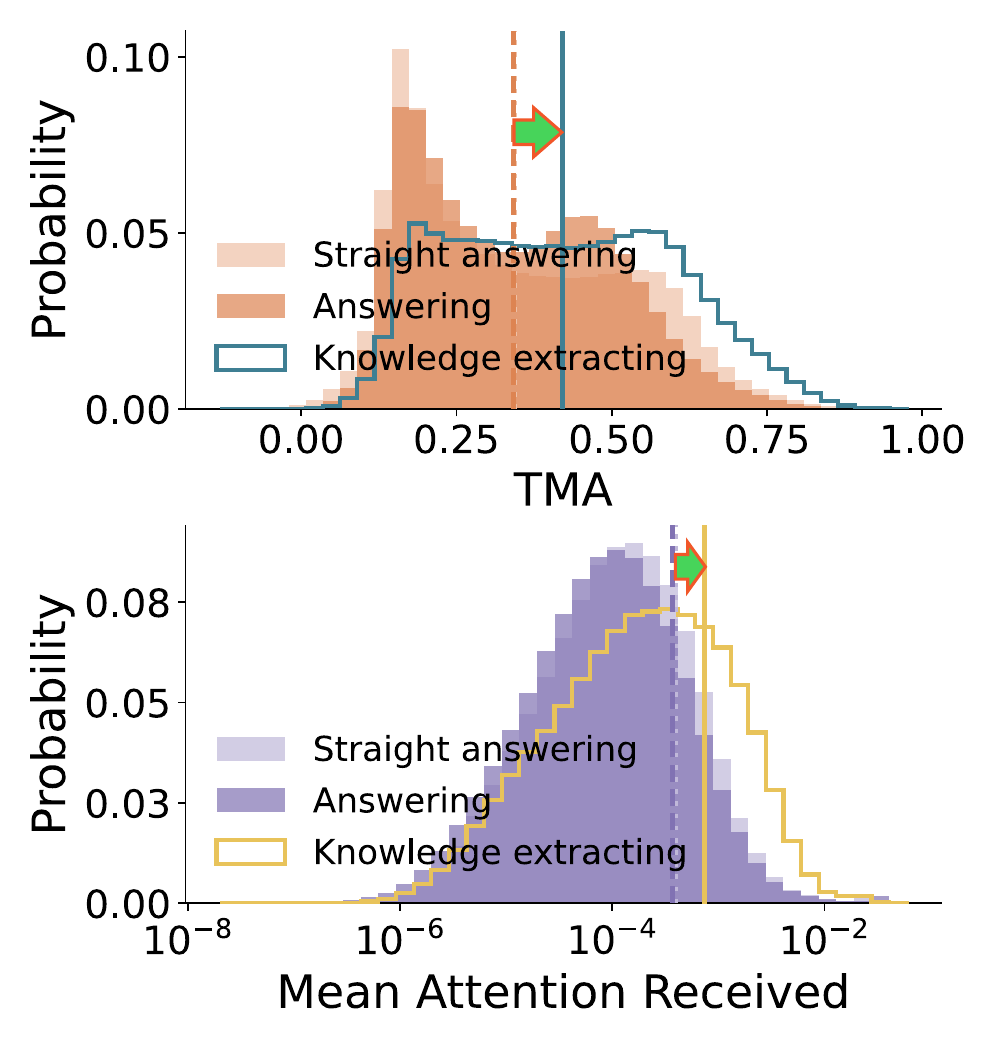}
      \vspace{-1.8em}
      \caption{Audio distribution}
      \label{fig:tma_attn_audio}
    \end{subfigure}
    \vspace{-0.8em}
    \caption{\textbf{TMA and attention distribution for each modality.}
    The knowledge extraction step of \ours{} selectively shifts TMA and attention for the weaker modality (here, audio).
    }
    \label{fig:tma_attn}
  \end{minipage}
  \hfill
  \begin{minipage}[t]{0.34\linewidth}
    \vspace{0pt} 
    \centering
    \includegraphics[width=\linewidth]{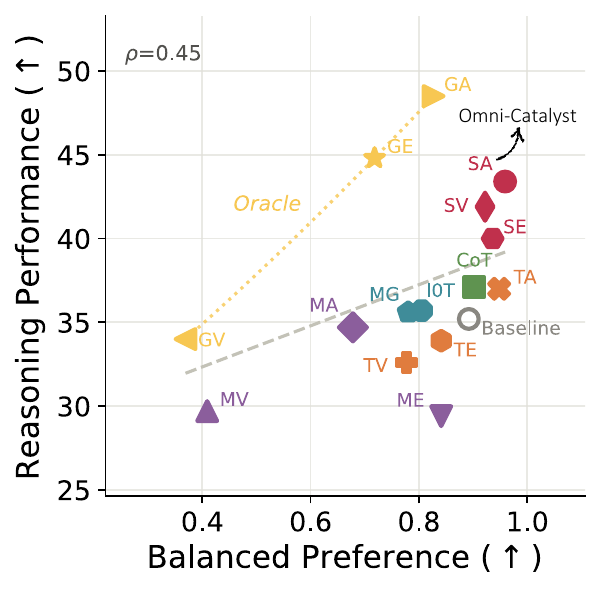}
    \vspace{-1.85em}
    \caption{\textbf{Performance and preference relation.}
    Reasoning performance correlates positively with modality preference balance.
    }
    \label{fig:preference_vs_perf}
  \end{minipage}%
  \vspace{-1em}
\end{figure}

We investigate how explicitly extracting modality information influences accessibility during generation.
Figure~\ref{fig:attn_tma_qualitative} reveals that direct answering often under-attends to the weaker modality (audio), whereas the knowledge-extraction step distributes attention more broadly over audio and converts its information into text.
The refined answering stage can then rely on this textualized audio information while focusing visual attention on the query-relevant region.
Importantly, TMA distinguishes where attention is merely allocated from where modality information aligns with the generated response: filtering attention by TMA ($s_j\!\ge\!\mathrm{median}(s)$) removes less relevant modality tokens and makes these response-aligned regions substantially clearer.
The aggregate distributions in Figure~\ref{fig:tma_attn} further show that knowledge extraction selectively increases both TMA and attention for the weaker modality.
Together, these results suggest that \ours{} improves access to underused modality information by textualization, while TMA characterizes how this information aligns with subsequent generation.

\subsection{Does bridging the gap improve reasoning and modality balance?}
\label{sec6:2}

We first examine whether making modality-specific information accessible during generation improves omni-modal reasoning and modality balance.
In Figure~\ref{fig:preference_vs_perf}, the pre-trained Omni-LLM (Baseline) achieves low accuracy (35\%) with an unbalanced modality preference, while existing modality-gap mitigation methods, MG~\citep{liang2022mind} and I0T~\citep{an-etal-2025-i0t}, leave both performance and modality preference largely unchanged.
This suggests that reducing representational disparity alone does not necessarily make heterogeneous modality information equally accessible during generation.
As an oracle, ground-truth audio descriptions (GA) move substantially toward higher performance and more balanced preference, yielding a 13\% performance gain.
\ours{} follows the same direction: its single-pass (SA, SV, SE) and two-pass (TA, TV, TE) variants generally improve both reasoning performance and preference balance.
The effect is particularly pronounced for audio descriptions (SA/TA), which shift modality preference toward a more balanced distribution (+10\%) while improving accuracy by up to 8\% (for SA).

These gains depend on preserving the original modality information rather than replacing it with text, since masking the original non-textual modalities with their descriptions (MA, MV, ME)~\citep{chen2025omnixr} degrades performance.
Meanwhile, Self-CoT, which gives the model a second chance to self-correct the initial answer, yields only marginal improvements in performance (+3\%) and preference balance (+1\%), suggesting that \ours{}'s gains are not simply due to additional inference.
Taken together, \ours{} helps Omni-LLMs better express information from heterogeneous modalities through language.
\section{Conclusion}
\label{sec6:conc}

Omni-LLMs can know more than they say.
We find that non-textual modalities can encode useful information that models fail to use effectively when generating answers.
TMA makes this gap visible at the token-level: non-textual representations are less aligned with generated response tokens and receive substantially less attention.
This suggests that modality imbalance is not necessarily a problem of missing knowledge, but of making that knowledge accessible to the language space used for answer generation. 
\ours{} addresses this gap with a simple intervention by self-generating textual descriptions of non-textual inputs. 
These descriptions increase modality alignment and make information already present in the inputs more available during answer generation. 
Across models and benchmarks, audio descriptions yield particularly consistent gains in both performance and modality balance.
We can automate the procedure in a learning process, where perceive-only training produces the strongest descriptions and performance among the strategies we evaluate.

\noindent\textbf{Limitations and future work.}
Our experiments cover only the modalities supported by current Omni-LLMs.
Nevertheless, TMA and \ours{} are modality-agnostic and can extend to emerging modalities such as tactile~\citep{tu2026unitac}, visual odometry~\citep{hilscher2026voca}, and olfactory~\citep{kim2026see}.
Our study also leaves the root causes of modality imbalance unresolved.
Possible sources include the LLM-centric architecture, multimodal pretraining data, and the difficulty of aligning heterogeneous representations.
More broadly, our findings suggest that progress toward omni-modal intelligence depends on how effectively models use information across modalities, beyond how many modalities they support.



\bibliography{iclr2027_conference}
\bibliographystyle{iclr2027_conference}

\clearpage
\newpage
\appendix

\renewcommand{\theHsection}{app.\arabic{section}}  
\renewcommand{\thefigure}{S\arabic{figure}}
\renewcommand{\thetable}{S\arabic{table}}
\renewcommand{\theequation}{S\arabic{equation}}
\setcounter{figure}{0}
\setcounter{table}{0}
\setcounter{equation}{0}

\section*{Appendix}

\section{AI Use Statement}\label{app:ai}

We used AI tools to edit and style the manuscript, identify relevant literature, check the bibliography for potential hallucinations or inaccuracies, assist with writing proofs in the supplementary materials, reconstruct the single-modality-to-text evaluation dataset, and edit software code.
The authors developed and verified all research ideas, conceptual development, methodological decisions, experimental design, analyses, and conclusions without AI assistance.
The authors reviewed and verified all AI-assisted text, proofs, and code and take full responsibility for the final content of this work.

\section{Ethics Statement}\label{app:ethic}

By simply encouraging Omni-LLMs to describe non-textual modality inputs, our approach improves the balance of modality usage and mitigates over-reliance on a single dominant modality.
We do not anticipate direct negative societal impacts beyond those inherent to the underlying pretrained models, whose biases our method neither introduces nor fully removes.

\section{Reproducibility Statement}\label{app:rep}

We provide sufficient details to reproduce our experiments, including model configurations, datasets, evaluation protocols, implementation details, and training settings. We describe all prompts, hyperparameters, and experimental procedures in the paper and supplementary materials. We use publicly available datasets and pretrained models, and provide the code and configuration files necessary to reproduce our main results. For trained variants, we additionally report the training data, objectives, and optimization settings. We will release the code and other necessary resources upon publication.

\clearpage
\newpage

\section{Implementation Details}\label{app:implementation}

\noindent\textbf{Datasets.}
We provide additional details on the datasets and evaluation settings used for the experiments that are omitted from the main paper.
For single-modality-to-text evaluation, we use
MS COCO~\citep{chen2015microsoft} for image-text,
MSVD QA~\citep{xu2017video} for video-text, 
Clotho~\citep{drossos2020clotho} for audio-text captioning,
Ego4D~\citep{grauman2022ego4d} for IMU-text, 
NSD~\citep{allen2022massive} for fMRI-text, 
Objaverse~\citep{deitke2023objaverse} for point-cloud-text tasks.

Because these datasets do not follow the standard Omni-LLM benchmark format, we reconstruct them into a common reasoning-oriented multiple-choice format. Table~\ref{supp_table:reconstructed_data} summarizes the resulting task construction, including the input modality, source annotation, and target category.
For each instance, we pool its ground-truth caption(s) or video QA text and use keyword-based scoring to map the source annotation to an abstract, modality-specific category (\eg, scene/activity, sound source, action, or object type).
The category is not required verbatim in the source text, so the mapping requires interpretation beyond surface-level vocabulary matching.
We select the highest-scoring category as the correct answer, sample three non-matching categories as distractors, and discard instances when we cannot obtain sufficiently distinct distractors.
Each resulting example therefore contains four text options, with correctness determined by exact option matching.

\begin{table*}[!ht]
\centering
\footnotesize
\setlength{\tabcolsep}{4pt}
\renewcommand{\arraystretch}{1.05}
\begin{tabularx}{\linewidth}{
  l|
  >{\raggedright\arraybackslash}X|
  >{\raggedright\arraybackslash}p{0.21\linewidth}
  >{\raggedright\arraybackslash}p{0.21\linewidth}|
  >{\centering\arraybackslash}p{0.10\linewidth}}
\toprule
\textbf{Modality} & \textbf{Ground-truth text (pooled)} & \multicolumn{2}{c|}{\textbf{Options}} & \textbf{Answer} \\
\midrule
\multirow{2}{*}{Image}
 & ``A child holding a flowered umbrella and petting a yak.''
 & A) A market, shop, or shopping scene & B) None of the above
 & \multirow{2}{=}{\centering C\newline{\scriptsize(\emph{animals\_\allowbreak pets})}} \\
 & ``A young man holding an umbrella next to a herd of cattle.'' \emph{(+3 more)}
 & C) An animal or pet is the main subject & D) People sharing a meal or eating food together & \\
\midrule
\multirow{2}{*}{Video}
 & ``who gives a wushu demonstration in front of a crowd?'' $\to$ man
 & A) An animal or pet is the main subject & B) A baby, infant, or child being cared for
 & \multirow{2}{=}{\centering C\newline{\scriptsize(\emph{dancing\_\allowbreak performing})}} \\
 & ``who demonstrated a sword routine for the crowd?'' $\to$ man \emph{(+34 more)}
 & C) Someone dancing or performing on stage & D) A person walking, standing, or sitting & \\
\midrule
\multirow{2}{*}{IMU}
 & \multirow{2}{=}{``stops painting the block of wood''}
 & A) Cooking or preparing food & B) None of the above
 & \multirow{2}{=}{\centering C\newline{\scriptsize(\emph{painting\_\allowbreak drawing\_\allowbreak writing})}} \\
 & & C) Painting, drawing, or writing & D) Picking up, placing, or putting down an object & \\
\midrule
\multirow{2}{*}{fMRI}
 & \multirow{2}{=}{``Three women are sitting at a table with pizza.''}
 & A) A market, shop, or shopping scene & B) None of the above
 & \multirow{2}{=}{\centering C\newline{\scriptsize(\emph{eating\_\allowbreak dining})}} \\
 & & C) People sharing a meal or eating food together & D) Someone preparing or cooking food & \\
\midrule
\multirow{2}{*}{\makecell[l]{Point\\cloud}}
 & \multirow{2}{=}{``a blue cartoon''}
 & A) An electronic device & B) None of the above
 & \multirow{2}{=}{\centering C\newline{\scriptsize(\emph{toy\_\allowbreak or\_\allowbreak character})}} \\
 & & C) A toy or cartoon/fictional character figure & D) A vehicle, such as a car, boat, train, or aircraft & \\
\bottomrule
\end{tabularx}
\vspace{-0.5em}
\caption{\textbf{Samples constructed for each single-modality domain.}
The pooled ground-truth caption(s) or QA text never state the correct category verbatim; the model must infer it and select the matching option.}
\label{supp_table:reconstructed_data}
\end{table*}

\noindent\textbf{Metrics.}
We use task-appropriate metrics for the three evaluation settings.
For the \emph{multimodal} benchmark, we report reasoning accuracy.
For the \emph{single modality-conditioned} benchmark, we report perception accuracy (correct yeses) and hallucination accuracy (correct nos) separately.
For the \emph{modality preference} benchmark (data along the x-axis in Figure~\ref{fig:alignment_metrics}), we quantify the balance of modality preference using the entropy of Normalized Modality Preference (NMP), where
$\mathrm{NMP}_m = \frac{P(m \mid I,A,T,C)}{P(m \mid C)}$.
Here, $I$, $A$, and $T$ denote image, audio, and text inputs, respectively, and $C$ denotes the set of modality options.
Higher NMP indicates stronger reliance on modality $m$ after normalizing for the model's textual prior, while higher entropy indicates a more distributed preference across modalities.
For audio captioning evaluation, we report standard automatic metrics, such as BLEU~\citep{papineni2002bleu} (averaged BLEU-1, BLEU-2, BLEU-3, and BLEU-4), METEOR~\citep{banerjee2005meteor}, ROUGE~\citep{lin2004rouge}, and CIDER~\citep{vedantam2015cider}.

\noindent\textbf{Model types and structures.}
To examine whether our findings extend beyond the models used in the main experiments, we additionally evaluate \ours{} on \texttt{Qwen3-Omni-30B-A3B-Instruct}~\citep{xu2025qwen3}, \texttt{OmniVinci}~\citep{ye2025omnivinci}, and \texttt{MiniCPM-o-2\_6}~\citep{yao2024minicpm}.
Table~\ref{supp_table:model_comparison} summarizes the architectures of all the tested Omni-LLMs, including their modality encoders and language-model backbones.
Most of the evaluated Omni-LLMs adopt a Qwen-family language model as the LLM backbone, whereas OneLLM and the Gemma family use different language-model backbones.

\begin{table*}[!ht]
\centering
\resizebox{\linewidth}{!}{%
\begin{tabular}{l|ccc}
\toprule
\textbf{Model} & \textbf{Audio Encoder} & \textbf{Visual Encoder} & \textbf{LLM} \\
\midrule
\texttt{OneLLM} & \multicolumn{2}{c}{CLIP ViT-L~\citep{radford2021learning} (\emph{shared})} & LLaMA2-7B~\citep{touvron2023llama} \\

\texttt{gemma-4-e2b-it} & USM-305M~\citep{zhang2023google} & ViT-150M~\citep{dosovitskiy2020image} & Gemma-4-2.3B~\citep{gemmateam2026} \\
\texttt{gemma-4-e4b-it} & USM-305M~\citep{zhang2023google} & ViT-150M~\citep{dosovitskiy2020image} & Gemma-4 4.5B~\citep{gemmateam2026} \\

\texttt{Qwen2.5-Omni-3B} & Whisper-large-v3~\citep{radford2023robust} & Qwen2.5-VL ViT~\citep{bai2025qwen25vltechnicalreport} & Qwen2.5-3B~\citep{qwen2025qwen25technicalreport} \\
\texttt{Qwen2.5-Omni-7B} & Whisper-large-v3~\citep{radford2023robust} & Qwen2.5-VL ViT~\citep{bai2025qwen25vltechnicalreport} & Qwen2.5-7B~\citep{qwen2025qwen25technicalreport} \\

\texttt{Qwen3-Omni-30B-A3B-Instruct} & AuT-650M~\citep{xu2025qwen3} & SigLIP2-So400M~\citep{tschannen2025siglip} & Qwen3-30B-A3B~\citep{yang2025qwen3technicalreport} \\

\texttt{omnivinci} & Audio Flamingo 3~\citep{ghosh2026audio} & SigLIP-So400M~\citep{zhai2023sigmoid} & Qwen2.5-7B~\citep{qwen2025qwen25technicalreport} \\

\texttt{HyperCLOVAX-SEED-Omni-8B} & Qwen2-Audio~\citep{chu2024qwen2} & Qwen2.5-VL ViT~\citep{bai2025qwen25vltechnicalreport} & HyperCLOVA~X SEED-8B~\citep{team2026hyperclova} \\

\texttt{MiniCPM-o-2\_6} & Whisper-medium~\citep{radford2023robust} & SigLIP-400M~\citep{zhai2023sigmoid} & Qwen2.5-7B~\citep{qwen2025qwen25technicalreport} \\
\bottomrule
\end{tabular}
}
\vspace{-0.3em}
\caption{\textbf{Omni-LLM architectures.} 
The Omni-LLM structures evaluated in this paper consist of modality-specific encoders and LLMs.
}
\label{supp_table:model_comparison}
\end{table*}

\noindent\textbf{Training details.}
For trained variants of \ours{}, we implement LoRA~\citep{hu2021lora} with rank 16 and $\alpha=32$ and AdamW~\citep{loshchilov2018decoupled} using weight decay $0.01$ and $\varepsilon=10^{-7}$.
We use a learning rate of $10^{-6}$ for Gemma and $3\times10^{-6}$ for Qwen.
STaR-based training~\citep{zelikman2022star} runs for four outer iterations with two epochs each and early stopping with patience 5 based on self-generated validation accuracy.
Descriptions are greedily decoded with up to 48 new tokens, and images are capped at 100,352 pixels.
We use seven VALID shards for training and two for validation, with at most four images and audio clips per sample and 2k MCQs per iteration.
We evaluate 100 validation examples every 200 optimizer steps and train on two A100-SXM-80GB GPUs.

\noindent\textbf{Self-training objectives.}
Our training variants use self-generated modality descriptions as supervision rather than relying on human-annotated reasoning traces.
STaR~\citep{zelikman2022star}, R$^3$V~\citep{cheng2025vision}, and SD-Zero~\citep{he2026self} differ in how they generate, refine, filter, and learn from self-generated supervision.
We adapt these strategies to generate descriptions of non-textual inputs, thereby providing a textual representation of modality-specific information that can be used during generation.
Commonly, the model's own binary reward
$r(\hat{\mathbf d})=\mathbb{1}[\arg\max_{y'}p_\theta(y'\mid\mathbf X,\hat{\mathbf d},q)=y]$
determines whether a generated description is useful for learning.
Each outer iteration is initialized from the original checkpoint, following~\citet{zelikman2022star}.

Distinctively, STaR~\citep{zelikman2022star} iteratively generates and filters rationales before fine-tuning on successful reasoning trajectories, while subsequent multimodal extensions such as
R$^3$V~\citep{cheng2025vision} and SD-Zero~\citep{he2026self} use self-generated supervision to improve reasoning without gold traces.
The detailed objectives for the three self-training strategies are given below.

\begin{itemize}
    \item \textbf{STaR}~\citep{zelikman2022star} retains greedily generated descriptions that yield a correct answer,
    $\mathcal{D}^{(t)}=\{(\mathbf{X},q,\hat{\mathbf{d}})\mid r(\hat{\mathbf{d}})=1\}$,
    optionally recovering failures by re-generating with the ground-truth caption as a hint:
    \begin{equation}
        \mathcal{L}_{\mathrm{STaR}}
        =\mathbb{E}_{\mathcal{D}^{(t)}}\!\left[\mathcal{L}_{\mathrm{U}}+\mathcal{L}_{\mathrm{G}}(\hat{\mathbf{d}})\right].
    \end{equation}

    \item \textbf{R3V}~\citep{cheng2025vision} samples $K$ descriptions per item and splits them into $\mathcal{D}^{+}$ and $\mathcal{D}^{-}$ by reward. The same model additionally acts as a refiner and a selector,
    where $\mathcal{C}=\{(d^{(k)},\hat{y}^{(k)})\}_{k=1}^{K}$:
    \begin{equation}
        \mathcal{L}_{\mathrm{R3V}}
        =\mathcal{L}_{\mathrm{STaR}}
        -\underset{\text{refiner}}{\underline{\log\pi^{\mathrm{Ref}}_\theta\!\left(d^{+}\mid d^{-}\right)}}
        -\underset{\text{selector}}{\underline{\log\pi^{\mathrm{Sel}}_\theta\!\left(y\mid\mathcal{C}\right)}}.
    \end{equation}

    \item \textbf{SD-Zero}~\citep{he2026self} uses the binary outcome reward $r(\cdot)$ not only to filter samples (like STaR) or pair samples for sequence-level SFT (like R3V) but as context for a self-reviser.
    One model acts as a generator $\pi^{\mathrm{Gen}}_\theta(\cdot)=p_\theta(\cdot\mid\mathbf{X},q_{\mathrm{G}})$
    and a reviser $\pi^{\mathrm{Rev}}_\theta(\cdot\mid\mathbf{d},P)=p_\theta(\cdot\mid\mathbf{X},q_{\mathrm{rev}},\mathbf{d},P)$.
    $r(\mathbf{d}_{\mathrm{init}})$ selects a control phrase $P_r$ (rephrase if correct, restart if incorrect), which is appended to the context before the model generates a revision $\mathbf{d}_{\mathrm{rev}}$; the resulting trace is kept only if $r(\mathbf{d}_{\mathrm{rev}})=1$,
    \begin{equation}
        \mathcal{L}_{\mathrm{SRT}}
        =\mathcal{L}_{\mathrm{U}}
        -\underset{\text{generator}}{\underline{\log\pi^{\mathrm{Gen}}_\theta\!\left(\mathbf{d}_{\mathrm{init}},P_r,\mathbf{d}_{\mathrm{rev}}\right)}}
        -\underset{\text{reviser}}{\underline{\log\pi^{\mathrm{Rev}}_\theta\!\left(\mathbf{d}_{\mathrm{rev}}\mid\mathbf{d}_{\mathrm{init}},P_r\right)}},
    \end{equation}
    after which the reviser, frozen at $\theta_{\mathrm{SRT}}$, serves as the teacher, and its token-level distribution is distilled into the generator on on-policy samples $\mathbf{d}\sim\pi^{\mathrm{Gen}}_\theta$:
    \begin{equation}
        \mathcal{L}_{\mathrm{SD}}
        =\mathbb{E}_{\mathbf{d}\sim\pi^{\mathrm{Gen}}_\theta}\biggl[
        \sum_{t=1}^{T_\mathbf{d}}
        \mathrm{KL}\bigl(
        \underset{\text{generator (student)}}{\underline{\pi^{\mathrm{Gen}}_\theta(\cdot\mid d_{<t})}}
        \,\|\,
        \underset{\text{frozen reviser (teacher)}}{\underline{\pi^{\mathrm{Rev}}_{\theta_{\mathrm{SRT}}}(\cdot\mid \mathbf{d},P_{r(\mathbf{d})},d_{<t})}}
        \bigr)\biggr]
        +\lambda\,\mathcal{L}_{\mathrm{U}}.
    \end{equation}
    
\end{itemize}

\begin{figure*}[!ht]
  \centering
    \includegraphics[width=\linewidth]{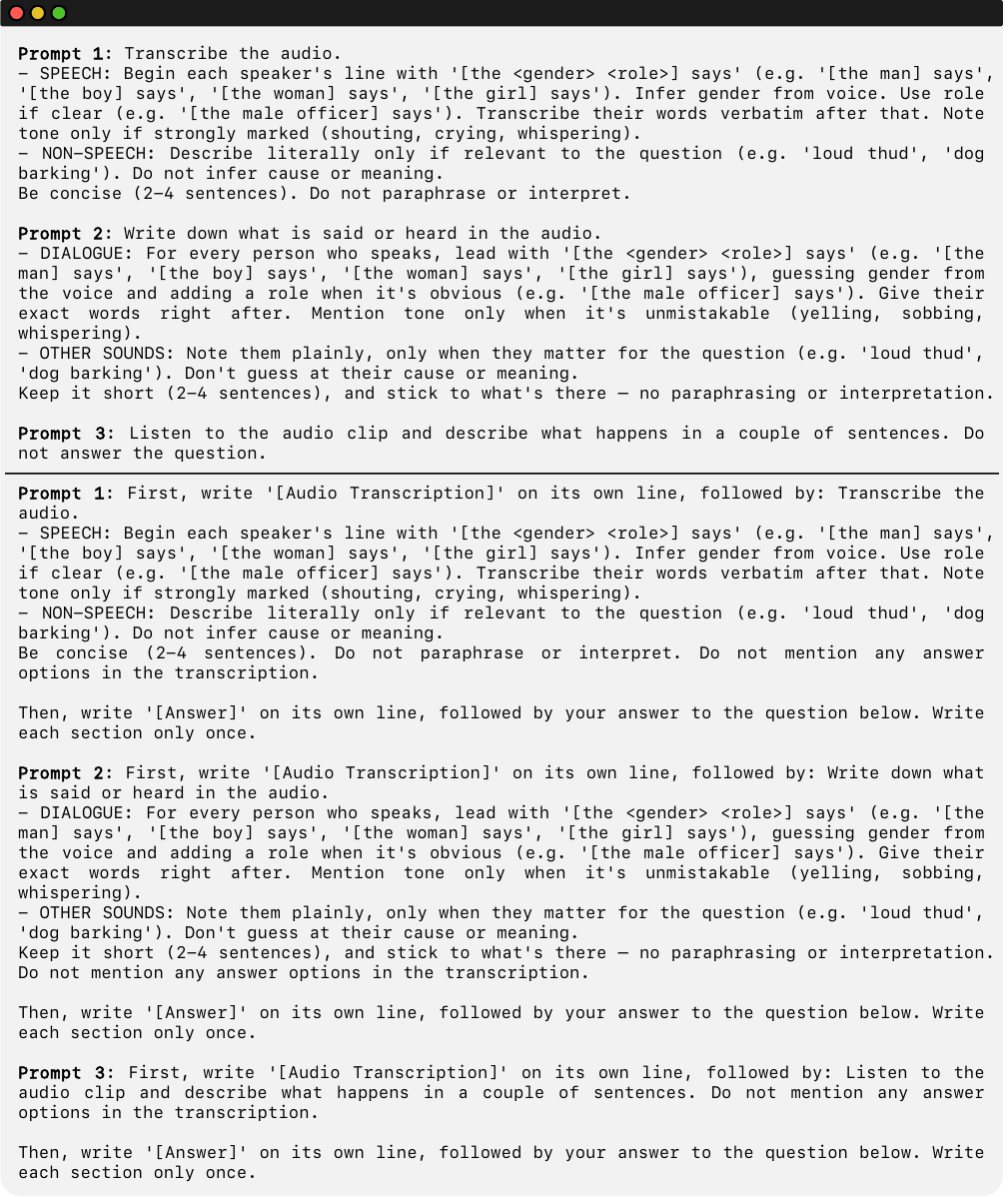}
    \vspace{-1.5em}
  \caption{\textbf{Prompt variations.}
  Alternative prompts for generating audio descriptions.
  Prompt 1 is used throughout the main experiments, which achieves performance within the range of the alternative prompts (Figure~\ref{supp_table:prompt_sensitivity}).}
  \label{supp_fig:prompts}
\end{figure*}

\noindent\textbf{Prompt details and sensitivity.}
The default prompt used to generate modality descriptions during training \ours{} is
(``\textit{Please listen carefully to the audio and observe the image. Describe what you see and hear.}'').
Figure~\ref{supp_fig:prompts} compares the default training-free \ours{} (Prompt 1) with alternative formulations that vary the wording and emphasis of the description instruction. As quantified in Table~\ref{supp_table:prompt_sensitivity}, \ours{} performance remains within a similar range across these prompt variants. Thus, the observed gains are not tied to a particular phrasing of the description instruction.

\begin{table*}[!ht]
\centering
\resizebox{0.6\linewidth}{!}{%
\begin{tabular}{l|cccc}
\toprule
\textbf{Model} & \textbf{Baseline} & \textbf{Prompt 1} & \textbf{Prompt 2} & \textbf{Prompt 3} \\
\midrule
\texttt{gemma-4-e2b-it} & 22.3 & 30.7\up{8.4} & 33.0\up{10.7} & 29.3\up{7.0} \\
\texttt{gemma-4-e4b-it} & 32.0 & 42.0\up{10.0} & 44.0\up{12.0} & 46.3\up{14.3} \\
\texttt{Qwen2.5-Omni-3B} & 39.3 & 41.7\up{2.4} & 42.3\up{3.0} & 40.0\up{0.7} \\
\texttt{Qwen2.5-Omni-7B} & 42.3 & 48.0\up{5.7} & 51.7\up{9.4} & 51.7\up{9.4} \\
\midrule
\emph{Avg.} & 34.0 & 40.6\up{6.6} & 42.8\up{8.8} & 41.8\up{7.9} \\
\bottomrule
\end{tabular}
}
\vspace{-0.5em}
\caption{\textbf{Prompt sensitivity.} 
Training-free \ours{} consistently improves over the baseline across prompt variants, with comparable gains across models, indicating limited sensitivity to prompt wording.
}
\label{supp_table:prompt_sensitivity}
\end{table*}

\begin{figure*}[!ht]
  \centering
  \includegraphics[width=0.5\linewidth]{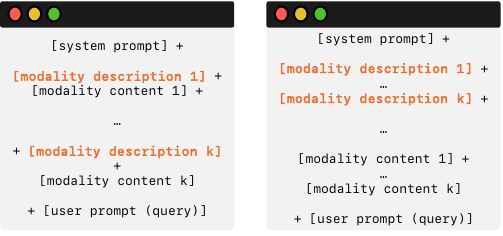}
  \vspace{-0.5em}
  \caption{\textbf{Effect of description position.}
  We compare two insertion positions for self-generated modality descriptions: before each corresponding modality (\emph{left}) and after the system prompt (\emph{right}).}
  \label{supp_fig:desc_position}
\end{figure*}

Figure~\ref{supp_fig:desc_position} illustrates the two description insertion positions considered in the prompt ablation: immediately before the corresponding modality and after the system prompt.
Table~\ref{supp_table:prompt_ablation} compares these alternative settings on both the multimodal (OmniBench~\citep{li2026omnibench}) and single modality-conditioned (CMM~\citep{leng2026curse}) benchmarks.
Complementing the original inputs with modality descriptions (TV, TA, TC, and TE) generally performs better than replacing the corresponding modalities with descriptions (MV, MA, and ME). On the modality-conditioned benchmark, conditionally complementing the description associated with the queried modality (TC) performs particularly well, suggesting that descriptions are most useful when they provide information relevant to the required modality in single-modality-conditioned tasks.

\begin{table*}[!ht]
\centering
\resizebox{\linewidth}{!}{%
\begin{tabular}{ll l cc ccc cccc ccc}
\toprule
\multirow{2}{*}{\textbf{Benchmark}} & \multirow{2}{*}{\textbf{Placement}} & \multirow{2}{*}{\textbf{Model}}
& \multicolumn{2}{c}{\textbf{Baseline}}
& \multicolumn{3}{c}{\textbf{Modality masking}}
& \multicolumn{4}{c}{\textbf{\ours{} (two-pass)}}
& \multicolumn{3}{c}{\textbf{Oracle description}} \\
\cmidrule(lr){4-5} \cmidrule(lr){6-8} \cmidrule(lr){9-12} \cmidrule(lr){13-15}
& & & \textbf{Orig} & \textbf{CoT} & \textbf{Visual} & \textbf{Audio} & \textbf{All} & \textbf{Visual} & \textbf{Audio} & \textbf{Cond.} & \textbf{All} & \textbf{Visual} & \textbf{Audio} & \textbf{All} \\
\midrule
\multirow{4}{*}{OmniBench}
& \multirow{2}{*}{Prepended}
 & Qwen  & 45.2 & 46.1 & 39.2 & 42.6 & 34.3 & 40.2 & 45.7$^{*}$   & 40.7 & 42.2 & 39.9 & \textbf{53.1}$^{***}$ & 50.7$^{***}$ \\
& & Gemma & 24.3 & 28.0$^{***}$ & 15.3 & 29.7$^{***}$ & 20.7 & 23.8 & 27.7$^{***}$ & 24.0 & 23.7 & 26.0 & \textbf{40.8}$^{***}$ & 38.3$^{***}$ \\
\cmidrule(lr){2-15}
& \multirow{2}{*}{System-level}
 & Qwen  & 45.2 & 46.2 & 38.8 & 40.3 & 35.6 & 41.0 & 46.9$^{*}$   & 41.0 & 42.9 & 40.9 & \textbf{54.0}$^{***}$ & 51.1$^{***}$ \\
& & Gemma & 24.3 & 28.0$^{***}$ & 20.5 & 29.0$^{***}$ & 23.2 & 24.1 & 27.0$^{***}$ & 24.0 & 24.8 & 27.2$^{*}$ & \textbf{42.9}$^{***}$ & 38.5$^{***}$ \\
\midrule
\multirow{4}{*}{CMM}
& \multirow{2}{*}{Prepended}
 & Qwen  & 84.0 & 75.0 & 79.0 & 80.8 & 71.7 & 83.8 & 83.8 & \textbf{84.2} & 81.0 & 83.8 & \textbf{84.2} & 80.4 \\
& & Gemma & 70.2 & 70.7 & 65.5 & 69.0 & 56.5 & 71.1 & 71.1 & \textbf{72.5} & 71.0 & 69.9 & 71.7 & 71.2 \\
\cmidrule(lr){2-15}
& \multirow{2}{*}{System-level}
 & Qwen  & 84.1 & 75.8 & 78.0 & 83.0 & 74.6 & 84.4 & 84.3 & \textbf{84.7} & 82.9 & 83.8 & 84.2 & 80.4 \\
& & Gemma & 72.0 & 70.3 & 67.0 & 71.9 & 60.6 & 71.8 & 71.5 & \textbf{72.4} & 71.8 & 69.9 & 71.7 & 71.2 \\
\bottomrule
\end{tabular}
}
\vspace{-0.5em}
\caption{\textbf{Effect of description position.}
Complementing the original modalities with descriptions generally outperforms replacing modalities with their descriptions. 
On the modality-conditional benchmark, conditionally complementing the query-relevant description (\textbf{Cond.}) performs best, while oracle audio descriptions provide the largest gains on OmniBench.
Results are consistent across both description placements (Figure~\ref{supp_fig:desc_position}).
}
\label{supp_table:prompt_ablation}
\end{table*}

\clearpage
\newpage

\section{Visualization Details}

\begin{wrapfigure}{r}{0.65\linewidth}
  \centering
  \vspace{-1.5em}
  \begin{subfigure}[t]{0.5\linewidth}
    \includegraphics[width=\linewidth]{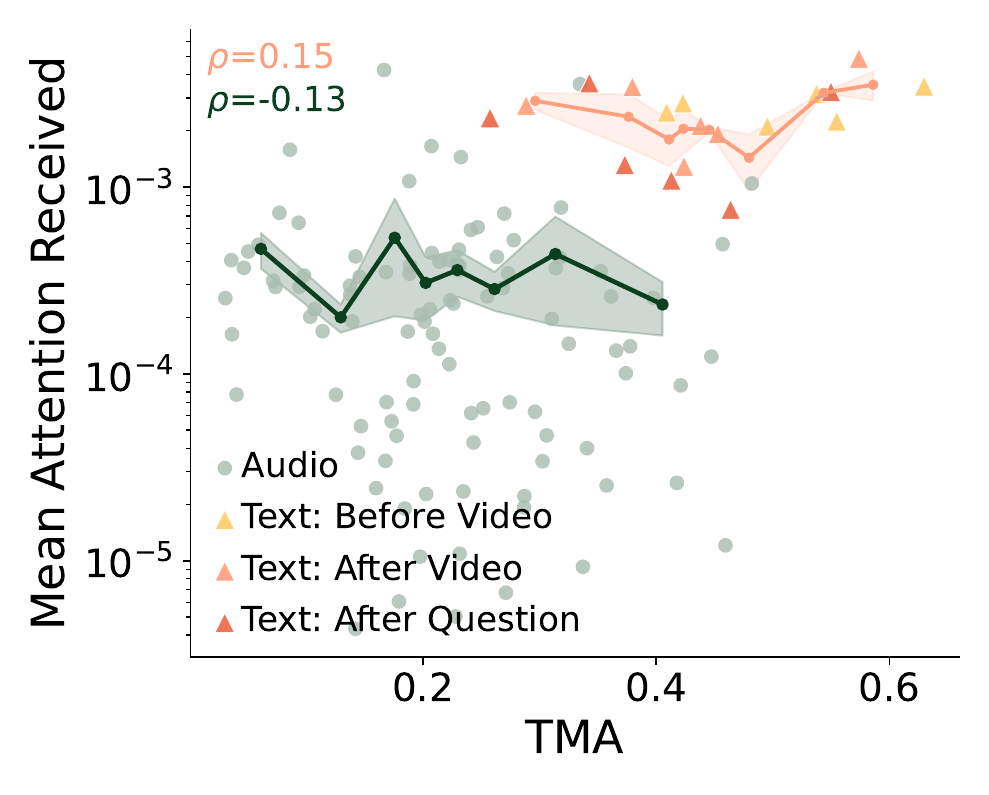}
    \vspace{-1.8em}
    \caption{Audio sample}
    \label{supp_fig:gap_vs_attention_1_gate25_audio}
  \end{subfigure}%
  \hspace{0pt}%
  \begin{subfigure}[t]{0.5\linewidth}
    \includegraphics[width=\linewidth]{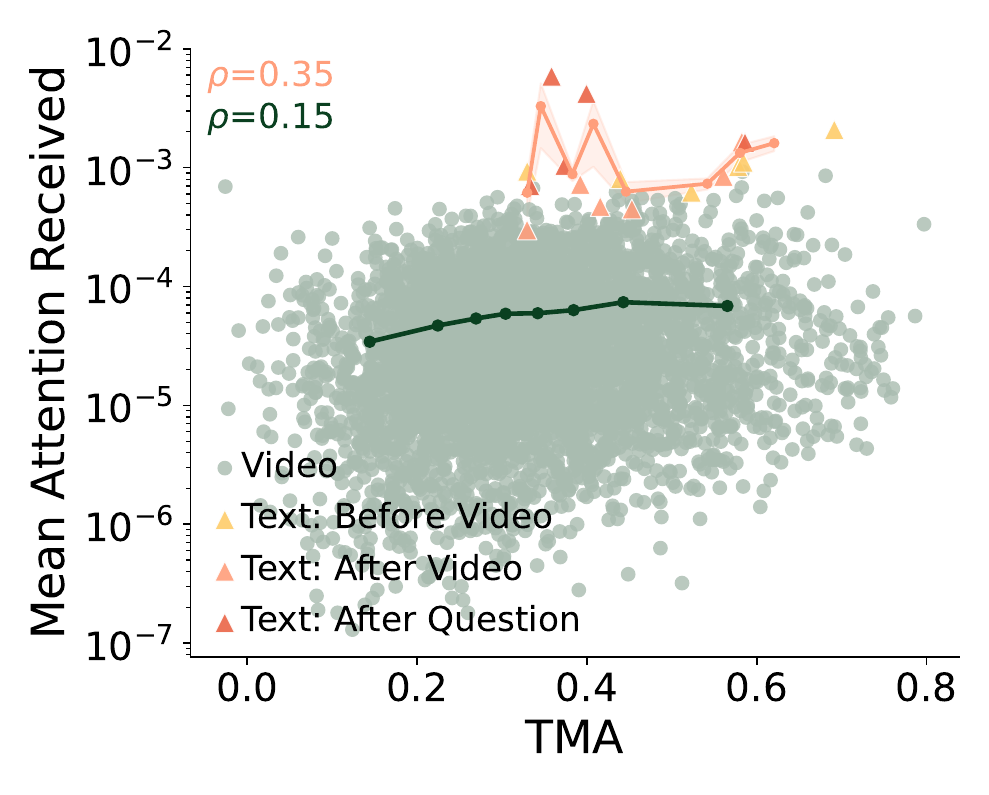}
    \vspace{-1.8em}
    \caption{Video sample}
    \label{supp_fig:gap_vs_attention_2_gate25_video}
  \end{subfigure}%
  \vspace{-0.7em}
  \caption{\textbf{Relation between TMA and attention received at the sample level.}
  Within samples, input modality (e.g., audio and video) tokens with higher TMA tend to receive higher attention weights from generated-token queries relatively, but exhibit low absolute TMA values.
  In contrast, text tokens yield higher attention weights and TMA values than non-text tokens, regardless of insertion position.
  This figure corresponds to audio and video from the two samples in Figure~\ref{fig:tma_vs_attn}.
  \label{supp_fig:gap_vs_attention}
  }
  \vspace{-2.0em}
\end{wrapfigure}

\noindent\textbf{TMA and attention.}
We first provide additional evidence for the relationship between TMA and attention.
Figure~\ref{fig:tma_vs_attn} extends the analysis in Figure~\ref{supp_fig:gap_vs_attention} with additional audio and video examples. Higher-TMA tokens tend to receive more generation-time attention, but the association is modest and absolute attention remains low. TMA and attention thus capture related but distinct aspects of generation accessibility: alignment with response representations and attention allocated during decoding, respectively.

\begin{figure}[!ht]
  \centering
  \hspace{-3em}
  \begin{subfigure}[b]{0.33\linewidth}
  \vspace{0pt}
    \includegraphics[width=\linewidth]{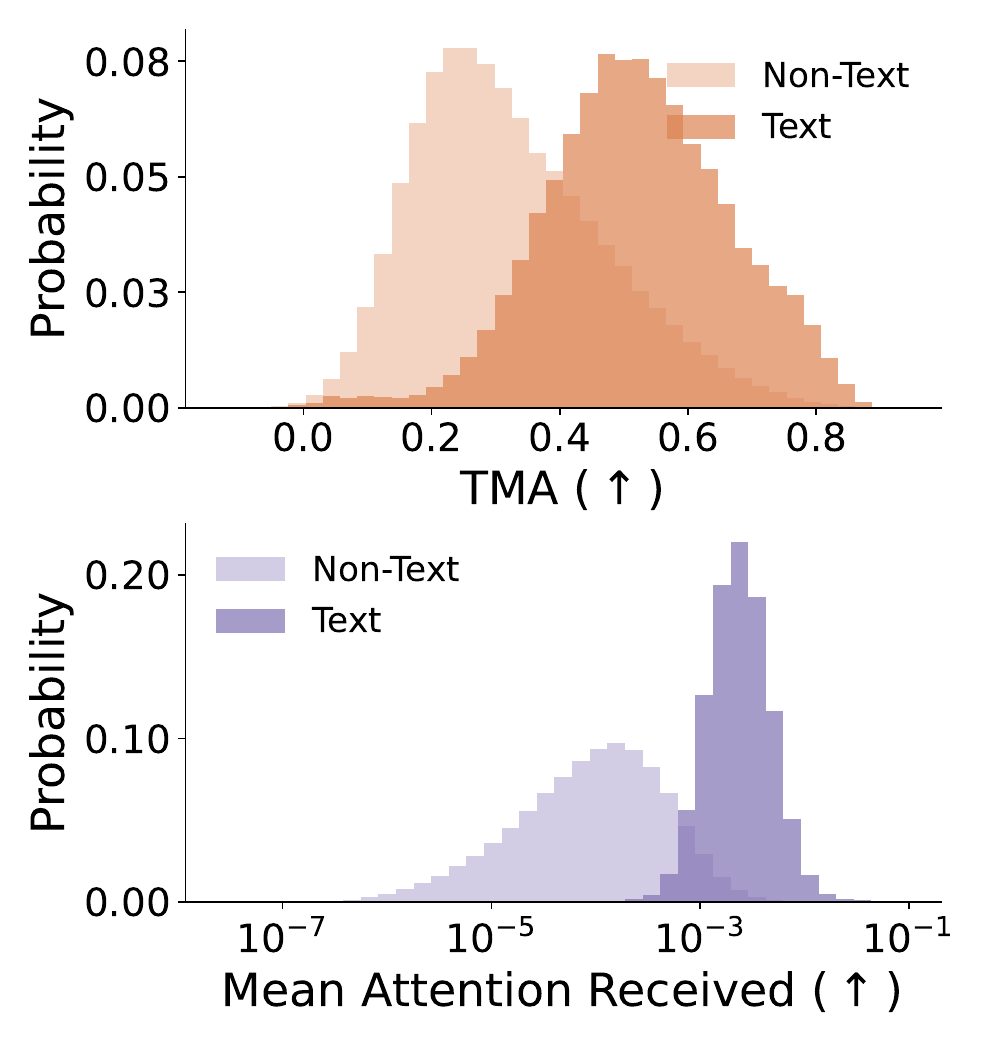}
    \vspace{-1.8em}
    \caption{Audio distribution}
    \label{supp_fig:gap_attn_distributions_audio}
  \end{subfigure}%
  \hspace{0pt}%
  \begin{subfigure}[b]{0.33\linewidth}
  \vspace{-3pt}
    \includegraphics[width=\linewidth]{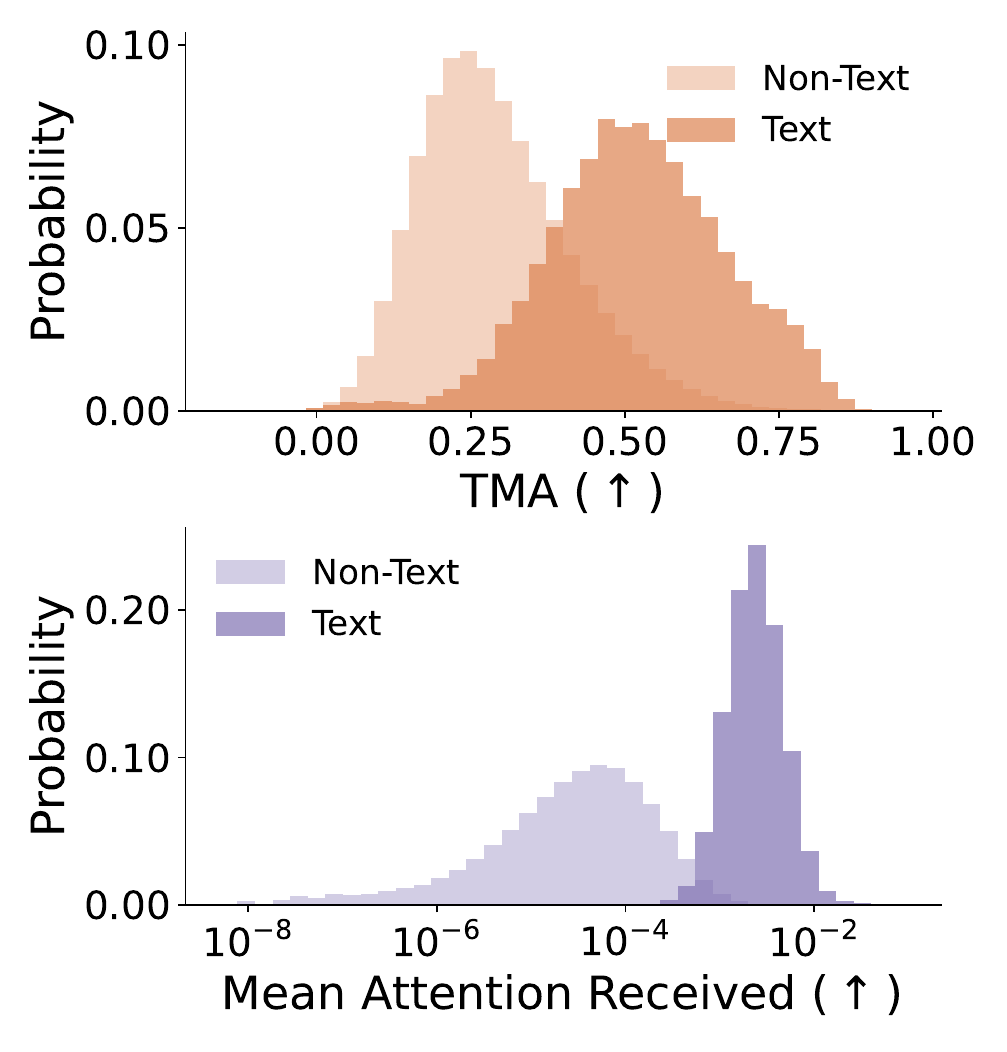}
    \vspace{-1.8em}
    \caption{Video distribution}
    \label{supp_fig:gap_attn_distributions_video}
  \end{subfigure}%
  \vspace{-0.7em}
  \caption{\textbf{TMA and attention distribution for each modality.}
  Text tokens exhibit higher TMA and attention than non-text tokens for both modalities, averaged in Figure~\ref{fig:tma_and_attn}.
  \label{supp_fig:gap_attn_distributions}
  }
  \vspace{-1em}
\end{figure}

\begin{figure}[!ht]
  \centering
  \hspace{-1em}
  \begin{subfigure}[b]{0.33\linewidth}
  \vspace{0pt}
    \includegraphics[width=\linewidth]{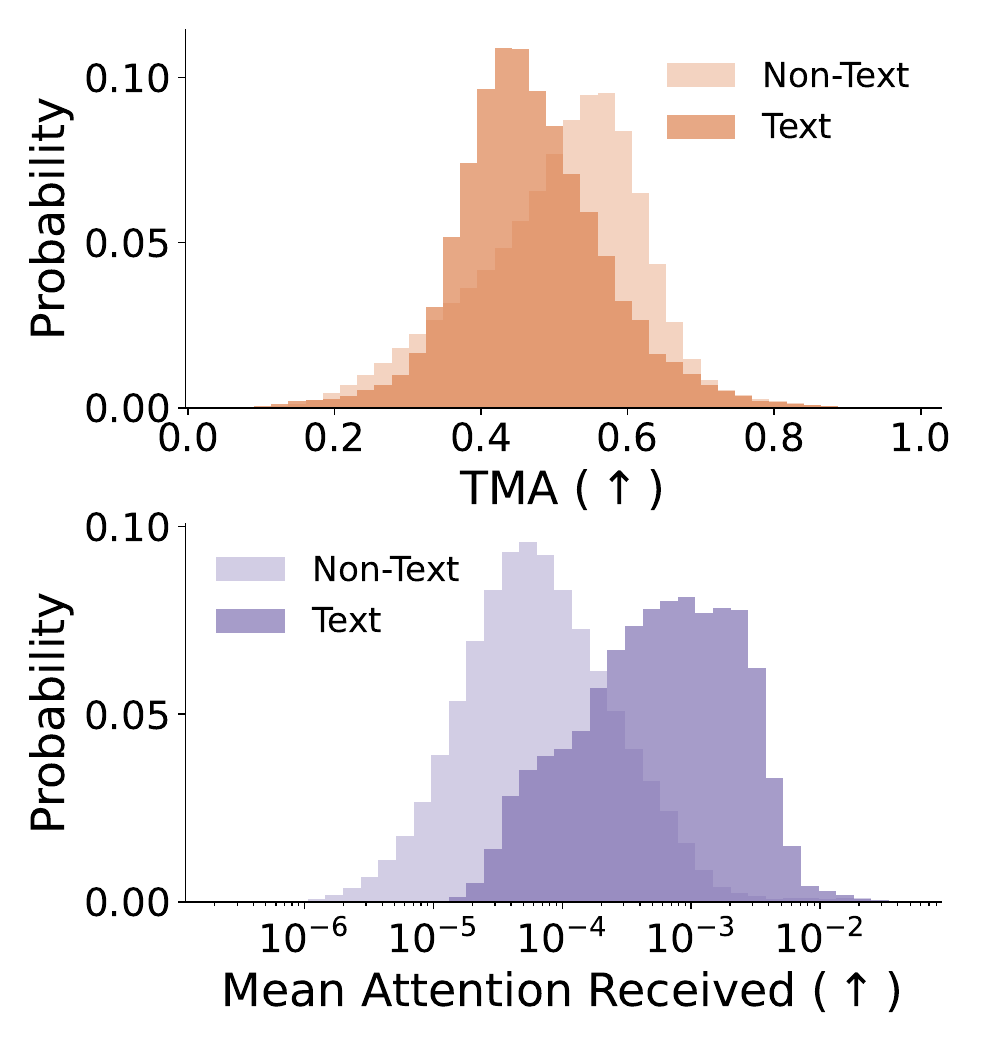}
    \vspace{-1.8em}
    \caption{\texttt{gemma-2b}}
    \label{supp_fig:gap_attn_distributions_gemma-4-E2B-it}
  \end{subfigure}%
  \hspace{0pt}%
  \begin{subfigure}[b]{0.33\linewidth}
  \vspace{-3pt}
    \includegraphics[width=\linewidth]{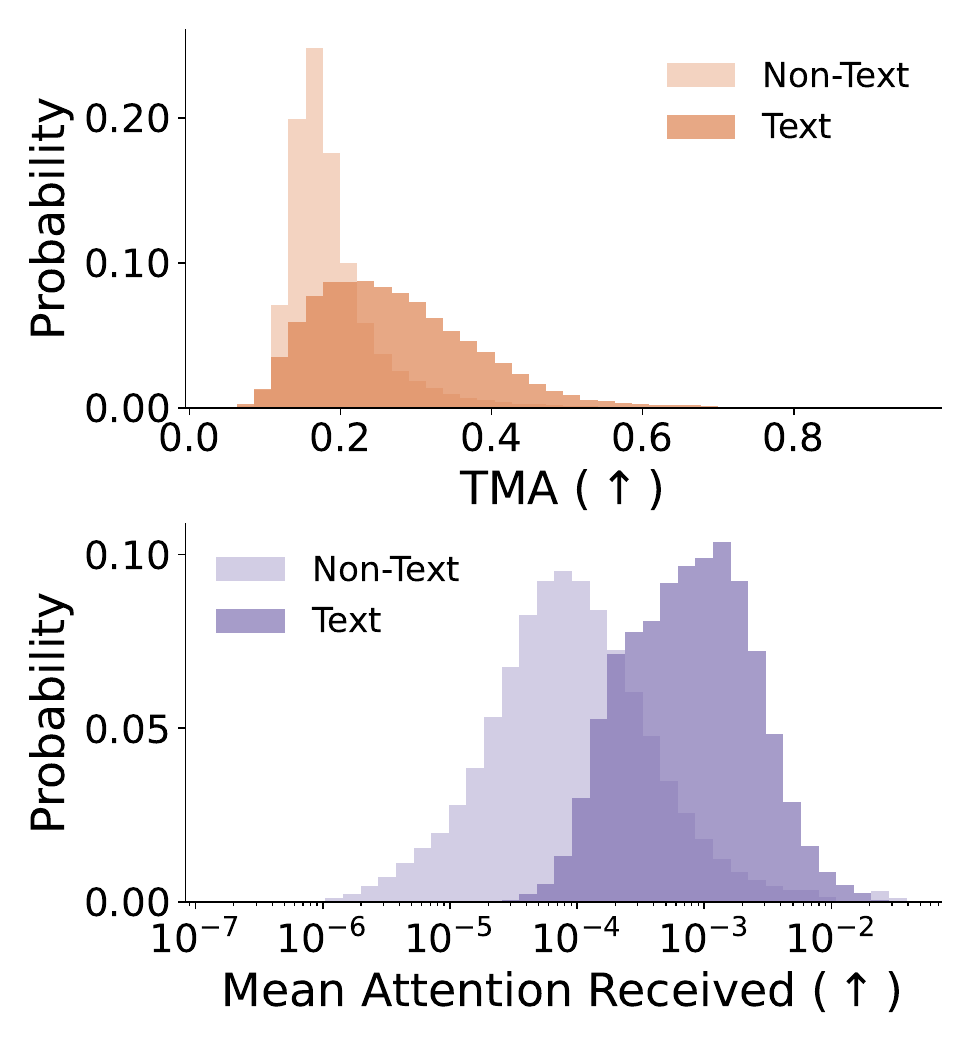}
    \vspace{-1.8em}
    \caption{\texttt{gemma-4b}}
    \label{supp_fig:gap_attn_distributions_gemma-4-E4B-it}
  \end{subfigure}%
  \begin{subfigure}[b]{0.33\linewidth}
  \vspace{-3pt}
    \includegraphics[width=\linewidth]{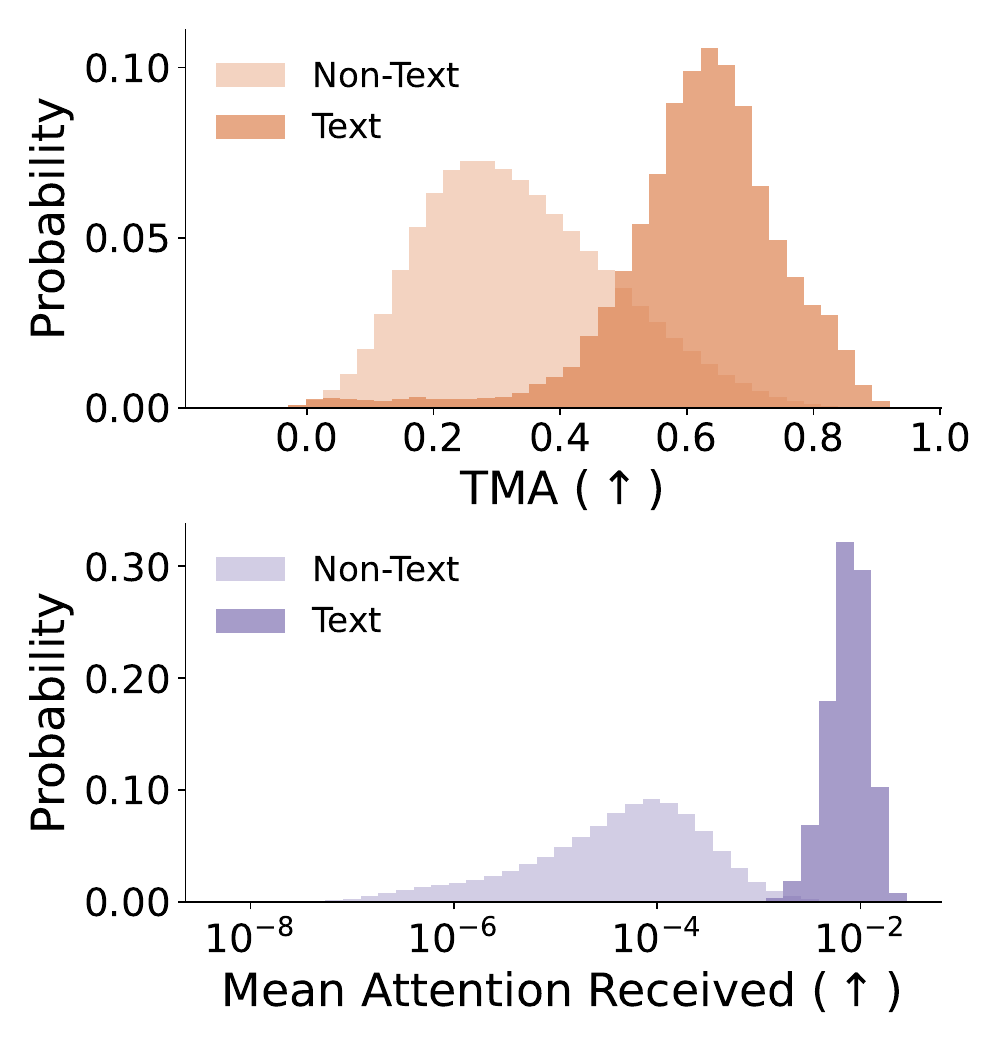}
    \vspace{-1.8em}
    \caption{\texttt{Qwen-3B}}
    \label{supp_fig:gap_attn_distributions_Qwen2.5-Omni-3B}
  \end{subfigure}%
  \vspace{-0.7em}
  \caption{\textbf{TMA and attention distribution for remaining Omni-LLMs.}
  Text tokens generally exhibit higher TMA and attention than non-text tokens in both modalities (\texttt{Qwen-7b} in Figure~\ref{fig:tma_and_attn}).
  \label{supp_fig:gap_attn_distributions_models}
  }
  \vspace{-4.0em}
\end{figure}

We further examine this relationship across token distributions and model families. Text tokens tend to exhibit higher TMA and attention than non-text tokens for both modalities.
Figures~\ref{supp_fig:gap_attn_distributions} and \ref{supp_fig:gap_attn_distributions_models} for each modality and different Omni-LLMs (corresponding to Figure~\ref{fig:tma_and_attn}).

Now, we provide conditions under which the empirical relationship hypothesized in \S~\ref{sec3:analyses} can arise.
Let $k^*(j)$ denote the response token maximizing the TMA of modality token $j$.
Suppose that the generated token representation contains a positively aligned component:
$
\mathbf h_{i,t}^{r}
=
\gamma_{i,t,j}\mathbf h_{i,k^*(j)}^{r}
+
\boldsymbol{\epsilon}_{i,t,j},
\gamma_{i,t,j}>0,
$
where the residual has a negligible contribution to its similarity with the modality token:
$
\boldsymbol{\epsilon}_{i,t,j}^{\top}\mathbf h_{i,j}^{m}\approx 0.
$
The attention logit from generated token $t$ to modality token $j$ is
$
\ell_{i,t,j}
=
\frac{
(W_Q\mathbf h_{i,t}^{r})^\top
(W_K\mathbf h_{i,j}^{m})
}{\sqrt{d_k}}
=
\frac{
(\mathbf h_{i,t}^{r})^\top
W_Q^\top W_K
\mathbf h_{i,j}^{m}
}{\sqrt{d_k}}.
$
As a special case, if the query-key projections preserve the ordering of similarities between the relevant response and modality representations, then increasing
$s_{i,j}^{m\rightarrow r}$
also increases the corresponding attention logit:
$
s_{i,j}^{m\rightarrow r}\uparrow
\quad\Longrightarrow\quad
\ell_{i,t,j}\uparrow.
$
Under the stated approximation,
$
\ell_{i,t,j}
\approx
(\mathbf h_{i,k^*(j)}^{(r)})^\top
\mathbf h_{i,j}^{m}
\approx
s_{i,j}^{m\rightarrow r}.
$
Since attention weights are obtained by softmax normalization,
$
\alpha_{i,t,j}
=
\operatorname{softmax}_{j}(\ell_{i,t,j}),
$
gives
$
s_{i,j}^{m\rightarrow r}\uparrow
\quad\Longrightarrow\quad
\alpha_{i,t,j}\uparrow
\quad\Longrightarrow\quad
\bar a_{i,j}^{m}\uparrow.
$
Thus, the observed positive association between TMA and attention follows under a specific representation and projection condition; this analysis does not imply that TMA and attention are equivalent.

\begin{figure*}[!h]
  \centering
    \includegraphics[width=\linewidth]{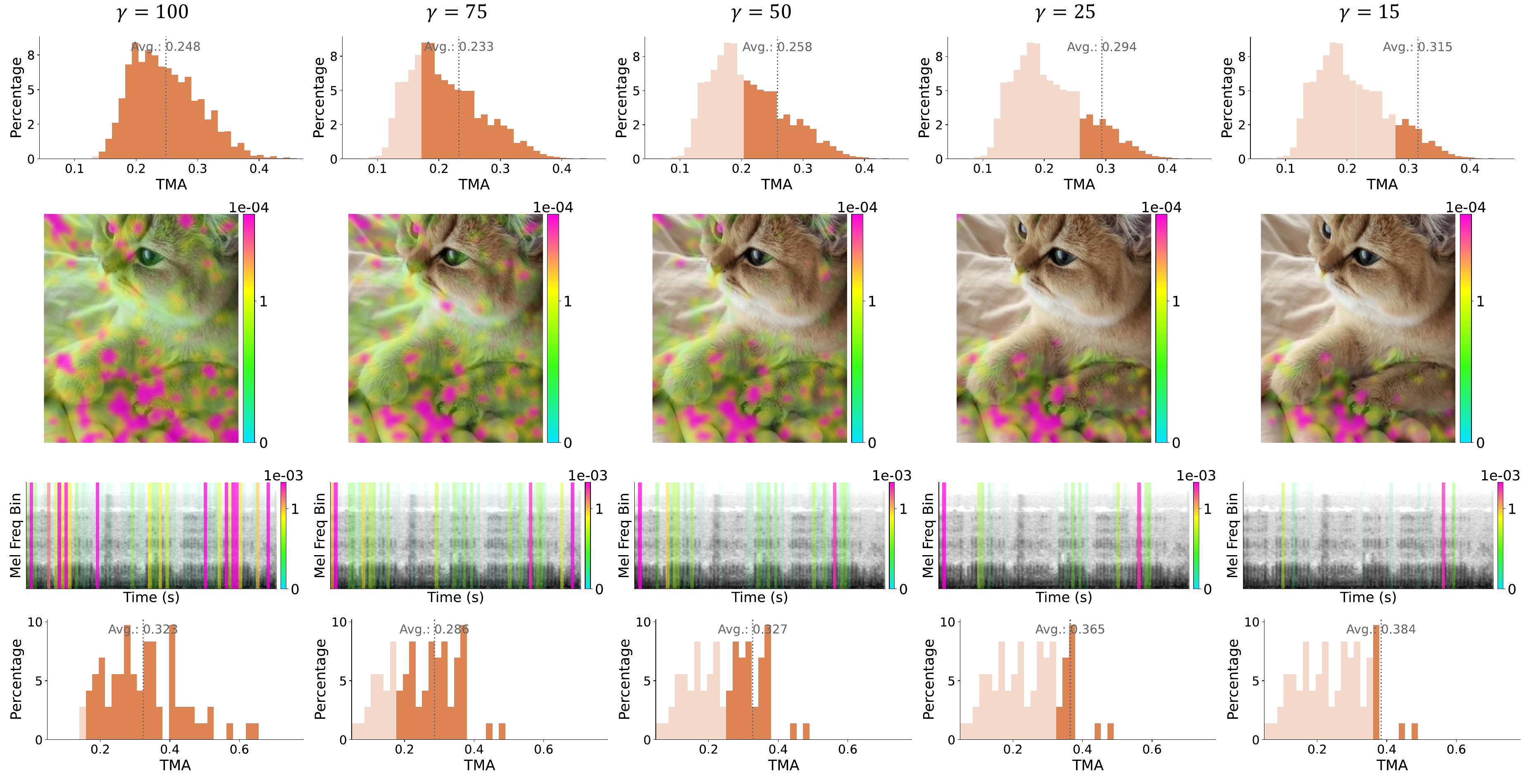}
  \vspace{-2em}
  \caption{\textbf{Probe of how Omni-LLM attention, capped with TMA, changes.} The highlighted blots remain in the prompt-relevant area (\eg, upper-left, near where the cat's paws are; less in the upper-left background regions) as $\gamma$ decreases.}
  \label{supp_fig:tma_variation}
  \vspace{-1em}
\end{figure*}

\begin{figure*}[!h]
  \centering
  \includegraphics[width=\linewidth]{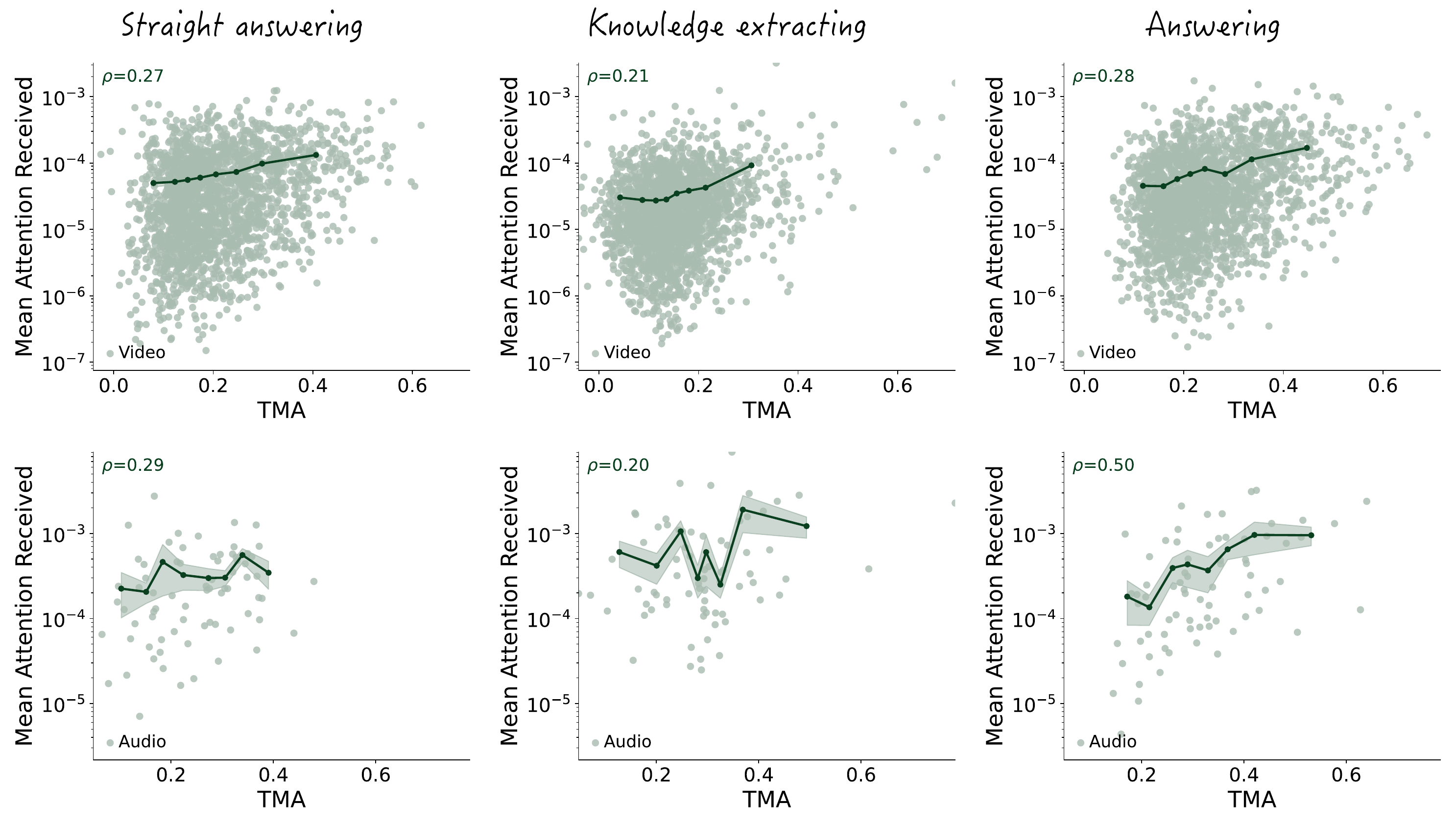}
  \vspace{-2em}
  \caption{\textbf{TMA reflects accessibility to the generation pathway.} Within samples, input modality tokens with higher TMA tend to receive greater attention from generated-token queries across steps.}
  \label{supp_fig:tma_vs_attention_kitten}
  \vspace{-2em}
\end{figure*}

\noindent\textbf{Attention map.}
To visualize the spatial and temporal distribution of attention within each non-textual modality, we project the token-level attention values back to their corresponding input positions. 
For each sample $i$ and modality $m$, we retain tokens in the upper $\gamma\%$ of the within-sample TMA distribution:
$
\mathcal{G}_i^m
=
\left\{ j\in M_i^m ;\middle|; \mathrm{TMA}_{i,j} \geq q_{\gamma}^{i,m} \right\}
$
where
$q_{\gamma}^{i,m}
=
\operatorname{Quantile}_{\gamma}
\left(
{\mathrm{TMA}_{i,k}:k\in M_i^m}
\right).$
We visualize only tokens in $\mathcal{G}_i^m$, leaving the remaining positions transparent.
Figure~\ref{supp_fig:tma_variation} visualizes how progressively restricting attention to higher-TMA tokens changes the resulting spatial attention map ($\gamma$ ranging from 100\% to 15\%). 
For video-only, we apply Gaussian smoothing to the attention, and TMA maps over the spatial patch grid before computing the normalization and selection threshold:
$
\tilde{x}_{i,j}
(G_{\sigma} * x_i)_j,
$
where $x_i$ denotes either the attention or TMA map ($\sigma=0.8$).
As the TMA threshold becomes more selective, the retained regions remain concentrated around prompt-relevant content, such as the upper-left region containing the cat's paws, rather than uniformly distributed over the input.
Figure~\ref{supp_fig:tma_vs_attention_kitten} additionally illustrates the relation between attention and TMA for each modality across stages.

\begin{wrapfigure}{r}{0.7\linewidth}
  \centering
  \vspace{-1.0em}
  \hspace{-1em}
  \begin{subfigure}[b]{\linewidth}
  \vspace{0pt}
    \includegraphics[width=\linewidth]{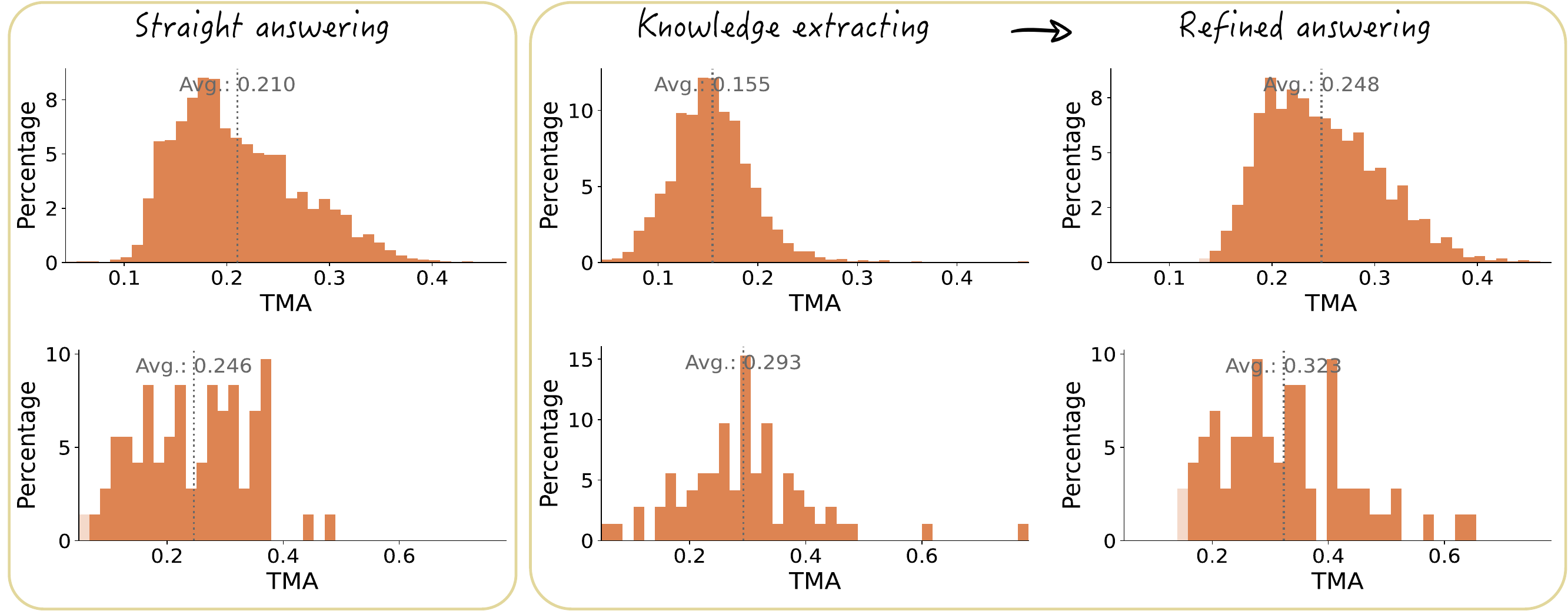}
    \vspace{-1.4em}
    \caption{Raw attention weights distribution for video (top) and audio (bottom)}
    \label{fig:attn_dist}
  \end{subfigure}
  \begin{subfigure}[b]{\linewidth}
  \vspace{1pt}
    \includegraphics[width=\linewidth]{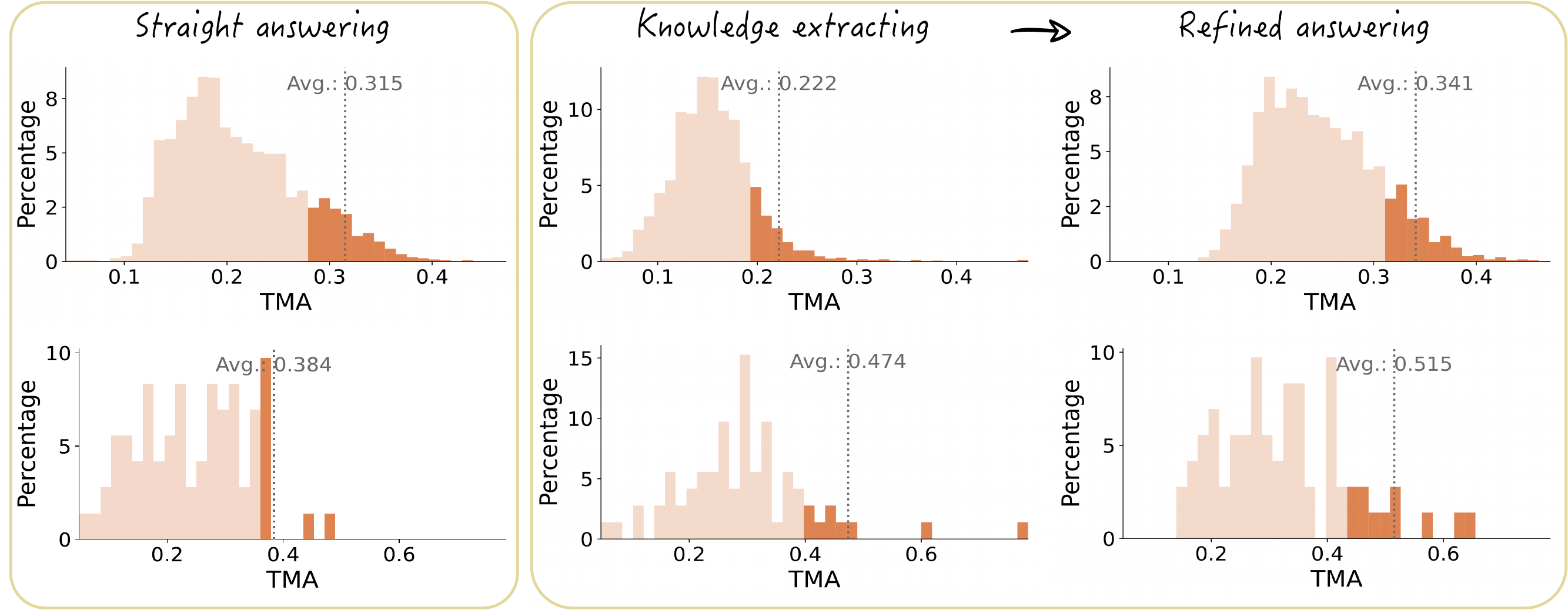}
    \vspace{-1.4em}
    \caption{Attention + TMA filter distribution for video (top) and audio (bottom)}
    \label{fig:attn_tma_dist}
  \end{subfigure}
  \vspace{-1.8em}
  \caption{\textbf{TMA and attention distribution for each modality.}
  The knowledge extraction stage shows higher (a) raw and (b) filtered attention weights for the audio modality, affecting the subsequent refined answering step.
  }
  \vspace{-1em}
\end{wrapfigure}

\noindent\textbf{Knowledge and reasoning performance.}
We next examine the distinction between perceptual knowledge and downstream reasoning under different modality conditions.
For each task, we compute reasoning performance directly from the benchmark: binary accuracy for AVHBench and multiple-choice accuracy for OmniBench.
We evaluate perceptual knowledge using the corresponding open-ended modality-description task.
An independent text-only judge extracts a small set of key content elements from the ground-truth description and checks whether each element appears in the model response verbatim or with semantic equivalence.
We report the fraction of examples with non-zero keyword recall.
For multimodal conditions, we evaluate only content associated with the modalities actually provided.
Figure \ref{fig:attn_dist} shows the raw attention distribution used during knowledge extraction, while ~\ref{fig:attn_tma_dist} shows the corresponding distribution after TMA filtering.
The two views illustrate how the extraction stage can preferentially attend to audio content before the subsequent answering stage.

\begin{figure*}[!ht]
  \centering
  \vspace{-1em}
  \begin{subfigure}[t]{0.31\linewidth}
    \centering
    \vspace{-45em}
    \includegraphics[width=\linewidth]{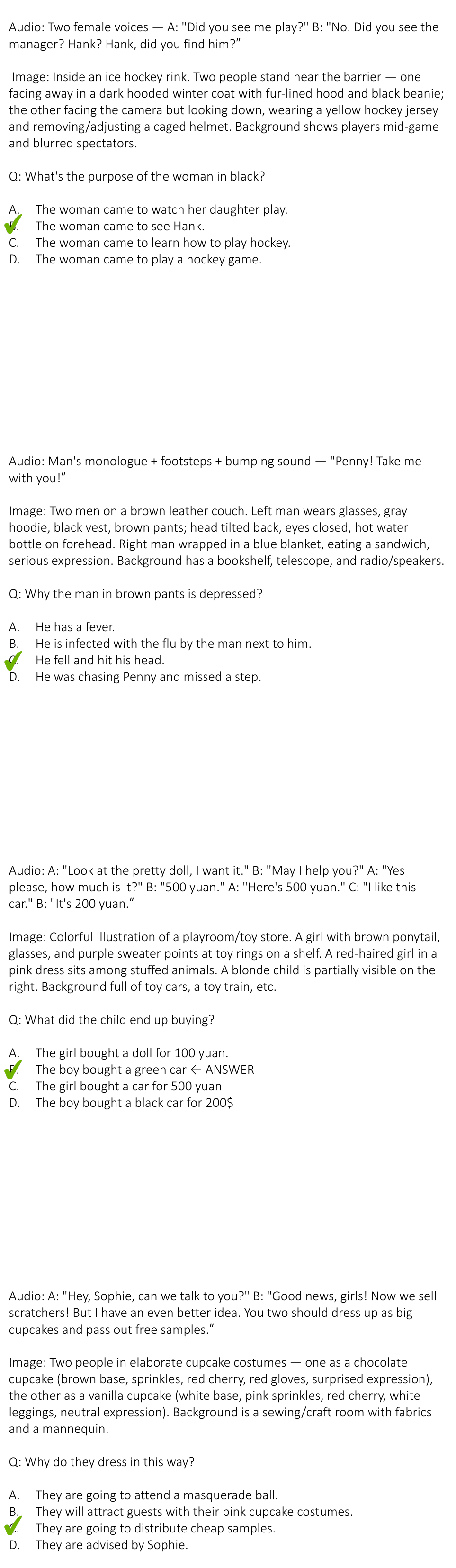}
    \vspace{1.6em}
    \caption{Benchmark question}
    \label{supp_fig:knowledge_reasoning_qwens}
  \end{subfigure}\hfill
  \begin{subfigure}[t]{0.34\linewidth}
    \centering
    \includegraphics[width=\linewidth]{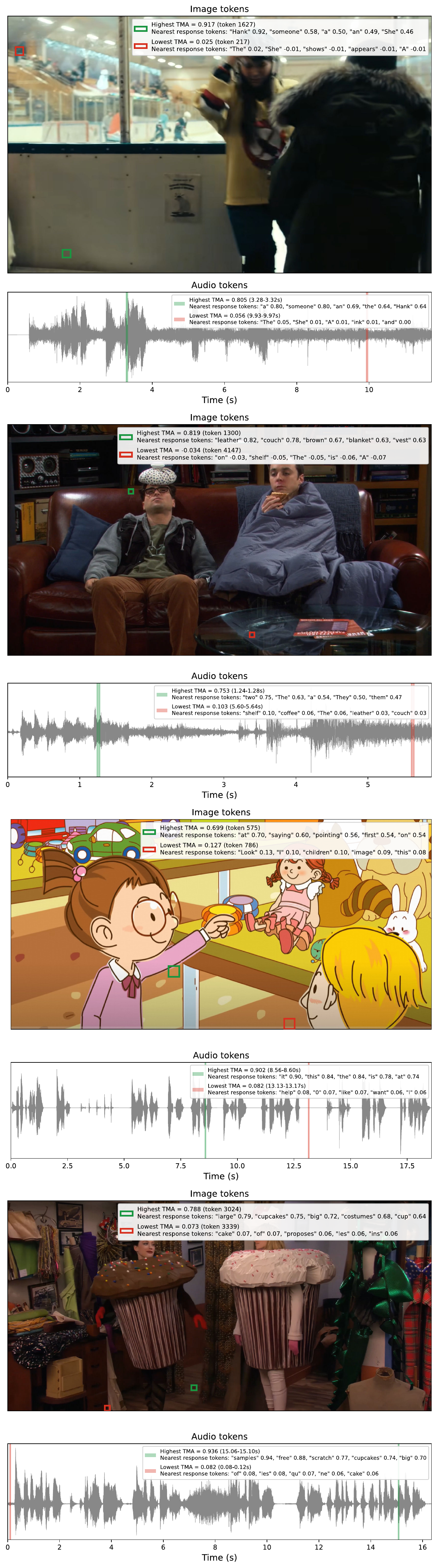}
    \caption{Knowledge task}
    \label{supp_fig:knowledge_qwens}
  \end{subfigure}\hfill
  \begin{subfigure}[t]{0.34\linewidth}
    \centering
    \includegraphics[width=\linewidth]{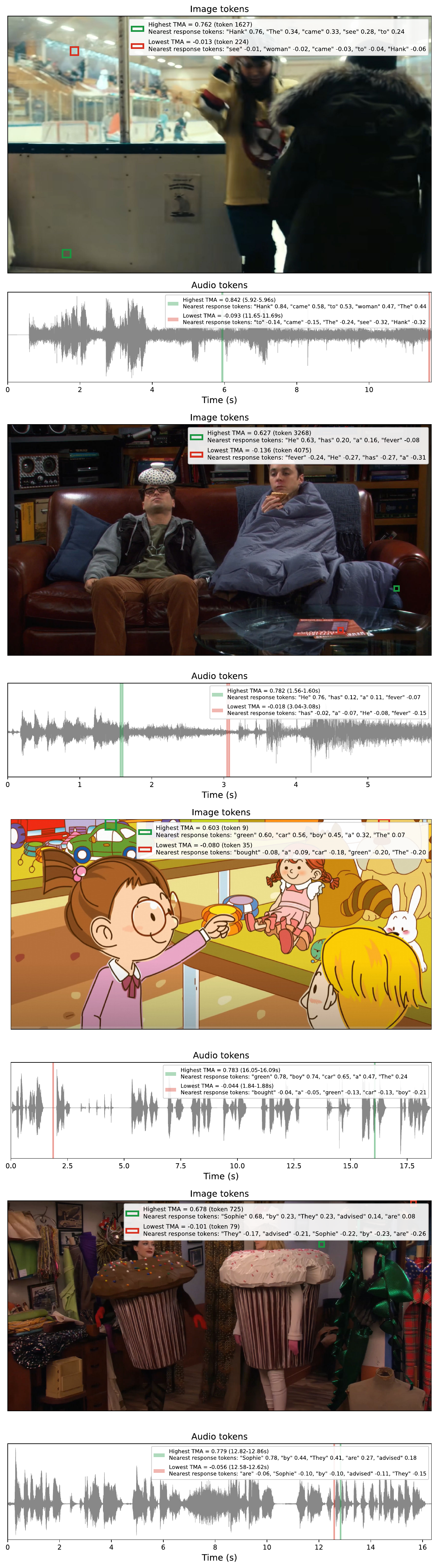}
    \caption{Reasoning task}
    \label{supp_fig:reasoning_qwens}
  \end{subfigure}
  \vspace{-0.7em}
  \caption{\textbf{Sample tokens extracted during knowledge and reasoning tasks for \texttt{Qwen-7B}.}
  Tokens with the highest TMA generally highlight key parts of the image/audio, while tokens with the lowest TMA point to semantically less meaningful regions/frames.}
  \label{supp_fig:knowledge_reasoning_tokens_qwens}
  \vspace{-2em}
\end{figure*}

\begin{figure*}[!ht]
  \centering
  \begin{subfigure}[t]{0.31\linewidth}
    \centering
    \vspace{-46em}
    \includegraphics[width=\linewidth]{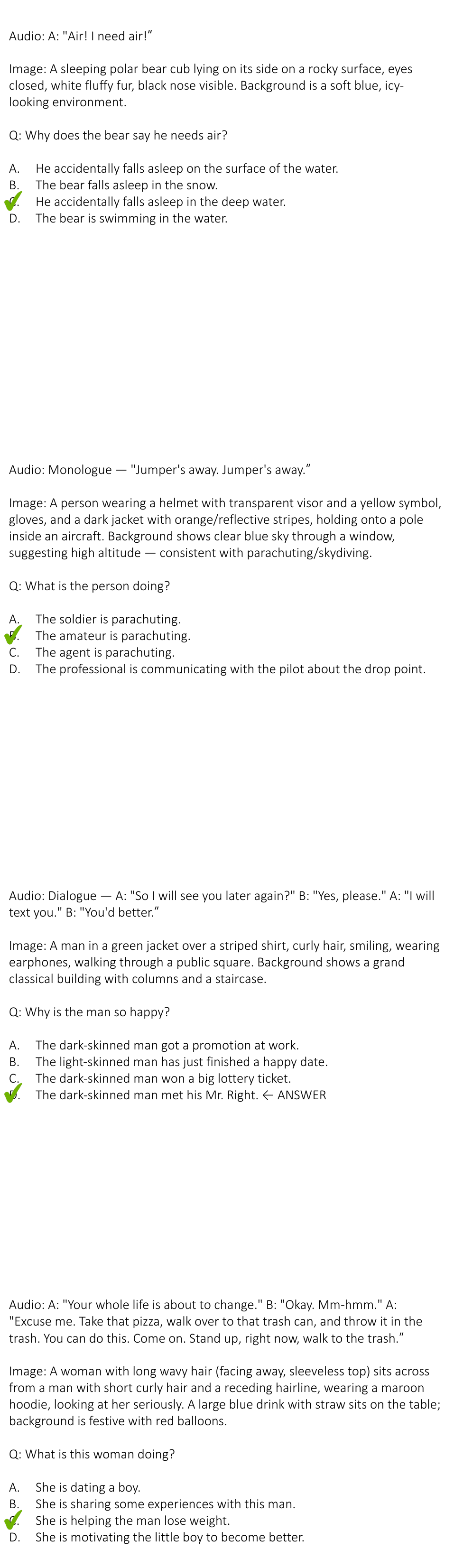}
    \vspace{2.3em}
    \caption{Benchmark question}
    \label{supp_fig:knowledge_reasoning_gemmas}
  \end{subfigure}\hfill
  \begin{subfigure}[t]{0.34\linewidth}
    \centering
    \includegraphics[width=\linewidth]{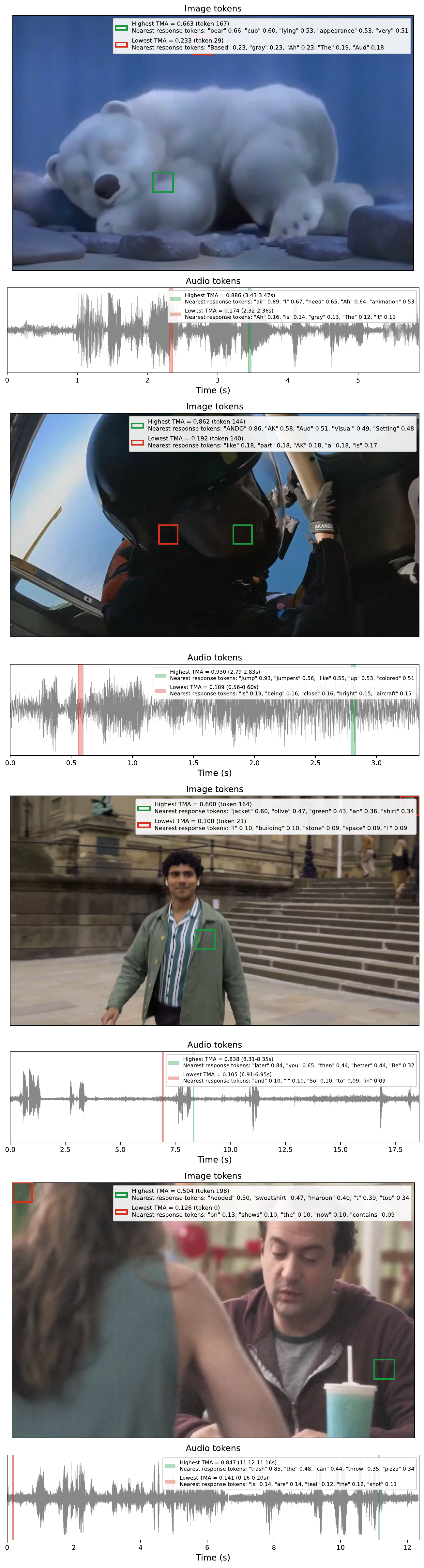}
    \vspace{-1em}
    \caption{Knowledge task}
    \label{supp_fig:knowledge_gemmas}
  \end{subfigure}\hfill
  \begin{subfigure}[t]{0.34\linewidth}
    \centering
    \includegraphics[width=\linewidth]{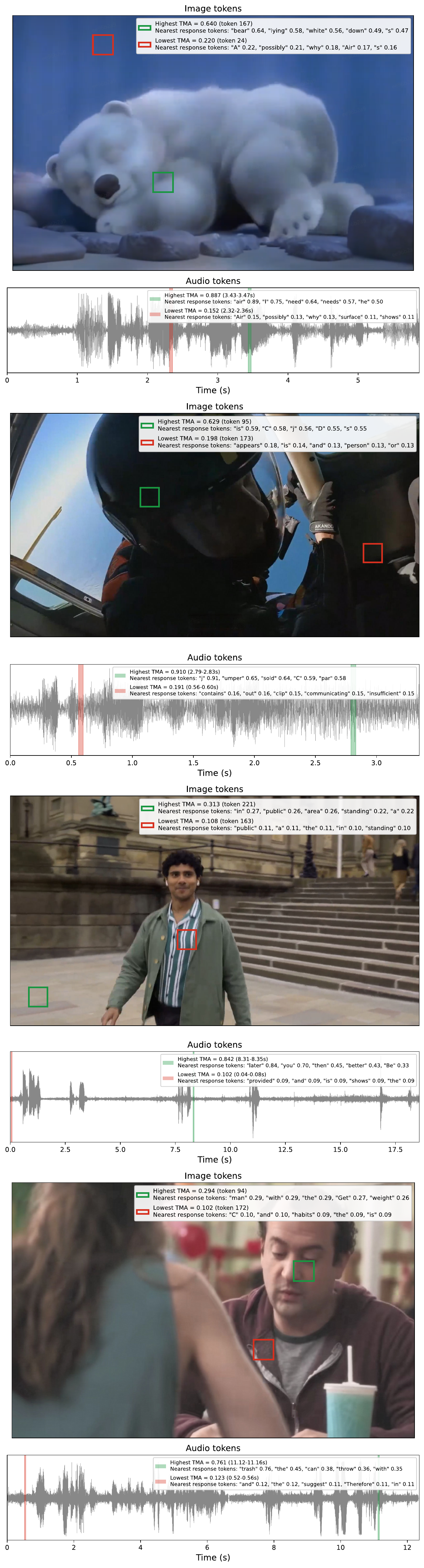}
    \vspace{-1em}
    \caption{Reasoning task}
    \label{supp_fig:reasoning_gemmas}
  \end{subfigure}
  \vspace{-0.7em}
  \caption{\textbf{Sample tokens extracted during knowledge and reasoning tasks for \texttt{gemma-4B}.}
  Tokens with the highest TMA generally highlight key parts of the image/audio, while tokens with the lowest TMA point to semantically less meaningful regions/frames.}
  \label{supp_fig:knowledge_reasoning_tokens_gemmas}
  \vspace{-2.5em}
\end{figure*}

We further inspect the modality token that is most strongly aligned with the generated response.
For each sample, we identify the single highest-TMA token and decode it through its nearest response token.
The accompanying benchmark questions and task-specific visualizations are included in Figures~\ref{supp_fig:knowledge_reasoning_tokens_qwens} and \ref{supp_fig:knowledge_reasoning_tokens_gemmas}.

Across $N=120$ samples per task pooled over four Omni-LLMs, the selected token exhibits a comparatively balanced modality distribution for knowledge evaluation (Shannon entropy $H=0.82$ bits out of a binary maximum of $1$ bit), with $H\in[0.47,1.00]$ across individual models.
For reasoning, the corresponding pooled entropy is substantially lower ($H=0.07$ bits, with $H\in[0,0.21]$ across models).
These examples are consistent with the view that successful reasoning may depend on a relatively localized piece of evidence that is explicitly verbalized, whereas perceptual knowledge can be distributed across multiple modalities and tokens.
In this sense, the model appears to \emph{know} more, diffusely, across modalities than its most response-aligned token alone would reveal during reasoning.

\begin{wrapfigure}{r}{0.30\linewidth}
  \centering
  \vspace{-2em}
  \includegraphics[width=\linewidth]{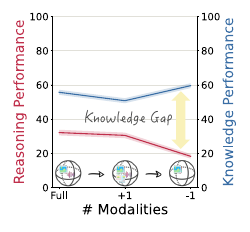}
  \vspace{-2em}
  \caption{\textbf{Modality sensitivity.}
  The gap increases with the \# of missing modalities.
  \label{supp_fig:perceptual_knowledge_omnibench_accuracy}
  }
  \vspace{-3em}
\end{wrapfigure}

\noindent\textbf{Modality-count analysis.}
We illustrate the knowledge and reasoning performance gap by modality count (AVHBench in Figure~\ref{fig:modality_num_performance}; OmniBench in Figure~\ref{supp_fig:perceptual_knowledge_omnibench_accuracy}).
We consider one real modality (averaged across video/image-only and audio-only), two real modalities (video/image+audio), and two real modalities plus task-irrelevant text. 
Together, these conditions distinguish retaining perceptual content from using that content to obtain the correct answer.

Using the same setting as above, we also examine how TMA changes with additional modalities (Figure~\ref{fig:modality_num_TMA}). For each example, we compute TMA for every available input modality and take the maximum value across modalities. We then average the scores across examples and models. The resulting curves therefore characterize how the strongest modality-to-generation alignment changes as additional inputs become available.
The effect is most pronounced when the added modality is semantically relevant to the query.
Because TMA measures alignment with the generated response rather than modality identity, a high score for an irrelevant modality can reflect a generic modality preference and should not be interpreted as evidence of semantic grounding.

\noindent\textbf{Correctness of TMA.}
We validate whether TMA identifies modality tokens that are functionally relevant to generation through tiered masking (Figure~\ref{fig:reasoning_vs_knowledge}).
For each sample, let
$
V_i=M_i^{\mathrm{Visual}}\cup M_i^{\mathrm{Audio}},
$
and partition its modality tokens into low-, mid-, and high-TMA tiers according to within-sample quantiles:
\begin{equation}
V_i^{\mathrm{low}}={j:\mathrm{TMA}_{i,j}\le q_{\alpha}^{(i)}},\quad
V_i^{\mathrm{mid}}={j:q_{\alpha}^{(i)}<\mathrm{TMA}_{i,j}\le q_{\beta}^{(i)}},\quad
V_i^{\mathrm{high}}={j:\mathrm{TMA}_{i,j}>q_{\beta}^{(i)}}.
\end{equation}
For each tier, we zero out the selected representations at every decoder layer during prefill and regenerate the response. We then evaluate both downstream correctness and perceptual knowledge across Omni-LLMs.
The mean TMA of the masked tokens
$
\overline{\mathrm{TMA}}_i^{\mathrm{tier}}
=
\frac{1}{|V_i^{\mathrm{tier}}|}
\sum_{j\in V_i^{\mathrm{tier}}}\mathrm{TMA}_{i,j},
$
is used as the corresponding ablated performance.
Masking high-TMA tokens causes a larger degradation in downstream performance, whereas perceptual knowledge remains comparatively stable.
This separation is consistent with TMA capturing accessibility to the generation pathway rather than simply the presence of information in the input representation, capturing the ``knows but cannot say’’ phenomenon.

\noindent\textbf{Robustness of TMA.}
To assess robustness to response variation, we resample each model's response $K$ times from the same fixed multimodal prompt and recompute TMA for each prompt token.
The prefill pass is identical across samples, so modality-token representations remain fixed; only the decoded response varies.
We quantify the stability of the continuous TMA values using the two-way random-effects, single-measure, absolute-agreement Intraclass Correlation Coefficient--ICC(2, 1)~\citep{koo2016guideline}:
\begin{equation}
\mathrm{ICC}(2,1) =
\frac{MS_R-MS_E}
{MS_R+(K-1)MS_E+\frac{K}{N}(MS_C-MS_E)},
\end{equation}
where $MS_R$, $MS_C$, and $MS_E$ denote between-modality token, between-response token, and residual mean squares, respectively. A high $\mathrm{ICC}(2,1)$ indicates that token-level differences in TMA are stable across sampled responses.

We assign each visual or audio token to the same low/mid/high TMG tiers used for masking.
We then quantify the stability of their low/mid/high tier assignments across the $K$ independent resamples using Fleiss' $\kappa$~\citep{fleiss1971measuring}, 
$
\kappa=\frac{\bar P-P_e}{1-P_e},
$
where $\bar P$ is the mean pairwise tier agreement per token and $P_e$ is the chance agreement implied by the marginal tier frequencies.

\section{Additional Results}\label{app:res}

\noindent\textbf{TMA robustness.}
Table~\ref{supp_tab:stability} reports $\mathrm{ICC}(2,1)$ for visual, audio, and random-text tokens and pooled $\kappa$ for visual+audio tokens across $K=5$ independently sampled responses and up to $N=300$ OmniBench samples. TMA shows high consistency across response samples, with $\mathrm{ICC}(2,1)$ of $0.76$, $0.73$, and $0.67$ for visual, audio, and random-text tokens, respectively, in \texttt{gemma-4-e2b-it}, accompanied by $\kappa=0.61$ for low/mid/high TMA tier agreement. The consistency is further improved for larger models, reaching $\mathrm{ICC}(2,1)=0.99$, $0.98$, and $0.92$ and $\kappa=0.93$ for \texttt{Qwen3-Omni-30B}. Although $k^*$ in Eq.~\ref{eq:tma} depends on the sampled response under the argmax formulation, these results indicate that the resulting TMA scores are largely stable across independently sampled responses rather than being driven by response-specific variation. In contrast, random-text tokens exhibit lower consistency, particularly for \texttt{Qwen2.5-Omni-7B} ($\mathrm{ICC}(2,1)=0.54$), as their TMA is unrelated to multimodal content and therefore depends more strongly on incidental lexical overlap with the sampled response.
\begin{wraptable}{r}{0.4\linewidth}
  \centering
  \footnotesize
  \vspace{-0em}
  \tabcolsep=0.15em
  \hspace{-0.5em}
  \begin{tabular}{l ccc c}
    \toprule
    Model & Vis. & Aud. & Txt. & $\kappa$ $\uparrow$ \\
    \midrule
    \texttt{gemma-4-e2b-it} & 76.4 & 72.7 & 67.3 & 60.8 \\
    \texttt{Qwen2.5-Omni-7B} & 81.9 & 75.8 & 53.8 & 59.9 \\
    \texttt{Qwen3-Omni-30B} & 98.9 & 97.9 & 92.4 & 93.0 \\
    \bottomrule
  \end{tabular}%
  \vspace{-0.8em}
  \caption{\textbf{Robustness of TMA to response sampling.}
  Omni-LLMs overall show low variation across $K$ sampled responses, indicated by
  Vis./Aud./Txt. ICC(2,1) ($\uparrow$) and $\kappa$, or agreement of low/mid/high TMA tier across the $K$ responses (all scaled by 100).}
  \label{supp_tab:stability}
  \vspace{-3em}
\end{wraptable}

\begin{figure*}[!ht]
    \centering
    \includegraphics[width=\linewidth]{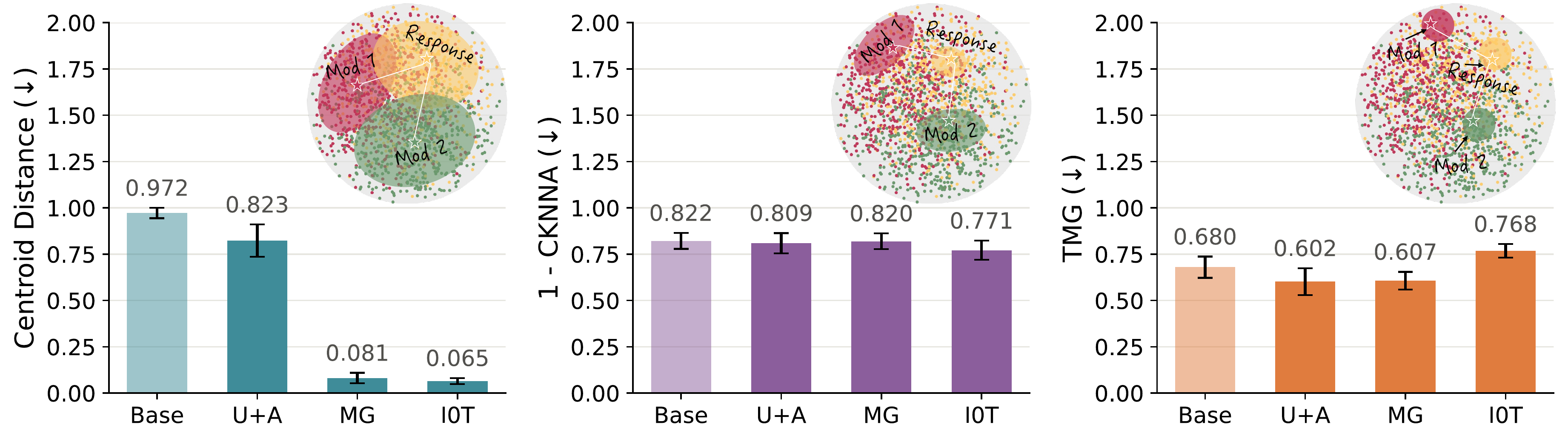}    
    \includegraphics[width=\linewidth]{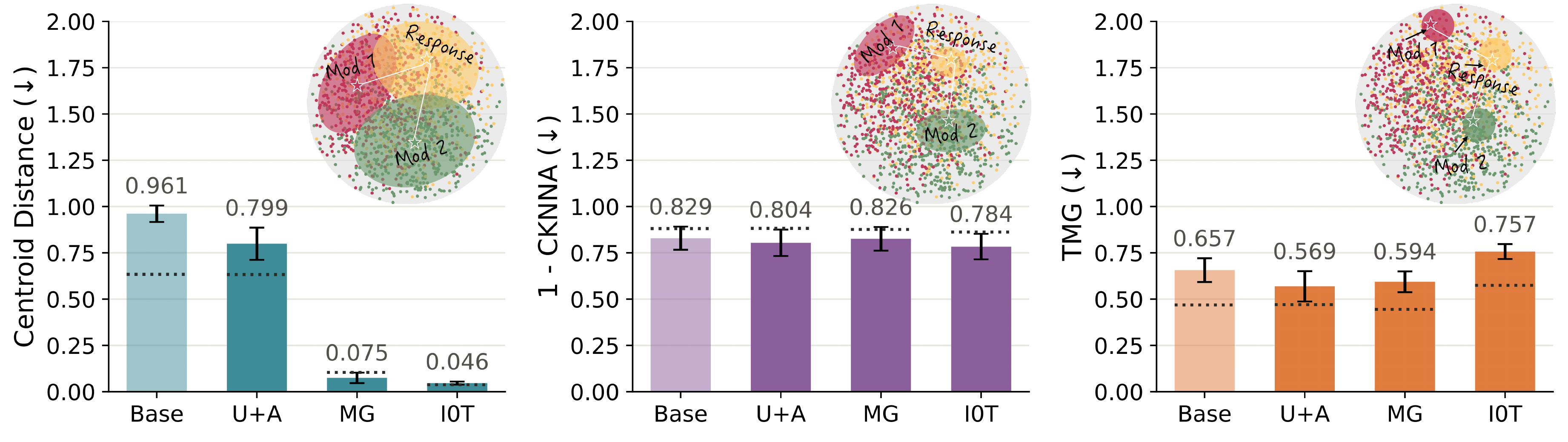}
    \vspace{-2em}
    \caption{\textbf{Modality gap across mitigation methods.}
    We compare global CD, local 1-CKNNA, and 1-TMA without (top) and with (bottom) random descriptions.
    Dotted lines show the corresponding gap for randomly inserted text inputs.
    Prior modality-mitigation methods do not consistently reduce the gap across all three measures, highlighting that improved representational alignment under one measure does not necessarily imply alignment under others.
    }
    \label{supp_fig:modality_gap_measures}
    \vspace{-1em}
\end{figure*}

\noindent\textbf{Modality gap.}
Figure~\ref{supp_fig:modality_gap_measures} compares three measures of modality alignment, expressed as modality gaps: centroid distance (CD), local $1$-CKNNA, and $1$-TMA.
The comparison shows that existing mitigation methods do not consistently reduce all three measures simultaneously.
In particular, randomly inserted text can reduce centroid distance and $1$-TMA despite being semantically unrelated to the response, illustrating that representation proximity alone does not guarantee semantic correspondence.
In contrast, its higher $ 1$-CKNNA reflects the absence of semantically aligned neighboring samples.

\begin{figure*}[!ht]
  \centering
  \includegraphics[width=\linewidth]{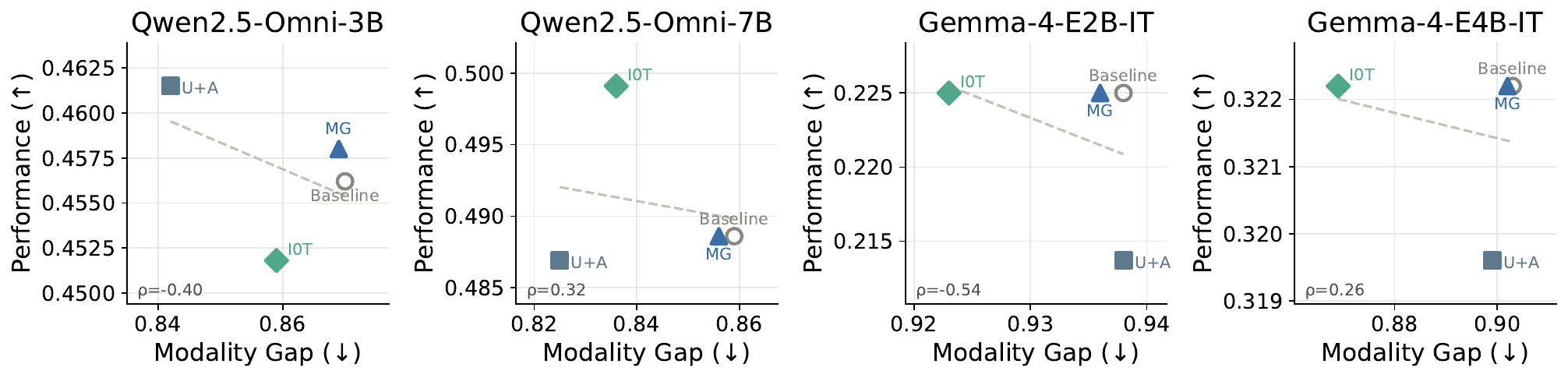}
  \vspace{-2em}
  \caption{\textbf{Modality gap and performance.}
  Existing modality gap mitigation methods yield incremental performance gains as the modality gap, measured by (1-CKNNA), decreases.}
  \label{supp_fig:gap_vs_performance}
  \vspace{-0.5em}
\end{figure*}

Figure~\ref{supp_fig:gap_vs_performance} further relates modality gap to downstream performance.
Across existing modality-gap mitigation methods, lower $1$-CKNNA is associated with incremental performance improvements.
Together, Figures~\ref{supp_fig:modality_gap_measures} and~\ref{supp_fig:gap_vs_performance} highlight the distinction between representational alignment and accessibility to generation.

\noindent\textbf{Modality preference.}
Table~\ref{supp_table:modality_dist} reports the detailed modality preference distributions for each model. Consistent with the main results, \ours{} generally produces a more distributed modality preference than prior modality-gap mitigation methods, as reflected by higher normalized preference entropy.
\begin{table*}[!ht]
\centering
\vspace{-0.5em}
\resizebox{\linewidth}{!}{%
\begin{tabular}{l| cccc| cccc| cccc| cccc| cccc}
\toprule
& \multicolumn{4}{c|}{\textbf{\texttt{Qwen2.5-Omni-3B}}} & \multicolumn{4}{c|}{\textbf{\texttt{Qwen2.5-Omni-7B}}}
& \multicolumn{4}{c|}{\textbf{\texttt{gemma-4-e2b-it}}} & \multicolumn{4}{c|}{\textbf{\texttt{gemma-4-e4b-it}}}
& \multicolumn{4}{c}{\textbf{\emph{Avg.}}} \\
\cmidrule(lr){2-5}\cmidrule(lr){6-9}\cmidrule(lr){10-13}\cmidrule(lr){14-17}\cmidrule(lr){18-21}
\textbf{Condition} & \textbf{T} & \textbf{I} & \textbf{A} & $\hat{\textbf{H}}$ & \textbf{T} & \textbf{I} & \textbf{A} & $\hat{\textbf{H}}$ & \textbf{T} & \textbf{I} & \textbf{A} & $\hat{\textbf{H}}$
& \textbf{T} & \textbf{I} & \textbf{A} & $\hat{\textbf{H}}$ & \textbf{T} & \textbf{I} & \textbf{A} & $\hat{\textbf{H}}$ \\
\midrule
\emph{Ideal} & 33.3 & 33.3 & 33.3 & 1.000 & 33.3 & 33.3 & 33.3 & 1.000 & 33.3 & 33.3 & 33.3 & 1.000 & 33.3 & 33.3 & 33.3 & 1.000 & 33.3 & 33.3 & 33.3 & 1.000 \\
\midrule
\multicolumn{21}{l}{\textit{Baselines}} \\
No augmentation & 23.4 & 56.1 & 20.5 & 0.900 & 8.6 & 59.1 & 32.3 & 0.807 & 24.8 & 51.1 & 24.1 & 0.939 & 21.3 & 61.3 & 17.4 & 0.849 & 19.5 & 56.9 & 23.5 & 0.892 \\
MG & 10.1 & 60.6 & 29.3 & 0.814 & 1.7 & 58.9 & 39.4 & 0.681 & 12.0 & 70.2 & 17.8 & 0.737 & 11.4 & 64.9 & 23.7 & 0.791 & 8.8 & 63.7 & 27.5 & 0.780 \\
IoT & 72.8 & 22.1 & 5.1 & 0.652 & 78.3 & 16.2 & 5.5 & 0.587 & 41.6 & 19.9 & 38.5 & \textbf{0.959} & 68.3 & 19.6 & 12.1 & 0.761 & 65.3 & 19.4 & 15.3 & 0.805 \\
Self-CoT & 22.5 & 45.1 & 32.5 & 0.965 & 9.2 & 58.8 & 32.0 & 0.816 & 20.4 & 53.8 & 25.8 & 0.917 & 20.1 & 62.5 & 17.3 & 0.837 & 18.1 & 55.0 & 26.9 & 0.902 \\
\midrule
\multicolumn{21}{l}{\textit{Single-modality masking with its description}} \\
Mask image & 1.1 & 87.7 & 11.2 & 0.372 & 0.5 & 92.1 & 7.4 & 0.269 & 1.8 & 85.0 & 13.2 & 0.436 & 3.4 & 80.7 & 15.9 & 0.528 & 1.7 & 86.4 & 11.9 & 0.409 \\
Mask audio & 1.9 & 7.7 & 90.5 & 0.329 & 3.0 & 9.4 & 87.6 & 0.405 & 7.5 & 39.7 & 52.8 & 0.818 & 12.6 & 31.1 & 56.2 & 0.863 & 6.3 & 22.0 & 71.8 & 0.678 \\
\midrule
\multicolumn{21}{l}{\textit{Oracle descriptions}} \\
GT text only & 33.3 & 48.4 & 18.3 & 0.936 & 11.7 & 59.6 & 28.7 & 0.835 & 38.8 & 39.7 & 21.5 & 0.969 & 29.2 & 58.8 & 12.0 & 0.844 & 28.2 & 51.6 & 20.1 & 0.930 \\
GT image only & 5.2 & 86.8 & 8.0 & 0.436 & 2.1 & 86.3 & 11.6 & 0.417 & 3.7 & 92.5 & 3.8 & 0.290 & 4.7 & 92.4 & 3.0 & 0.292 & 3.9 & 89.5 & 6.6 & 0.369 \\
GT audio only & 12.0 & 27.1 & 60.8 & 0.829 & 1.5 & 26.0 & 72.5 & 0.589 & 13.5 & 38.4 & 48.1 & 0.902 & 12.7 & 35.6 & 51.8 & 0.883 & 9.9 & 31.8 & 58.3 & 0.827 \\
GT all & 4.2 & 57.3 & 38.4 & 0.747 & 0.4 & 53.0 & 46.6 & 0.651 & 5.1 & 72.3 & 22.6 & 0.658 & 5.6 & 65.1 & 29.3 & 0.728 & 3.8 & 61.9 & 34.2 & 0.718 \\
\midrule
\rowcolor{gray!15} \multicolumn{21}{l}{\textit{\ours{} (two-pass)}} \\
\rowcolor{gray!15} Text only & 32.4 & 49.6 & 18.0 & 0.930 & 13.6 & 56.9 & 29.5 & \underline{0.867} & 44.6 & 35.5 & 19.9 & 0.955 & 32.0 & 54.7 & 13.3 & 0.877 & 30.6 & 49.2 & 20.2 & 0.942 \\
\rowcolor{gray!15} Image only & 15.7 & 66.0 & 18.3 & 0.797 & 7.5 & 64.1 & 28.4 & 0.761 & 12.1 & 64.0 & 23.9 & 0.804 & 13.0 & 73.1 & 13.9 & 0.700 & 12.1 & 66.8 & 21.1 & 0.777 \\
\rowcolor{gray!15} Audio only & 25.4 & 43.6 & 30.9 & 0.977 & 4.5 & 45.9 & 49.6 & 0.768 & 30.6 & 47.1 & 22.3 & \underline{0.957} & 23.5 & 56.8 & 19.7 & 0.894 & 21.0 & 48.4 & 30.7 & \underline{0.948} \\
\rowcolor{gray!15} All & 13.9 & 52.4 & 33.7 & 0.892 & 2.1 & 54.6 & 43.3 & 0.705 & 14.9 & 59.3 & 25.8 & 0.858 & 13.7 & 65.6 & 20.7 & 0.796 & 11.1 & 58.0 & 30.9 & 0.841 \\
\midrule
\rowcolor{gray!15} \multicolumn{21}{l}{\textit{\ours{} (single-pass)}} \\
\rowcolor{gray!15} Text only & 27.0 & 41.6 & 31.4 & \textbf{0.985} & 10.9 & 46.2 & 42.9 & \textbf{0.875} & 20.6 & 49.4 & 30.1 & 0.942 & 25.2 & 55.9 & 18.9 & 0.898 & 20.9 & 48.3 & 30.8 & \underline{0.948} \\
\rowcolor{gray!15} Image only & 23.9 & 47.9 & 28.1 & 0.957 & 8.2 & 61.7 & 30.1 & 0.787 & 11.3 & 46.0 & 42.7 & 0.881 & 30.5 & 51.9 & 17.5 & 0.917 & 18.5 & 51.9 & 29.6 & 0.922 \\
\rowcolor{gray!15} Audio only & 25.6 & 32.2 & 42.2 & \underline{0.981} & 7.4 & 35.4 & 57.2 & 0.802 & 10.5 & 44.2 & 45.4 & 0.870 & 36.0 & 41.2 & 22.8 & \textbf{0.974} & 19.9 & 38.2 & 41.9 & \textbf{0.959} \\
\rowcolor{gray!15} All & 26.1 & 42.7 & 31.2 & \underline{0.981} & 9.6 & 63.7 & 26.7 & 0.786 & 13.2 & 48.1 & 38.7 & 0.899 & 32.6 & 48.0 & 19.4 & \underline{0.943} & 20.4 & 50.6 & 29.0 & 0.936 \\
\bottomrule
\end{tabular}
}
\vspace{-0.5em}
\caption{\textbf{Modality preference distributions and balance.}
For each model, we report the softmax-normalized
preference mass over Text (T), Image (I), and Audio (A), together with the normalized entropy $\hat{H}=H/\log_2 3$, where $\hat{H}=1$ denotes perfectly balanced preference. 
\textbf{Bold} and \underline{underline} mark the highest and second-highest $\hat{H}$ per column
among non-oracle conditions.}
\vspace{-0.5em}
\label{supp_table:modality_dist}
\end{table*}

\noindent\textbf{Additional Omni-LLMs.}
Table~\ref{supp_table:remaining_ollms} extends the main evaluation to additional Omni-LLMs.
\ours{} consistently improves performance across these models, with particularly large gains for \texttt{OmniVinci} (+11.5 percentage points on average).
The larger \texttt{Qwen3-Omni-30B-A3B-Instruct} also benefits from \ours{}, although the gain is smaller than that observed for the smaller models.
These results indicate that the method is not limited to a particular model scale or architecture.
\begin{table*}[!ht]
\centering
\resizebox{0.5\linewidth}{!}{%
\begin{tabular}{l|cccc}
\toprule
\textbf{Method} & \textbf{Music} & \begin{tabular}[c]{@{}c@{}}\textbf{Sound}\\ \textbf{Event}\end{tabular} & \textbf{Speech} & \textbf{\emph{Avg.}} \\
\midrule
\texttt{Qwen3-Omni-30B-A3B-Instruct} & 53.8 & 60.0 & 48.4 & 51.6 \\
\rowcolor{gray!15} \ours{}      & 53.8\up{0.0} & 58.9\down{1.1} & 50.3\up{1.9} & 52.6\up{1.0} \\

\texttt{omnivinci}              & 44.3 & 38.1 & 35.3 & 36.8 \\
\rowcolor{gray!15} \ours{}      & 57.6\up{13.3} & 48.3\up{10.2} & 47.0\up{11.7} & 48.3\up{11.5} \\

\texttt{MiniCPM-o-2\_6}         & 50.0 & 40.0 & 37.6 & 39.1 \\
\rowcolor{gray!15} \ours{}      & 48.1\down{1.9} & 45.7\up{5.7} & 39.8\up{2.2} & 42.0\up{2.9} \\
\bottomrule
\end{tabular}
}
\vspace{-0.5em}
\caption{\textbf{OmniBench performance on additional Omni-LLMs.} 
\ours{} improves performance across additional Omni-LLMs, with particularly large gains for \texttt{omnivinci}.
}
\vspace{-1.0em}
\label{supp_table:remaining_ollms}
\end{table*}

\noindent\textbf{Training ablation.}
Table~\ref{supp_table:selftrain_comparison} reports per-model results for the different training strategies.
Applying training-free \ours{} at inference time improves trained models' performance, while incorporating modality-description generation into training provides additional gains.
This comparison separates the benefit of the inference-time mechanism from that of learning the description capability.
\begin{table*}[!ht]
\centering
\small
\setlength{\tabcolsep}{3pt}
\resizebox{0.5\linewidth}{!}{%
\begin{tabular}{l|cccc|c}
\toprule
\textbf{Method} & \textbf{\texttt{gemma-2B}} & \textbf{\texttt{gemma-4B}} & \textbf{\texttt{Qwen-3B}} & \textbf{\texttt{Qwen-7B}} & \textbf{\emph{Avg.}} \\
\midrule
Baseline & 21.4 & 27.1 & 43.8 & 46.5 & 34.7 \\
\rowcolor{gray!15} \quad \ours{} & 32.6\up{11.2} & 40.5\up{13.4} & 49.6\up{5.8} & 51.7\up{5.2} & 43.6\up{8.9} \\

\rowcolor{gray!15} Infer-only & 22.0 & 32.1 & 44.7 & 48.1 & 36.7 \\
\rowcolor{gray!15} \quad \ours{} & 32.0\up{10.0} & 39.8\up{7.7} & 46.2\up{1.5} & 52.8\up{4.7} & 42.7\up{6.0} \\

\rowcolor{gray!15} Infer+Perceive & 22.8 & 32.6 & 45.1 & 48.4 & 37.2 \\
\rowcolor{gray!15} \quad \ours{} & 32.1\up{9.3} & 40.3\up{7.7} & 44.0\down{1.1} & 53.0\up{4.6} & 42.4\up{5.1} \\

\rowcolor{gray!15} STaR~\citep{zelikman2022star} & 22.6 & 31.7 & 44.9 & 48.9 & 37.0 \\
\rowcolor{gray!15} \quad \ours{} & 32.2\up{9.6} & 40.1\up{8.4} & 48.2\up{3.3} & 52.1\up{3.2} & 43.2\up{6.1} \\

\rowcolor{gray!15} R3V~\citep{cheng2025vision} & 22.6 & 32.0 & 45.5 & 48.2 & 37.1 \\
\rowcolor{gray!15} \quad \ours{} & 32.1\up{9.5} & 40.2\up{8.2} & 47.0\up{1.5} & 52.2\up{4.0} & 42.9\up{5.8} \\

\rowcolor{gray!15} SD-Zero~\citep{he2026self} & 22.6 & 32.2 & 45.2 & 49.1 & 37.3 \\
\rowcolor{gray!15} \quad \ours{} & 31.5\up{8.9} & 40.4\up{8.2} & 46.9\up{1.7} & 52.4\up{3.3} & 42.8\up{5.5} \\

\rowcolor{gray!15} Perceive-only & 22.9 & 31.5 & 46.2 & 49.3 & 37.5 \\
\rowcolor{gray!15} \quad \ours{} & 31.6\up{8.7} & 40.2\up{8.7} & 48.5\up{2.3} & 53.2\up{3.9} & 43.4\up{5.9} \\

\rowcolor{gray!15} Perceive + STaR & 22.2 & 33.0 & 45.9 & 49.1 & 37.6 \\
\rowcolor{gray!15} \quad \ours{} & 31.6\up{9.4} & 40.4\up{7.4} & 44.8\down{1.1} & 51.3\up{2.2} & 42.0\up{4.5} \\

\bottomrule
\end{tabular}%
}
\vspace{-0.5em}
\caption{\textbf{Adaptability across training strategies.}
Training-free \ours{} consistently improves models trained with different objectives applied with our \emph{perceive} $\rightarrow$ \emph{infer} paradigm.}
\vspace{-2em}
\label{supp_table:selftrain_comparison}
\end{table*}

Table~\ref{supp_table:training_ablation} further examines modality-description configurations for Omni-LLMs trained with the single-pass formulation across target modalities.
Single-pass training generally improves the effectiveness of inference-time descriptions across trained modalities and description types.
Among the two-turn configurations, audio descriptions provide the most consistent improvement, regardless of whether the model was trained with audio, visual, or all-modality descriptions.
Visual or all-modality descriptions produce smaller or occasionally negative gains.
This pattern is consistent with the main observation that audio is often the less accessible modality and therefore benefits particularly from an explicit textual description.
\begin{table*}[!ht]
\centering
\vspace{-0.5em}
\resizebox{0.86\linewidth}{!}{%
\begin{tabular}{l|cc|cccc|c}
\toprule
\textbf{Method} & \textbf{Trained Modality Desc} & \textbf{$+$Modality Desc} & \textbf{\texttt{gemma-2B}} & \textbf{\texttt{gemma-4B}} & \textbf{\texttt{Qwen-3B}} & \textbf{\texttt{Qwen-7B}} & \textbf{\emph{Avg.}} \\
\midrule
\rowcolor{gray!15} ST \ours{} & Audio & - & 15.1 & \textbf{27.5} & 44.8 & 49.8 & 34.3 \\
\rowcolor{gray!15} ST \ours{} & Audio & Audio & \underline{29.3}\up{14.2} & \underline{36.1}\up{8.6} & \textbf{47.7}\up{2.9} & \textbf{55.9}\up{6.1} & \textbf{42.3}\up{8.0} \\
\rowcolor{gray!15} ST \ours{} & Audio & Visual & \underline{30.1}\up{15.0} & \textbf{38.9}\up{11.4} & \underline{42.2}\down{2.6} & \underline{52.9}\up{3.1} & \textbf{41.0}\up{6.7} \\
\rowcolor{gray!15} ST \ours{} & Audio & All & \textbf{29.3}\up{14.2} & \textbf{35.7}\up{8.2} & \textbf{43.7}\down{1.1} & \textbf{53.9}\up{4.1} & \textbf{40.7}\up{6.4} \\

\rowcolor{gray!15} ST \ours{} & Audio & Audio (two turns) & 19.3\up{4.2} & 26.4\down{1.1} & \textbf{48.1}\up{3.3} & 51.0\up{1.2} & \underline{36.2}\up{1.9} \\
\rowcolor{gray!15} ST \ours{} & Audio & Visual (two turns) & 15.8\up{0.7} & \underline{23.8}\down{3.7} & 41.7\down{3.1} & 47.3\down{2.5} & 32.2\down{2.1} \\
\rowcolor{gray!15} ST \ours{} & Audio & All (two turns) & 18.4\up{3.3} & 22.9\down{4.6} & 43.4\down{1.4} & \underline{48.7}\down{1.1} & 33.4\down{0.9} \\

\midrule

\rowcolor{gray!15} ST \ours{} & Perceive$\rightarrow$Audio & - & 2.2 & 3.9 & 45.4 & \underline{50.3} & 25.5 \\
\rowcolor{gray!15} ST \ours{} & Perceive$\rightarrow$Audio & Audio & 18.0\up{15.8} & 21.1\up{17.2} & 42.3\down{3.1} & \underline{54.0}\up{3.7} & 33.9\up{8.4} \\
\rowcolor{gray!15} ST \ours{} & Perceive$\rightarrow$Audio & Visual & 18.5\up{16.3} & 26.3\up{22.4} & 38.2\down{7.2} & \textbf{53.7}\up{3.4} & 34.2\up{8.7} \\
\rowcolor{gray!15} ST \ours{} & Perceive$\rightarrow$Audio & All & 12.9\up{10.7} & 25.0\up{21.1} & 39.6\down{5.8} & \underline{53.1}\up{2.8} & 32.7\up{7.2} \\

\rowcolor{gray!15} ST \ours{} & Perceive$\rightarrow$Audio & Audio (two turns) & 8.5\up{6.3} & 11.0\up{7.1} & 47.0\up{1.6} & \textbf{51.8}\up{1.5} & 29.6\up{4.1} \\
\rowcolor{gray!15} ST \ours{} & Perceive$\rightarrow$Audio & Visual (two turns) & 1.5\down{0.7} & 3.2\down{0.7} & 41.9\down{3.5} & \textbf{49.4}\down{0.9} & 24.0\down{1.5} \\
\rowcolor{gray!15} ST \ours{} & Perceive$\rightarrow$Audio & All (two turns) & 6.3\up{4.1} & 7.6\up{3.7} & 43.6\down{1.8} & \textbf{49.7}\down{0.6} & 26.8\up{1.3} \\

\midrule

\rowcolor{gray!15} ST \ours{} & Visual & - & 18.6 & \underline{27.4} & 44.5 & 48.2 & \underline{34.7} \\
\rowcolor{gray!15} ST \ours{} & Visual & Audio & 25.6\up{7.0} & \textbf{36.3}\up{8.9} & \underline{47.0}\up{2.5} & 53.6\up{5.4} & \underline{40.6}\up{5.9} \\
\rowcolor{gray!15} ST \ours{} & Visual & Visual & 26.8\up{8.2} & \underline{37.7}\up{10.3} & 41.4\down{3.1} & 50.2\up{2.0} & \underline{39.0}\up{4.3} \\
\rowcolor{gray!15} ST \ours{} & Visual & All & 20.5\up{1.9} & 34.8\up{7.4} & 42.3\down{2.2} & 50.4\up{2.2} & \underline{37.0}\up{2.3} \\

\rowcolor{gray!15} ST \ours{} & Visual & Audio (two turns) & 20.0\up{1.4} & \underline{27.3}\down{0.1} & 46.2\up{1.7} & \underline{51.1}\up{2.9} & \underline{36.2}\up{1.5} \\
\rowcolor{gray!15} ST \ours{} & Visual & Visual (two turns) & 16.0\down{2.6} & 22.6\down{4.8} & 43.2\down{1.3} & 45.6\down{2.6} & 31.9\down{2.8} \\
\rowcolor{gray!15} ST \ours{} & Visual & All (two turns) & 18.3\down{0.3} & 23.7\down{3.7} & 44.8\up{0.3} & 48.4\up{0.2} & 33.8\down{0.9} \\

\midrule

\rowcolor{gray!15} ST \ours{} & Perceive$\rightarrow$Visual & - & \underline{19.9} & 21.1 & 44.5 & 49.6 & 33.8 \\
\rowcolor{gray!15} ST \ours{} & Perceive$\rightarrow$Visual & Audio & 21.1\up{1.2} & 32.1\up{11.0} & 44.8\up{0.3} & \underline{54.0}\up{4.4} & 38.0\up{4.2} \\
\rowcolor{gray!15} ST \ours{} & Perceive$\rightarrow$Visual & Visual & 24.5\up{4.6} & 36.0\up{14.9} & 39.5\down{5.0} & 50.9\up{1.3} & 37.7\up{3.9} \\
\rowcolor{gray!15} ST \ours{} & Perceive$\rightarrow$Visual & All & 17.2\down{2.7} & 33.3\up{12.2} & 42.5\down{2.0} & 51.2\up{1.6} & 36.1\up{2.3} \\

\rowcolor{gray!15} ST \ours{} & Perceive$\rightarrow$Visual & Audio (two turns) & 17.7\down{2.2} & 22.9\up{1.8} & 46.9\up{2.4} & 50.5\up{0.9} & 34.5\up{0.7} \\
\rowcolor{gray!15} ST \ours{} & Perceive$\rightarrow$Visual & Visual (two turns) & 9.5\down{10.4} & 18.0\down{3.1} & 41.9\down{2.6} & \underline{47.5}\down{2.1} & 29.2\down{4.6} \\
\rowcolor{gray!15} ST \ours{} & Perceive$\rightarrow$Visual & All (two turns) & 13.7\down{6.2} & 15.7\down{5.4} & 44.3\down{0.2} & 46.8\down{2.8} & 30.1\down{3.7} \\

\midrule

\rowcolor{gray!15} ST \ours{} & All & - & \textbf{22.9} & 23.2 & \textbf{47.6} & 49.7 & \textbf{35.9} \\
\rowcolor{gray!15} ST \ours{} & All & Audio & \textbf{30.5}\up{7.6} & 30.3\up{7.1} & 45.7\down{1.9} & 52.1\up{2.4} & 39.7\up{3.8} \\
\rowcolor{gray!15} ST \ours{} & All & Visual & \textbf{30.5}\up{7.6} & 30.2\up{7.0} & 40.0\down{7.6} & 48.4\down{1.3} & 37.3\up{1.4} \\
\rowcolor{gray!15} ST \ours{} & All & All & \underline{28.7}\up{5.8} & 26.1\up{2.9} & 42.7\down{4.9} & 46.9\down{2.8} & 36.1\up{0.2} \\

\rowcolor{gray!15} ST \ours{} & All & Audio (two turns) & \underline{22.1}\down{0.8} & \textbf{27.4}\up{4.2} & 47.6\up{0.0} & 49.8\up{0.1} & \textbf{36.7}\up{0.8} \\
\rowcolor{gray!15} ST \ours{} & All & Visual (two turns) & \textbf{18.5}\down{4.4} & \textbf{26.2}\up{3.0} & \underline{45.2}\down{2.4} & 47.1\down{2.6} & \textbf{34.3}\down{1.6} \\
\rowcolor{gray!15} ST \ours{} & All & All (two turns) & \underline{21.3}\down{1.6} & \textbf{26.4}\up{3.2} & \underline{45.9}\down{1.7} & 47.5\down{2.2} & \textbf{35.3}\down{0.6} \\

\midrule

\rowcolor{gray!15} ST \ours{} & Perceive$\rightarrow$All & - & 18.6 & 21.0 & \underline{46.6} & \textbf{50.5} & 34.2 \\
\rowcolor{gray!15} ST \ours{} & Perceive$\rightarrow$All & Audio & 24.3\up{5.7} & 34.0\up{13.0} & 43.3\down{3.3} & 53.2\up{2.7} & 38.7\up{4.5} \\
\rowcolor{gray!15} ST \ours{} & Perceive$\rightarrow$All & Visual & 29.1\up{10.5} & 29.9\up{8.9} & \textbf{44.8}\down{1.8} & 49.0\down{1.5} & 38.2\up{4.0} \\
\rowcolor{gray!15} ST \ours{} & Perceive$\rightarrow$All & All & 21.5\up{2.9} & \underline{35.1}\up{14.1} & \underline{42.8}\down{3.8} & 48.3\down{2.2} & 36.9\up{2.7} \\

\rowcolor{gray!15} ST \ours{} & Perceive$\rightarrow$All & Audio (two turns) & \textbf{26.4}\up{7.8} & 23.7\up{2.7} & \underline{47.8}\up{1.2} & 49.0\down{1.5} & \textbf{36.7}\up{2.5} \\
\rowcolor{gray!15} ST \ours{} & Perceive$\rightarrow$All & Visual (two turns) & \underline{16.9}\down{1.7} & 23.5\up{2.5} & \textbf{46.0}\down{0.6} & 45.8\down{4.7} & \underline{33.1}\down{1.1} \\
\rowcolor{gray!15} ST \ours{} & Perceive$\rightarrow$All & All (two turns) & \textbf{21.5}\up{2.9} & \underline{25.3}\up{4.3} & \textbf{46.2}\down{0.4} & 46.1\down{4.4} & \underline{34.8}\up{0.6} \\

\bottomrule
\end{tabular}%
}
\vspace{-0.5em}
\caption{\textbf{Self-trained (ST) \ours{} ablation results.} 
Performance of self-trained \ours{} under different trained and inference-time modality-description configurations.
\textbf{Bold} and \underline{underline} denote the best and second-best results, respectively, for each $+$Modality Desc setting across trained modalities.}
\vspace{-0.8em}
\label{supp_table:training_ablation}
\end{table*}

\noindent\textbf{Transfer to audio captioning.}
Table~\ref{supp_table:captioning} evaluates trained models on an out-of-distribution audio-captioning task.
Perceive + STaR improves captioning performance for both Gemma and Qwen compared with both the baseline and perceive-only training.
The improvement suggests that the learned ability to generate modality descriptions can transfer beyond the multiple-choice reasoning tasks used for training.
\begin{table*}[!ht]
\centering
\vspace{-0.5em}
\small
\setlength{\tabcolsep}{3pt}
\resizebox{0.5\linewidth}{!}{%
\begin{tabular}{l|l|cccc}
\toprule
\textbf{Model} & \textbf{Method} & \textbf{BLEU} & \textbf{METEOR} & \textbf{ROUGE-L} & \textbf{CIDEr} \\
\midrule
\texttt{gemma} & Baseline & 12.8 & 8.1 & 23.0 & 4.4 \\
\rowcolor{gray!15} \texttt{gemma} & Perceive-only & 12.7\down{0.1} & 8.1\down{0.0} & 23.0\down{0.0} & 4.4\down{0.0} \\
\rowcolor{gray!15} \texttt{gemma} & Perceive + STaR & 13.0\up{0.3} & 8.2\up{0.1} & 23.1\up{0.1} & 4.6\up{0.2} \\

\texttt{Qwen} & Baseline & 28.2 & 15.8 & 34.9 & 35.6 \\
\rowcolor{gray!15} \texttt{Qwen} & Perceive-only & 28.7\up{0.5} & 16.0\up{0.2} & 35.7\up{0.8} & 36.5\up{0.9} \\
\rowcolor{gray!15} \texttt{Qwen} & Perceive + STaR & 31.4\up{3.2} & 16.3\up{0.5} & 36.0\up{1.1} & 37.1\up{1.5} \\
\bottomrule
\end{tabular}%
}
\vspace{-0.5em}
\captionof{table}{\textbf{Out-of-distribution audio captioning.}
Audio-captioning performance of models trained with different description objectives. Perceive + STaR improves over both the baseline and perceive-only training for both Gemma and Qwen.}
\label{supp_table:captioning}
\vspace{-2em}
\end{table*}

\noindent\textbf{External description models.}
Table~\ref{supp_table:external_model} evaluates descriptions generated by larger external Omni-LLMs.
Using these external descriptions increases average performance from 34.2 to 39.8, indicating that \ours{} benefits from higher-quality modality descriptions.
This experiment also separates the utility of the textual modality representation from the identity of the model that generates it.
\begin{table*}[!ht]
\centering
\vspace{-0.8em}
\resizebox{0.6\linewidth}{!}{%
\begin{tabular}{l|cccc}
\toprule
\textbf{Model} & \textbf{No Description}  & \textbf{\texttt{Qwen2.5-Omni-7B}} & \textbf{\texttt{Qwen3-Omni-30B}} \\
\midrule
\texttt{gemma-4-e2b-it} & 23.0  &29.0\up{6.0} & 29.3\up{6.3} \\
\texttt{gemma-4-e4b-it} & 32.0  & 35.0\up{3.0} & 37.7\up{5.7} \\
\texttt{Qwen2.5-Omni-3B} & 39.3  &45.3\up{6.0} & 44.3\up{5.0} \\
\texttt{Qwen2.5-Omni-7B} & 42.3  &44.0\up{1.7} & 48.0\up{5.7} \\
\midrule
\emph{Avg.} & 34.2 & 38.3\up{4.2} & 39.8\up{5.7} \\
\bottomrule
\end{tabular}
}
\vspace{-0.5em}
\caption{\textbf{External model usage for modality description generation in \ours{}.} 
The final omni-modal reasoning performance does not significantly differ when using modality descriptions generated by external models (spanned across columns) for the two-call scheme.
}
\label{supp_table:external_model}
\vspace{-1.2em}
\end{table*}

\noindent\textbf{Thinker mode.}
Table~\ref{supp_table:thinker_mode} evaluates \ours{} with thinker mode enabled for Gemma.
Although the main Gemma experiments use non-thinker mode for consistency with Qwen, \ours{} improves performance in both settings.
Thus, the benefit of modality descriptions complements the model's internal thinking mode rather than depending on a particular decoding configuration.
\begin{table*}[!ht]
\centering
\resizebox{0.6\linewidth}{!}{%
\begin{tabular}{l|cccc}
\toprule
& \multicolumn{2}{c}{\textbf{No Thinker Mode}} & \multicolumn{2}{c}{\textbf{Thinker Mode}} \\
\textbf{Model} & \textbf{Baseline} & \textbf{\ours{}} & \textbf{Baseline} & \textbf{\ours{}} \\
\midrule
\texttt{gemma-4-e2b-it} & 23.7 & 28.3\up{4.6} & 36.1 & 37.3\up{1.2} \\
\texttt{gemma-4-e4b-it} & 34.0 & 42.0\up{8.0} & 42.0 & 42.3\up{2.3} \\
\bottomrule
\end{tabular}
}
\vspace{-0.5em}
\caption{\textbf{Effect of thinker mode on the Gemma family.} 
\ours{} improves performance in both thinker and non-thinker modes, indicating that its benefit complements the model’s thinking mode.
}
\label{supp_table:thinker_mode}
\vspace{-1.5em}
\end{table*}

\begin{figure*}[!ht]
  \centering
  \vspace{-1em}
  \begin{subfigure}{\linewidth}
    \centering
    \includegraphics[width=\linewidth]{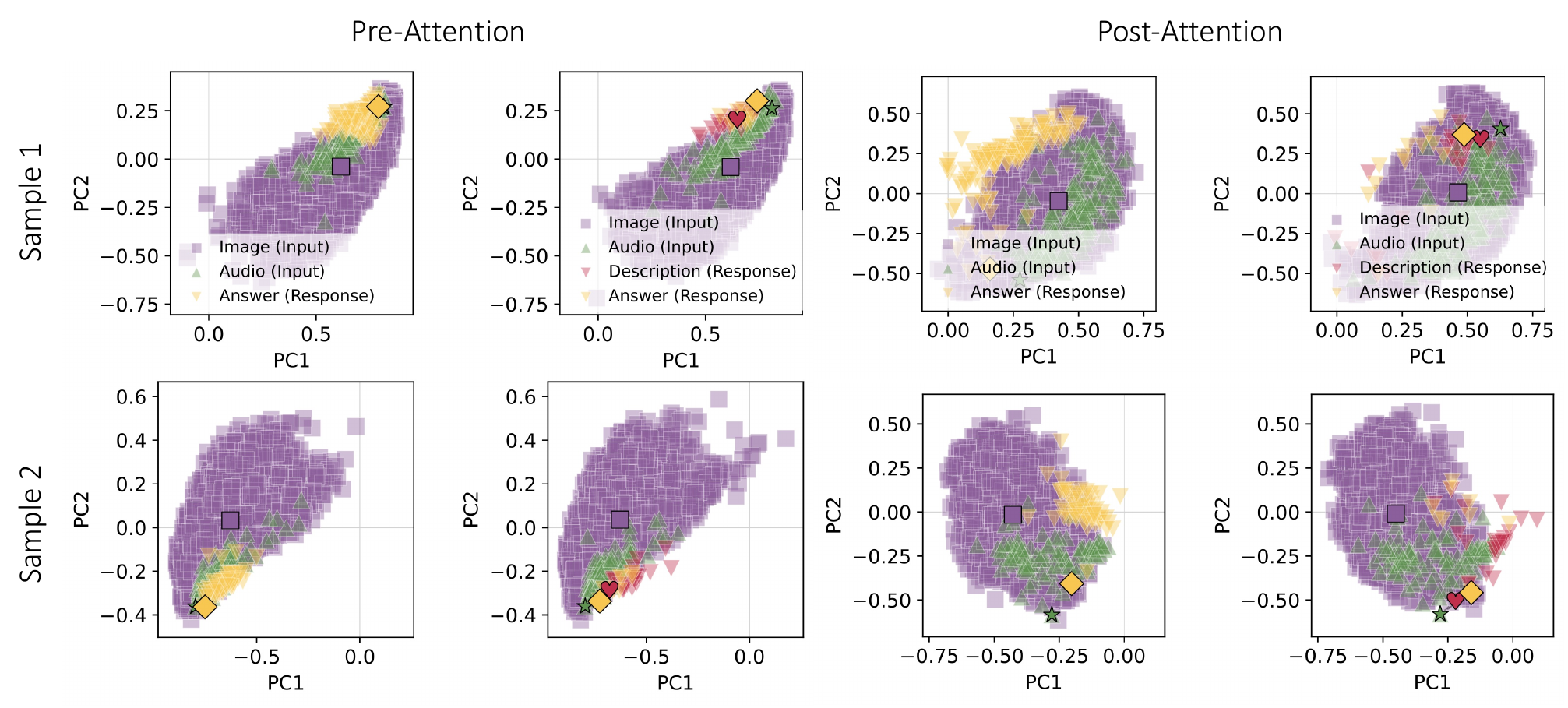}
    \vspace{-1.5em}
    \caption{\texttt{Qwen-7b}}
    \label{supp_fig:hiddenrep_qwen7b}
  \end{subfigure}
  \vspace{0.5em}
  \begin{subfigure}{\linewidth}
    \centering
    \includegraphics[width=\linewidth]{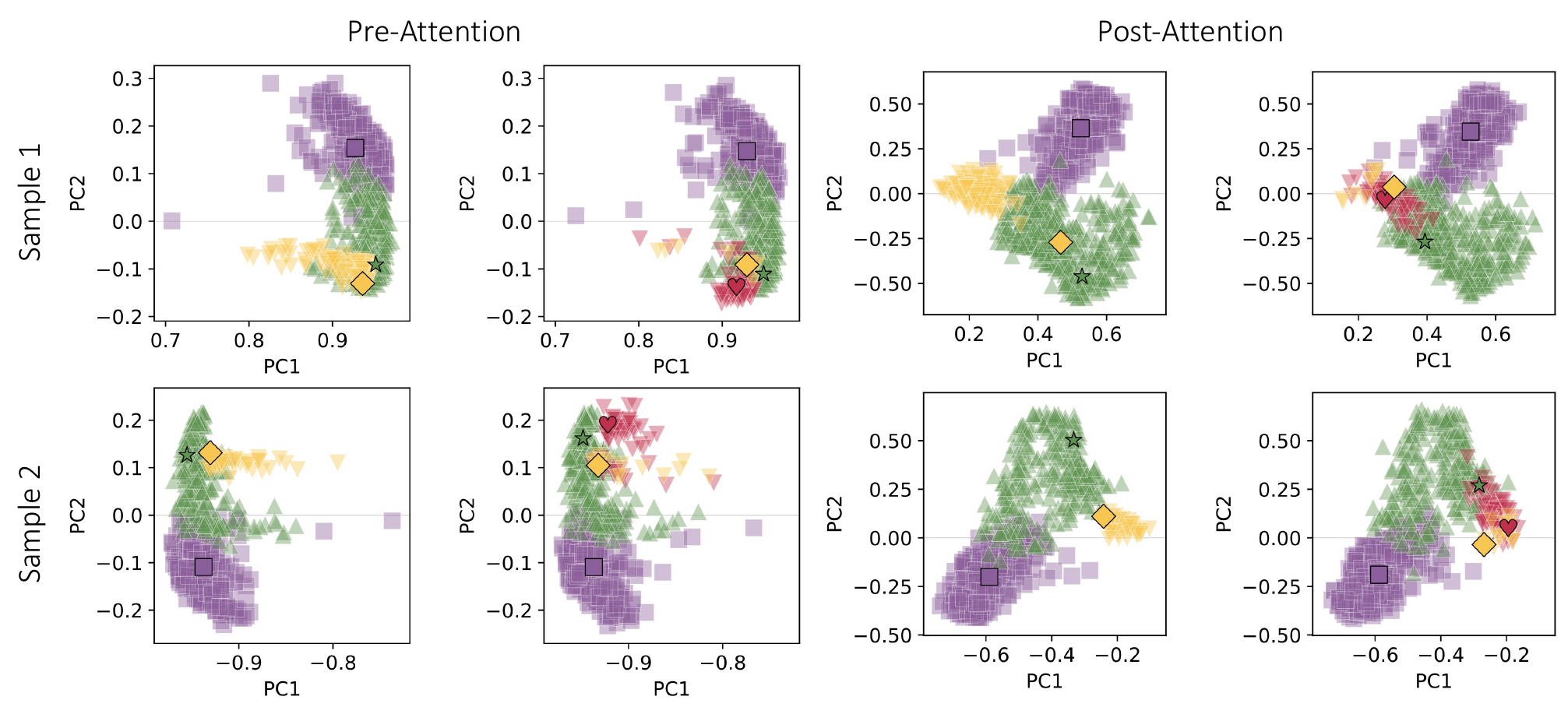}
    \vspace{-1.5em}
    \caption{\texttt{gemma-4b}}
    \label{supp_fig:hiddenrep_gemma4b}
  \end{subfigure}
  \vspace{-2em}
  \caption{\textbf{PCA projections of token representations before and after attention in the last layer for (a) \texttt{Qwen-7b} and (b) \texttt{gemma-4b}.}
  Each point per plot corresponds to a token of the image, audio, answer, and self-generated audio description.
  In both models, the audio description representations share the same PCA coordinate space as the audio and answer representations, unlike the image.}
  \label{supp_fig:hiddenrep}
  \vspace{-0.5em}
\end{figure*}

\noindent\textbf{Hidden-state visualization.}
Figures \ref{supp_fig:hiddenrep_qwen7b} and \ref{supp_fig:hiddenrep_gemma4b} visualize PCA projections of token representations before and after attention in the final layer.
For both \texttt{Qwen-7B} and \texttt{Gemma-4B}, the self-generated audio-description representations occupy a representation space closer to the audio and answer representations than to the image representations.
This qualitative pattern is consistent with the description's role as a textual medium that makes audio information more accessible to the language-generation pathway.

\begin{figure*}[!ht]
  \centering
  \includegraphics[width=\linewidth]{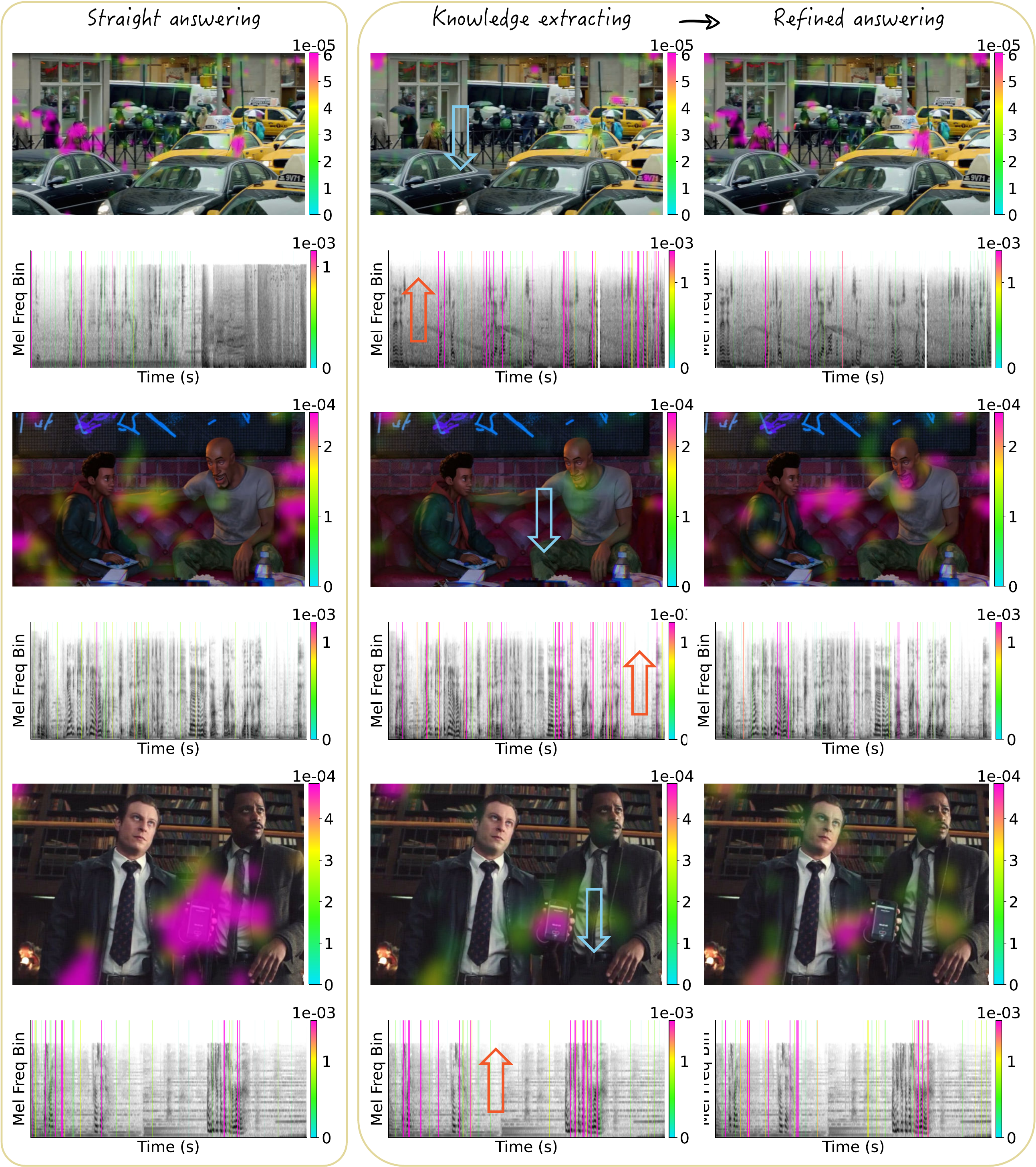}
  \vspace{-1.5em}
  \caption{\textbf{Probe of how Omni-LLM attention changes using our TMA filter.}
  The knowledge-extraction step in \emph{trained} \ours{} enables the model to ``hear'' better, which in turn affects the refined answering stage, showing more balanced ``seeing'' and ``hearing.''}
  \label{supp_fig:qualitative_trained_qwen7b}
\end{figure*}

\noindent\textbf{Qualitative results.}
Figure~\ref{supp_fig:qualitative_trained_qwen7b} shows an example from trained \ours{}.
The knowledge-extraction stage places greater emphasis on audio evidence, which subsequently affects the refined answering stage.
The resulting attention pattern is more balanced between visual and auditory evidence, illustrating how learned modality descriptions can alter the information available to downstream generation.

\begin{figure*}[!ht]
  \centering
  \begin{subfigure}[t]{0.33\linewidth}
    \centering
    \vspace{-47em}
    \includegraphics[width=\linewidth]{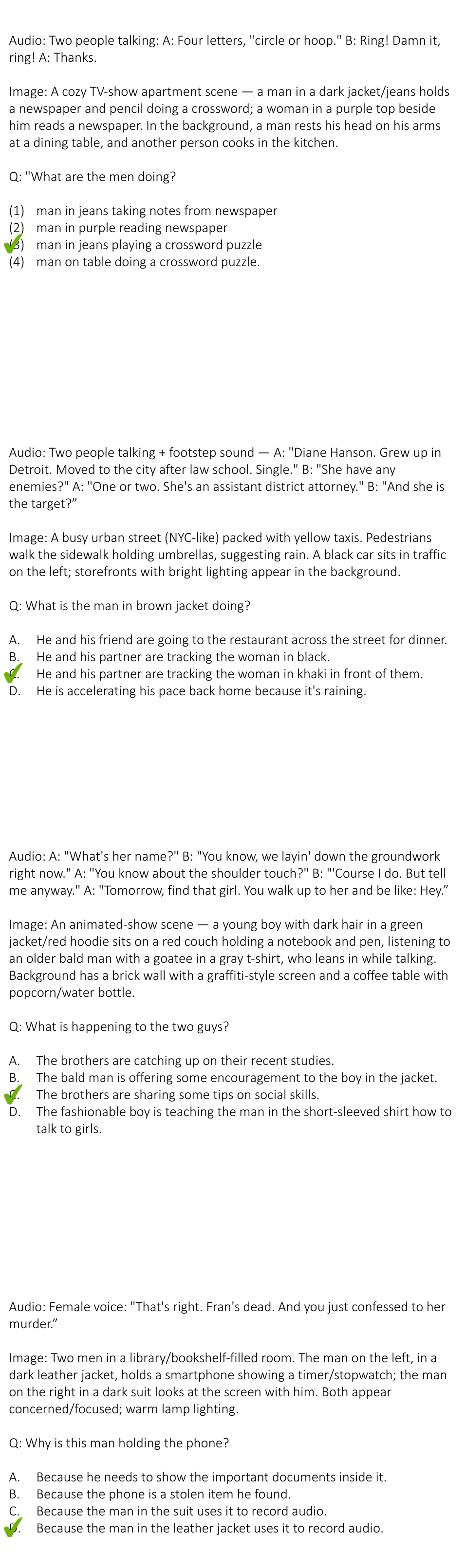}
    \vspace{1.6em}
    \caption{Benchmark question}
    \label{supp_fig:qualitative_tma_decoded_qa}
  \end{subfigure}\hfill
  \hspace{-1em}
  \begin{subfigure}[t]{0.33\linewidth}
    \centering
    \includegraphics[width=\linewidth]{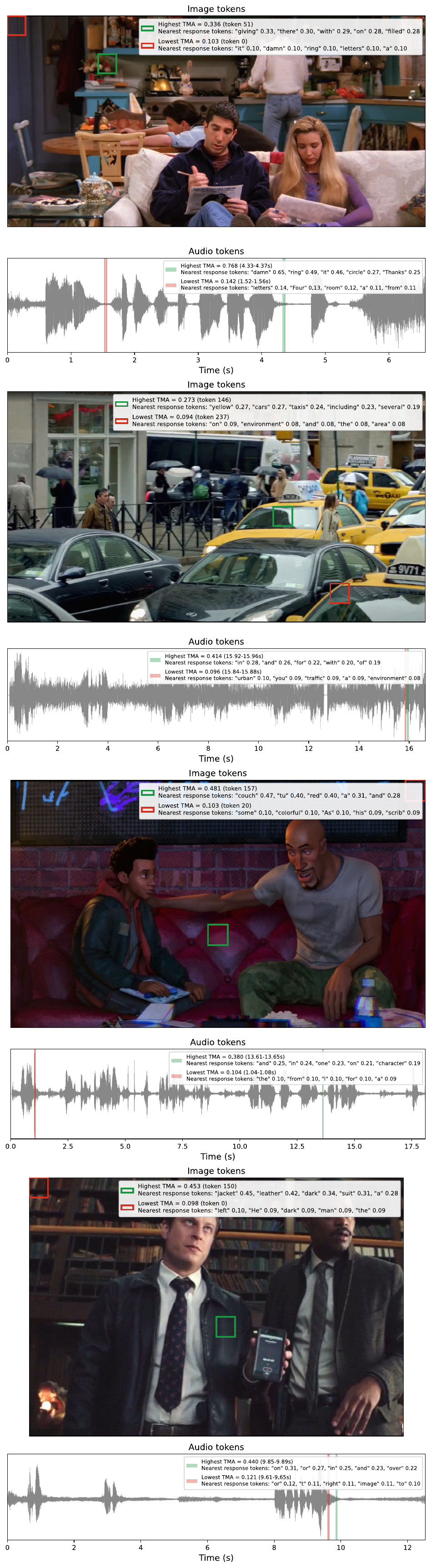}
    \caption{\texttt{gemma-4b}}
    \label{supp_fig:qualitative_tma_decoded_gemma}
  \end{subfigure}\hfill
  \begin{subfigure}[t]{0.33\linewidth}
    \centering
    \includegraphics[width=\linewidth]{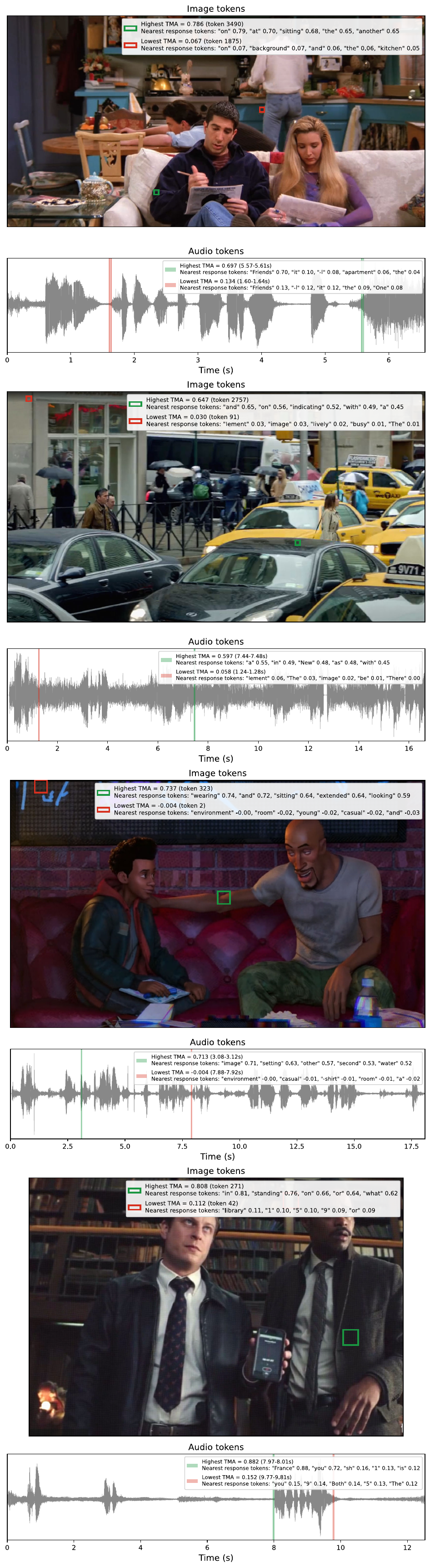}
    \caption{\texttt{Qwen-7b}}
    \label{supp_fig:qualitative_tma_decoded_qwen}
  \end{subfigure}
  \vspace{-0.5em}
  \caption{\textbf{Successful qualitative samples.}
  While the lowest-TMA tokens fall on background or silence, the highest-TMA image and audio tokens land on the visual regions and speech segments the answer depends on.}
  \label{supp_fig:qual_decoded_success}
\end{figure*}

\begin{figure*}[!ht]
  \centering
  \begin{subfigure}[t]{0.33\linewidth}
    \centering
    \vspace{-46em}
    \includegraphics[width=\linewidth]{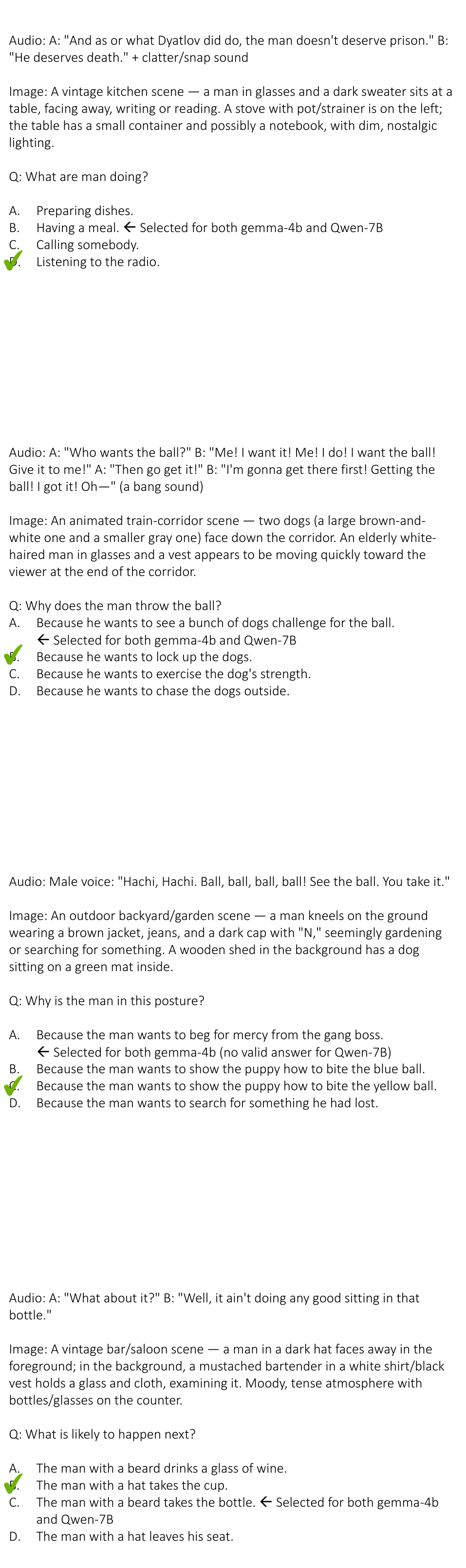}
    \vspace{0.7em}
    \caption{Benchmark question}
    \label{supp_fig:qualitative_tma_decoded_qa_unsuccessful}
  \end{subfigure}\hfill
  \hspace{-1em}
  \begin{subfigure}[t]{0.33\linewidth}
    \centering
    \includegraphics[width=\linewidth]{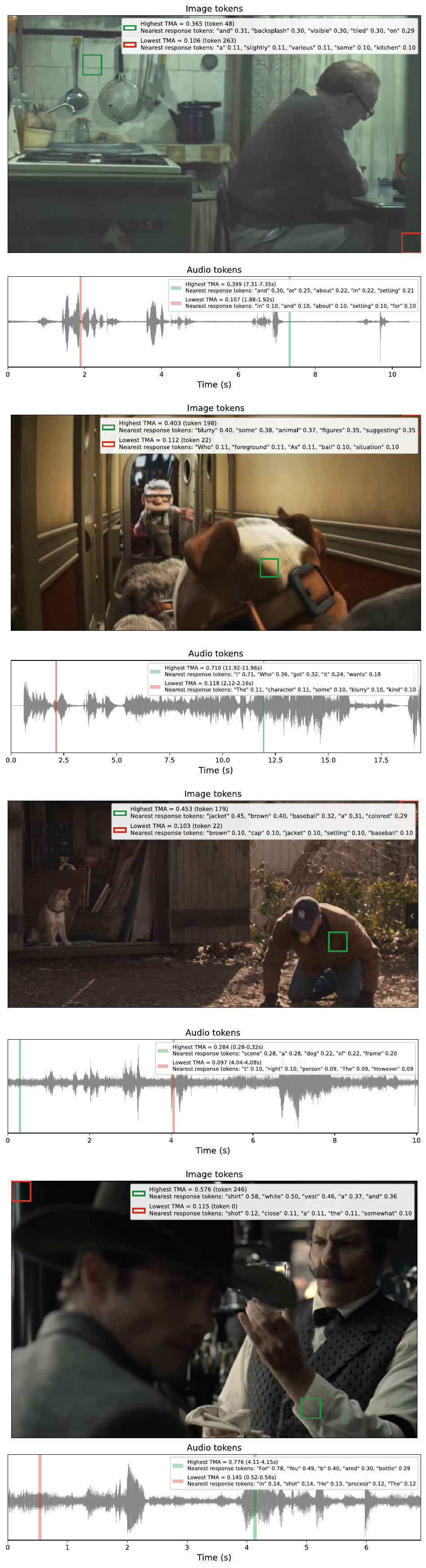}
    \caption{\texttt{gemma-4b}}
    \label{supp_fig:qualitative_tma_decoded_gemma_unsuccessful}
  \end{subfigure}\hfill
  \begin{subfigure}[t]{0.33\linewidth}
    \centering
    \includegraphics[width=\linewidth]{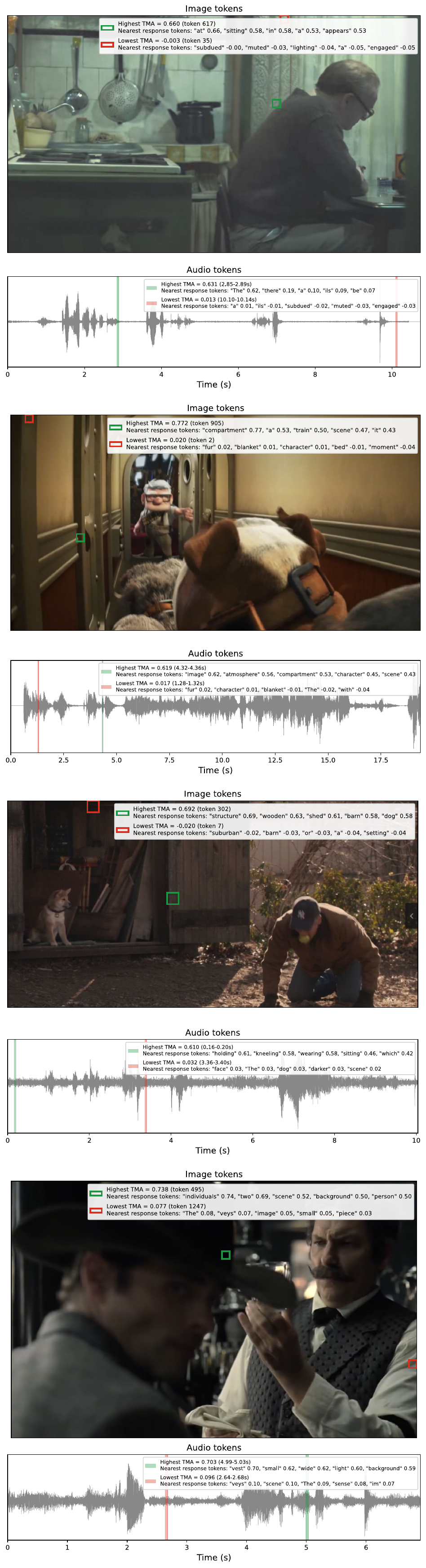}
    \caption{\texttt{Qwen-7b}}
    \label{supp_fig:qualitative_tma_decoded_qwen_unsuccessful}
  \end{subfigure}
  \vspace{-0.5em}
  \caption{\textbf{Unsuccessful qualitative samples.}
  When the answer requires reasoning beyond direct perception, such as off-screen sounds, intent, or future events, the highest-TMA tokens spread onto weakly related regions and segments.
  In other words, the evidence is too diffuse to capture with a single most-aligned token.}
  \label{supp_fig:qual_decoded_unsuccess}
\end{figure*}

Figures~\ref{supp_fig:qual_decoded_success} and~\ref{supp_fig:qual_decoded_unsuccess} present additional successful and unsuccessful examples of training-free \ours{}, together with the corresponding decoded TMA. In each panel, we highlight the image and audio tokens with the highest TMA values in red and the lowest in green, using bounding boxes on video frames and vertical markers on audio waveforms. We also report their TMA values and the response tokens closest to them in embedding space.
In the successful examples (Figure~\ref{supp_fig:qual_decoded_success}), the lowest-TMA tokens consistently correspond to background image regions or silent audio segments, whereas the highest-TMA tokens correspond to visual regions or speech segments relevant to the correct answer. These include the newspaper and crossword-related speech in the first example, and the pedestrian walking behind the others and the dialogue identifying the target in the second. This spatial and temporal alignment suggests that token-level TMA reflects the relevance of the input evidence to the generated response.
In the unsuccessful examples (Figure~\ref{supp_fig:qual_decoded_unsuccess}), this correspondence is less apparent. When a question requires reasoning beyond directly observable evidence, such as inferring off-screen sounds, intentions, or future events, the highest-TMA token corresponds to a region or temporal segment with limited relevance to the answer. Examples include the wall or shirt rather than the man, and the bartender's sleeves or the man's hat rather than the cup. These cases illustrate the limitations of relying on a single maximally aligned token as sufficient evidence, as some questions require integrating information across multiple regions or modalities.

\end{document}